\documentclass{article} 
\usepackage{iclr2027_conference,times}

\usepackage{amsmath,amsfonts,bm}

\def\eqref#1{equation~\ref{#1}}

\def\1{\bm{1}}

\DeclareMathAlphabet{\mathsfit}{\encodingdefault}{\sfdefault}{m}{sl}
\SetMathAlphabet{\mathsfit}{bold}{\encodingdefault}{\sfdefault}{bx}{n}

\usepackage{hyperref}
\usepackage{url}
\usepackage{graphicx}
\usepackage{tikz}
\usepackage{adjustbox}
\usepackage{amsthm}
\usepackage{booktabs}
\usepackage{tcolorbox}
\usepackage{multirow}
\newtheorem{definition}{Definition}
\usetikzlibrary{arrows.meta, positioning, shapes.geometric}

\title{Estimating Uncertainty of Omnimodal Large Language Models via External Omnimodal Embeddings}

\author{Iacopo Ghinassi, Hy Nguyen Gia La \& Luke Ong \\
CCDS\\
Nanyang Technological University\\
Singapore, 639798, Singapore \\
\texttt{iacopo.ghinass@ntu.edu.sg}
}

\iclrfinalcopy

\begin{document}

\maketitle

\begin{abstract}
Recent work has investigated Omni-modal Large Language Models (OLLMs), models which can take multiple modalities and their combinations as input, as a way forward to have systems really capable of interacting with their environment. Uncertainty quantification (UQ) for OLLMs, however, remains an open problem: existing methods largely target at most two modalities at a time, require access to model internals, or re-invocation of the evaluated model.
We propose a model-agnostic UQ framework that scores a Chain-of-Thought response entirely through external omni-modal embeddings, measuring the \emph{grounding} of the reasoning in the multimodal input and its internal \emph{coherence}. Our contribution is the family of signals and an empirical characterisation of when it is reliable, rather than any single estimator: we find the geometric scores to be largely redundant with one another (mean within-split rank correlation $0.91$--$0.94$) but strongly complementary to answer-dispersion baselines, and their reliability to depend on the generator and the modality. The resulting confidence estimates require no access to logits, hidden states or gradients of the evaluated model and no additional inference from it; they do require segmenting the reasoning trace with an auxiliary model and, for the aggregated score, a labelled calibration set that we show can be as small as 25 examples --- although we also show that such a set buys little over fixing one signal a priori.
On UNO-Bench, with both a closed and an open model, our scores yield small gains over self-consistency --- systematic with the closed generator, mixed with the open one --- and are competitive with self-verbalisation and internal-state baselines, especially in open-ended, omni-modal questions where previous state-of-the-art baselines struggle.\footnote{Code: https://anonymous.4open.science/r/Multimodal-CoT-Confidence-Scoring-C12E}
\end{abstract}

\section{Introduction}
Omni-modal large language models (OLLMs) process and reason over diverse modalities including text, images, audio and video. Assessing the reliability of their outputs through Uncertainty Quantification (UQ) is an open problem.

Recent attempts at UQ for multimodal models \citep{lau2026uncertainty} exhibit major limitations: (1) they require access to model internal states or final logits, limiting their scope to open-source models; (2) they are designed primarily for bimodal interactions (e.g. audio-text) instead of omni-modal (e.g. audio, image and text at the same time); and (3) they typically rely solely on the final answer, ignoring the information contained in the model's reasoning trace or Chain-of-Thought (CoT).

To address these issues, we introduce a novel model-agnostic UQ method for omni-modal contexts, which derives its confidence signal from the geometry of the model's reasoning trace in the shared space of omni-modal embedding models external to the analysed model. Our approach maps to a shared embedding space both the multi-modal inputs, the reasoning steps and the final answer, deriving \emph{grounding} and \emph{coherence} scores based on the distance between such embeddings. Such scores are then used to grade the majority-selected answer 
so that ungrounded and low-coherence answers get less \emph{voting power}. We make three claims. First, useful uncertainty signals for omni-modal reasoning can be read off an external embedding space alone, with no access to the evaluated model's logits, hidden states or gradients and no additional inference from it. Second, these scores are interpretable diagnostics of \emph{one} underlying quantity, the geometric regularity of a reasoning trace, rather than independent evidence sources: they correlate at $0.91$--$0.94$ within a split, which is why combining them does not beat selecting one, while being strongly decorrelated from --- and complementary to --- between-sample answer dispersion. Third, after a small calibration step they give small but statistically systematic gains over self-consistency with a closed generator and smaller, split-dependent gains with an open one, largest at low sampling budgets and on open-ended questions spanning more than two modalities. We regard the characterisation of \emph{when} these signals are reliable as a contribution on a par with the scores themselves.

The comparison with self-verbalisation and internal-state baselines is mixed: our scores remain the strongest family on open-ended, omni-modal questions under both answer-generating models, while with the open model the strongest self-verbalisation baseline and the internal-state baseline match our aggregated score on discrimination averaged across splits; a simpler, reasoning-free measure of answer dispersion that we also evaluate is the one signal that repairs the audio failure mode where every agreement-based signal inverts, while dropping to chance level exactly on the omni-modal splits where our scores are strongest; this suggests that such signals are complementary rather than one subsuming the other. Our scores are nonetheless markedly more interpretable than self-consistency and, especially, previous embedding-based methods relying on internal states and answer-only embeddings, since they have a straightforward geometric interpretation rooted in well-defined individual scores.
Finally, we release both CoT datasets used in our experiments, derived from answering the questions of an indicative benchmark with a closed and an open-source model respectively.
We hope that by doing so we will be able to foster research in this interesting and still relatively under-developed area of research.

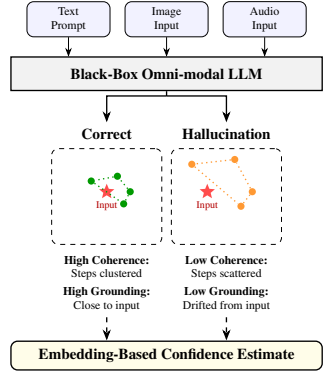
\begin{figure}[t]
    \centering
    \resizebox{0.3\columnwidth}{!}{%
    \begin{tikzpicture}[
        >=Stealth,
        node distance=1.5cm,
        inputbox/.style={draw, rounded corners, fill=blue!5, text width=1.5cm, align=center, font=\scriptsize},
        modelbox/.style={draw, thick, fill=gray!10, minimum width=6.5cm, minimum height=0.7cm, font=\small\bfseries},
        spacebox/.style={
    draw,
    dashed,
    rounded corners,
    minimum width=2.3cm,
    minimum height=2.0cm
},
        staranchor/.style={star, star points=5, star point ratio=2.25, fill=red!70, inner sep=1.5pt},
        cotstep/.style={circle, fill=green!60!black, inner sep=1.5pt},
        cotstepbad/.style={circle, fill=orange!80, inner sep=1.5pt},
        textdesc/.style={font=\scriptsize, align=center}
    ]

    \node[inputbox] (text) {Text\\Prompt};
    \node[inputbox, right=0.3cm of text] (image) {Image\\Input};
    \node[inputbox, right=0.3cm of image] (audio) {Audio\\Input};

    \node[modelbox, below=0.4cm of image] (ollm) {Black-Box Omni-modal LLM};

    \draw[->] (text.south) -- (ollm.north -| text.south);
    \draw[->] (image.south) -- (ollm.north);
    \draw[->] (audio.south) -- (ollm.north -| audio.south);

    \node[spacebox] (reliable_space) at ([xshift=-1.25cm,yshift=-2.2cm]ollm.south) {};
\node[spacebox] (uncertain_space) at ([xshift=1.25cm,yshift=-2.2cm]ollm.south) {};

    \node[textdesc, above=0.07cm of reliable_space.north, font=\footnotesize\bfseries] (rel_title) {Correct};
    \node[textdesc, above=0.07cm of uncertain_space.north, font=\footnotesize\bfseries] (unc_title) {Hallucination};

    \coordinate (fork) at ([yshift=-0.1cm]ollm.south);
    \draw[thick] (ollm.south) -- (fork);
    \draw[->, thick] (fork) -| (rel_title.north);
    \draw[->, thick] (fork) -| (unc_title.north);

    \node[staranchor, label={[font=\tiny, text=red!70!black]below:Input}] at ([yshift=0.1cm]reliable_space.center) (anchor_rel) {};
    
    \node[cotstep, above right=0.2cm and 0.1cm of anchor_rel] (s1) {};
    \node[cotstep, above left=0.1cm and 0.2cm of anchor_rel] (s2) {};
    \node[cotstep, below right=0.1cm and 0.2cm of anchor_rel] (s3) {};
    \node[cotstep, right=0.3cm of anchor_rel] (s4) {};
    \draw[dotted, thick, green!60!black] (s1) -- (s2) -- (s3) -- (s4) -- (s1);
    
    \node[textdesc, below=0.1cm of reliable_space.south] (rel_desc) {\textbf{High Coherence:}\\Steps clustered\\[1mm]\textbf{High Grounding:}\\Close to input};

    \node[staranchor, label={[font=\tiny, text=red!70!black]below:Input}] at ([yshift=0.1cm, xshift=-0.4cm]uncertain_space.center) (anchor_unc) {};
    
    \node[cotstepbad, above right=0.55cm and 0.45cm of anchor_unc] (u1) {};
\node[cotstepbad, above left=0.45cm and 0.20cm of anchor_unc] (u2) {};
\node[cotstepbad, right=0.75cm of anchor_unc] (u3) {};
\node[cotstepbad, below right=0.15cm and 0.55cm of anchor_unc] (u4) {};
    \draw[dotted, thick, orange!80] (u1) -- (u2) -- (u4) -- (u3) -- (u1);

    \node[textdesc, below=0.1cm of uncertain_space.south] (unc_desc) {\textbf{Low Coherence:}\\Steps scattered\\[1mm]\textbf{Low Grounding:}\\Drifted from input};

    \node[draw, thick, fill=yellow!10, minimum width=6.5cm, minimum height=0.6cm, rounded corners, font=\small\bfseries] (output) at ([yshift=-0.8cm]ollm.south |- rel_desc.south) {Embedding-Based Confidence Estimate};
    
    \draw[->, thick, dashed] (rel_desc.south) -- (rel_desc.south |- output.north);
    \draw[->, thick, dashed] (unc_desc.south) -- (unc_desc.south |- output.north);

    \end{tikzpicture}
    }
    \vspace{-3mm}
    \caption{We map the omni-modal inputs and generated Chain-of-Thought reasoning steps into a shared semantic space to compute \textbf{Coherence} and \textbf{Grounding}, which grade the majority-selected answer without internal model logits and without any additional call to the evaluated model.}
    \label{fig:teaser}
\end{figure}

\section{Related Work}
\subsection{LLM Uncertainty Quantification}
Literature on uncertainty quantification for LLMs is abundant, starting from the seminal work by \citet{wang2022selfconsistency} showing that self-consistency is a good signal of LLMs certainty over their output. Most works in this context, then, have developed further the self-consistency direction with better answer matching \citep{farquhar2024semanticentropy}. At the same time, other work has shown that the log-probabilities of the models could also serve as a strong signal of model (un)certainty \citep{fadeeva-etal-2024-fact}, even though its alignment to true probabilities has been questioned \citep{toney-wails-2025-certain}. Similarly, another line of work has explored the internal states of the LLMs as a strong signal for uncertainty measure \citep{10.1609/aaai.v40i36.40259}. The preponderance of private models, for which neither log probabilities nor hidden states are available, has however led much of the literature to focus on black-box, external methods such as self-consistency. One such method consists in asking the LLMs themselves to \emph{verbalize} their confidence: such methods tend to work well for powerful models, but they fail when the LLM is not powerful enough and they tend to be overconfident \citep{mei-etal-2026-reasoning}.

\subsection{Uncertainty Quantification for Multimodal LLMs}
Recent work has started exploring the space of multimodal LLMs, starting from Visual LLMs \citep{zhang2024vl}. The biggest challenge in this context is potential conflict between modalities \citep{pmlr-v258-bezirganyan25a}. However, current methods mostly follow approaches such as self-verbalisation, collapsing the role of different modalities into a single judgment \citep{chen-etal-2025-unveiling}. A different approach, UMPIRE \citep{lau2026uncertainty}, was recently proposed which instead uses the internal states of the LLM to adjust Semantic Volume. Also this method, however, presents the problem of needing access to some model internal states. In case of private models, the authors circumvented this problem by feeding the input and answer to an open source multimodal LLM, but also this solution can be seen as suboptimal since it involves recomputing the entire states for a second model and it is contingent on how \emph{similar} to the private model the open source model is. 

\section{Methodology}
\label{sec:methodology}

To address the limitations of prior work, we introduce a fully external framework that assesses the confidence of a generated answer from the \emph{coherence} and \emph{grounding} of its reasoning chain, measured in the embedding space of an external encoder. The resulting scores need no access to the internals of the evaluated model --- logits, hidden states or gradients --- and no additional inference from it; segmenting the trace into steps is done once by a separate auxiliary model, and the aggregated score assumes a small labelled calibration sample. They are used to \emph{weight the votes} of each candidate answer, so that answers supported by coherent, well-grounded reasoning count more than answers supported by erratic or ungrounded chains. 

Specifically, we define \emph{coherence} and \emph{grounding} in such omni-modal embedding space as:

\begin{enumerate}
\item \emph{grounding}: the semantic alignment, in the shared embedding space, between each intermediate reasoning step and the non-textual multi-modal inputs.
\item \emph{coherence}: the dispersion of the embeddings of the steps of the generated reasoning chain (based on text only), which we further decompose into internal reasoning smoothness, density and convergence toward the predicted answer.
\end{enumerate}

\subsection{Problem Formulation}

Let $\mathcal{M}$ denote a blackbox omni-modal model, which receives a multimodal context

\[
\mathcal{V} = \bigcup_{k=1}^K V^{(k)}, \quad V^{(k)} = \{v^{(k)}_1, \dots, v^{(k)}_{n_k}\},
\]

where each modality-specific subset $V^{(k)}$ collects the elements of one input modality. The pool contains only the non-textual inputs (audio, image, video); the textual prompt is deliberately excluded, so grounding cannot be satisfied by aligning a step with the question text. We assume access to the chain-of-thought process of the model, thus we define the model-generated reasoning chains

\[
R = [r_1,r_2,\dots,r_m]
\]

followed by a final answer $a$.

Because internal representations of $\mathcal{M}$ might be unavailable, we instead rely on an external embedding model

\[
\mathcal{E}: x \rightarrow \phi_x \in \mathbb{R}^d .
\]

We embed all components into a shared space:

\[
\Phi_{\mathcal{V}} = \{\phi_{v} : v \in \mathcal{V}\}, \quad
\Phi_R = [\phi_{r_1},\dots,\phi_{r_m}], \quad
\phi_a .
\]

Let $\mathcal{S}(\cdot,\cdot)$ denote the normalized cosine similarity defined as
\[
\mathcal{S}(x,y) = \frac{1 + \cos(\phi_x, \phi_y)}{2},
\]
which maps similarity scores to the interval $[0,1]$.

\subsection{Grounding}
\label{subsec:grounding}

We quantify grounding by measuring the alignment between reasoning steps and input modalities in a shared omni-modal embedding space.

Recall that the input context $\mathcal{V} = \bigcup_{k=1}^{K} V^{(k)}$ is partitioned into the $K$ modality-specific subsets introduced above. The embedding model $\mathcal{E}$ maps each subset to its set of embeddings in the shared space:
\begin{equation}
\Phi^{(k)} = \{\phi_{v} : v \in V^{(k)}\}, \quad k = 1, \dots, K.
\end{equation}

\begin{definition}[Omni-modal Grounding]
\label{def:grounding}
For a reasoning step $r_i$ with embedding $\phi_{r_i} = \mathcal{E}(r_i)$, its alignment to modality $k$ is defined as:
\begin{equation}
A^{(k)}(r_i) = \max_{\phi_v \in \Phi^{(k)}} \mathcal{S}(\phi_{r_i}, \phi_v),
\label{eq:align}
\end{equation}
where $\mathcal{S}(\cdot, \cdot)$ is the normalized cosine similarity defined above.

Since we expect an individual reasoning step to refer to at most one modality at a time, we then obtain the omni-modal alignment $A^{\mathrm{omni}}(r_i)$ by applying the max operator again, this time over the marginal alignments $A^{(k)}(r_i)$:
\begin{equation}
    A^{\mathrm{omni}}(r_i)=\max_{k = 1,\ldots,K}A^{(k)}(r_i)
\end{equation}

Then, having the alignments for each reasoning step $i$, we can aggregate with simple averaging:

\begin{equation}
    G_{\text{avg}}=\frac{1}{m}\sum^m_{i=1}A^{\mathrm{omni}}(r_i)
\end{equation}

We further consider the max variant to factor in the most grounded step.

\begin{equation}
    G_{\max}=\max_{i=1,\ldots,m} A^{\mathrm{omni}}(r_i)
\end{equation}

\end{definition}

\paragraph{Answer-level grounding.} We additionally embed each candidate generation as a single unit and align it with the same input pool, giving three further scores $G^{\max}_{\text{ans}}$, $G^{\text{avg}}_{\text{ans}}$ and $G^{\min}_{\text{ans}}$ (Appendix~\ref{app:aggregator-evidence}, Equation~\ref{eq:ansground}); the last requires support from \emph{every} present modality. They cost one embedding comparison per candidate and no extra model call.

We note that, when a single shared encoder embeds all modalities the maximum over the marginal alignments $A^{(k)}(r_i)$ coincides with the maximum over the pooled inputs, so the per-modality formulation adds discriminative power only once modality-specific encoders or richer aggregation schemes are used; we evaluate one such modality-aware aggregation in Appendix~\ref{app:grounding-attention}.

\subsection{Answer Coherence}

A \emph{coherent} answer should exhibit gradual semantic transitions rather than abrupt jumps.  
We capture this property using three complementary geometric measures measured directly from the textual output of the OLLM (i.e. the reasoning chain).

\paragraph{Local Smoothness}
This quantity measures the local transition between one reasoning step to the next one by comparing pairwise, consecutive steps in the embedding space. Specifically, we compute: 

\begin{equation}
S_{\text{smooth}} =
\frac{1}{m-1}
\sum_{i=1}^{m-1}
\mathcal{S}(\phi_{r_i},\phi_{r_{i+1}})
\label{eq:smooth}
\end{equation}

Low smoothness indicates erratic reasoning transitions, highlighting low local coherence and possible hallucinations. Equations~\ref{eq:smooth} and~\ref{eq:dens}, and $\Delta_{\text{bwd}}$ below, are undefined for single-step chains; our implementation then assigns all three scores their maximum value $1$, which inflates the coherence of terse chains (Appendix~\ref{app:aggregator-evidence} reports how often this occurs).

\paragraph{Semantic Density}
Semantic density measures the global coherence of the reasoning chain by comparing each step embeddings with each other and normalizing:

\begin{equation}
S_{\text{dens}} =
\frac{1}{m(m-1)}
\sum_{i \neq j}
\mathcal{S}(\phi_{r_i},\phi_{r_j})
\label{eq:dens}
\end{equation}

A low density in the embedding space indicates low global compactness within the reasoning chain, which has been shown to correlate with lower coherence \citep{ghinassi-etal-2024-cohesion}. 

\begin{figure*}
    \centering
    \includegraphics[width=0.8\linewidth]{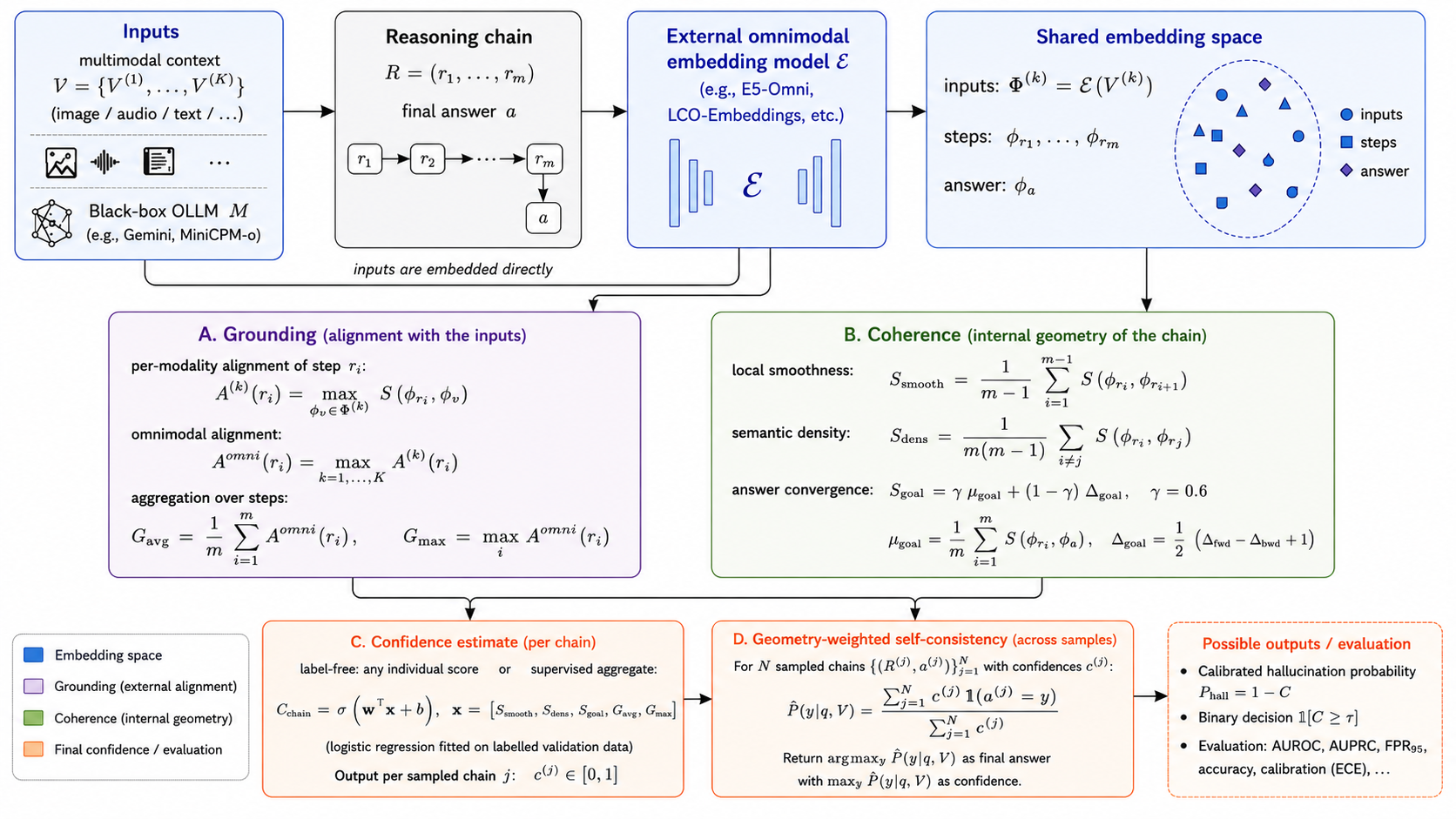}
    \caption{Visual overview of the proposed method, as explained in Methodology section. 
    }
    \label{fig:method-overview}
\end{figure*}

\paragraph{Answer Convergence}

A reliable reasoning trajectory should also progressively move toward the final answer, a property also of a text coherently oriented towards a goal.

Let

\[
\mu_{\text{goal}} =
\frac{1}{m}
\sum_{i=1}^{m}
\mathcal{S}(\phi_{r_i},\phi_a)
\]

denote the average proximity of reasoning steps to the answer embedding.

To capture directional progression, we define forward and backward
alignments as

\begin{equation}
\Delta_{\text{fwd}} =
\mathbb{E}_{i>m/2}
[\mathcal{S}(\phi_{r_i},\phi_a)]
\end{equation}

\begin{equation}
\Delta_{\text{bwd}} =
\mathbb{E}_{i\le m/2}
[\mathcal{S}(\phi_{r_i},\phi_a)]
\end{equation}

and compute the directional progression score as

\begin{equation}
\Delta_{\text{goal}} =
\frac{1}{2}
\left(
\Delta_{\text{fwd}}
-
\Delta_{\text{bwd}}
+1
\right).
\end{equation}

which measures whether later reasoning steps move closer to the answer.

The final convergence score is

\begin{equation}
S_{\text{goal}} =
\gamma \mu_{\text{goal}}
+
(1-\gamma)\Delta_{\text{goal}}
\end{equation}

where $\gamma$ is a weighting hyperparameter, fixed a priori to $\gamma = 0.6$ in all experiments.

\subsection{Unified Chain-Level Score}

The extracted chain-level signals
\begin{equation}
\mathbf{x} =
\left[
S_{\text{smooth}},
S_{\text{dens}},
S_{\text{goal}},
G_{\text{avg}},
G_{\max}
\right]
\label{eq:feature-vector}
\end{equation}
are used as feature inputs to an $\ell_1$-regularised logistic regression classifier. The classifier is
fitted on the validation set using answer correctness as the target label,
learning the relationship between these signals and the probability that a
generated answer is correct. Specifically, the predicted correctness
probability is computed as

\begin{equation}
\mathcal{C}_{\text{chain}}
=
\sigma
\left(
\mathbf{w}^{\top}\mathbf{x}+b
\right),
\end{equation}

where $\sigma(\cdot)$ denotes the sigmoid function, and $\mathbf{w}$ and $b$
are learned parameters. Fitting has two stages (Appendix~\ref{app:experiments}): a grid search on validation AUROC, then an explicit comparison against each single-feature scorer, which replaces the fitted model whenever strictly better. That selection step, not the $\ell_1$ penalty, is what yields the single-feature solutions of Appendix~\ref{app:ablations}, so $\mathcal{C}_{\text{chain}}$ is validation-based selection with a logistic-regression fallback rather than a multivariate combination.

We also report $\mathcal{C}^{+}_{\text{chain}}$, fitted with the identical protocol on Equation~\ref{eq:feature-vector} extended by the three answer-level grounding scores of Equation~\ref{eq:ansground} (Appendix~\ref{app:aggregator-evidence}).

The final uncertainty estimate is then defined as the complement of the
predicted correctness probability:

\begin{equation}
\mathcal{U}
=
1-\mathcal{C}_{\text{chain}} .
\end{equation}

\subsection{Geometry-Weighted Aggregation of Sampled Answers}
\label{sec:methodology-gwsc}

Standard self-consistency treats all sampled reasoning chains equally: given $N$ sampled generations $\{(R^{(j)},a^{(j)})\}_{j=1}^{N}$, it estimates $\hat{P}(y|q,\mathcal{V})=\frac{1}{N}\sum_{j}\mathbf{1}(a^{(j)}=y)$ which is treated as the model confidence. We instead compute for each sampled chain the feature vector $\mathbf{x}^{(j)}$ of Equation~\ref{eq:feature-vector} and a chain-level confidence $c^{(j)}$ (either from the logistic-regression aggregator, $c^{(j)}=\mathcal{C}_{\text{chain}}(\mathbf{x}^{(j)})$, or directly from a single score in the individual-score variants) and replace the vote frequencies by their geometry-weighted counterpart:
\begin{equation}
\hat{P}(y|q,\mathcal{V})
=
\frac{
\sum_{j=1}^{N}
c^{(j)}\mathbf{1}(a^{(j)}=y)
}{
\sum_{j=1}^{N} c^{(j)}
}.
\label{eq:gwsc}
\end{equation}

Equation~\ref{eq:gwsc} admits two modes. In the \emph{primary} protocol of the main tables the returned answer is the count-majority cluster and the geometry only attaches a confidence to it, so every method grades the same answer; in the \emph{active} mode (Appendix~\ref{app:gwsc}) the answer is $\arg\max_y \hat{P}(y|q,\mathcal{V})$ and the geometry can change the prediction. The overall pipeline of the proposed framework is summarized in Figure~\ref{fig:method-overview}. Unlike previous embedding-based methods, our procedure can be applied to any model, private or open, as it requires only externally observable artifacts: a generated reasoning chain and its final answer, embedded with an external encoder.

\section{Experimental Setup}
\label{sec:experimental-setup}
\textbf{Dataset \& Models}: We use the uni- and omni-modal benchmark UNO-bench \citep{chen2025unobenchunifiedbenchmarkexploring}, and report results on its four subsets corresponding to Audio only (Audio, 1{,}000 questions), Vision only (Visual, 1{,}480), Audio + Vision in a multiple choice answer setting (MC, 1{,}000) and Audio + Vision in an open-ended answer setting (MO, 250), together with their average (Avg).

We employ Gemini 3.1 Flash Lite and  MiniCPM-o 4.5 \citep{cui2026minicpmo45realtimefullduplex} to generate 6 answers for each question in the UNO-bench dataset: our methods, Self-Consistency, and UMPIRE aggregate over candidates, whereas the self-verbalisation baselines score the selected answer through one additional call to the evaluated model. For the UMPIRE baseline with the closed generator, we extract internal states using MiniCPM-o 4.5 as a proxy model; in-depth experimental details are documented in Appendix~\ref{app:experiments}.

\textbf{Embedding Model}: We used E5-Omni \citep{chen-etal-2026-e5} to generate embeddings. Appendix~\ref{app:lco-encoder} repeats the evaluation with a different omnimodal encoder, LCO-Embedding-Omni-7B \citep{NEURIPS2025_e839f274}, showing that our results are robust to this choice.

\textbf{Answer Selection}: 
Following \citet{farquhar2024semanticentropy}, sampled answers are grouped into semantic clusters by an LLM, and the candidate belonging to the most frequent cluster is selected with the methods scores breaking ties. The tie-break is method-dependent, so on tied questions --- $28.9\%$ with Gemini, $14.2\%$ with MiniCPM --- different methods can return different answers (Appendix~\ref{app:aggregator-evidence}). 
A variant in which the geometric evidence also \emph{selects} the answer, via the weighted vote of Section~\ref{sec:methodology-gwsc}, is evaluated in Appendix~\ref{app:gwsc}.

\textbf{Baselines}: We compare against two families of methods. The first requires repeated sampling from the evaluated model: \emph{self-consistency}, which uses the raw frequency of the selected answer's cluster among the $n=6$ candidates as its confidence, and \emph{UMPIRE}, which additionally requires access to internal states (extracted through a proxy model for closed-source generators). To the same sampling-based family belong two embedding-dispersion baselines computed from the candidate-\emph{answer} embeddings of the same external encoder used by our method: \emph{RDS} (radial dispersion score), the mean Euclidean distance of the $N$ candidate-answer embeddings from their centroid, and \emph{Semantic Volume}, the log-determinant of the Gram matrix of the mean-centred answer embeddings, $\log\det(XX^{\top}+\varepsilon I)$. 
The second family avoids repeated sampling but requires one additional call to the evaluated model: three SOTA self-verbalisation baselines. The first, which we call Level-Based, asks the model to assign a five-level confidence score to its answer, mapped to numerical values. The second and third follow Self-Probing~\citep{liu-etal-2024-llms-learn-uncertainty} and Two-Step CoT~\citep{tian-etal-2023-just}, respectively, where the model estimates answer correctness from the question-answer pair and previously generated reasoning context. All baseline implementation details are in Appendix~\ref{app:baselines}; Appendix~\ref{app:additional-baselines} analyses the dispersion baselines in depth and further compares against two cluster-based baselines (discrete Semantic Entropy and the number of semantic sets) computed with the same clustering step.

\textbf{Metrics}: We evaluate Area Under the Receiver Operating Characteristic curve (AUROC) and Expected Calibration Error (ECE). Since the compared methods produce confidence scores on heterogeneous scales, before computing the metrics each method's scores are min--max rescaled to $[0,1]$ over the evaluated split; this preprocessing leaves the rank-based AUROC unchanged and makes ECE comparable across methods. It is computed on the evaluated split itself, so it reflects range-normalised spread rather than deployment calibration (Appendix~\ref{app:variance}).
Appendix~\ref{app:aurac} additionally reports AURAC metric.
The main tables report mean values across 100 randomised runs. Statistical support is reported separately, in two complementary forms. First, the paired delta Table~\ref{tab:delta-combined} tests the difference between $\mathcal{C}_{\text{chain}}$ and Self-Consistency on each split with a paired $t$-test. Second, Section~\ref{sec:item-level} reports targeted paired-bootstrap comparisons over test questions on a canonical partition, which quantify generalisation to new questions; these are reported with their raw $p$-values. Appendix~\ref{app:variance} reports the underlying run-to-run standard deviations and item-level 95\% confidence intervals, together with a deployment-realistic calibration analysis.

\section{Results}
\label{sec:results}
We compare our individual scores and the aggregated $\mathcal{C}_{\text{chain}}$ against the sampling-based methods (Self-Consistency, UMPIRE, and the RDS and Semantic Volume answer-dispersion baselines) and the three self-verbalisation baselines; every method scores the same majority-selected answer (Section~\ref{sec:experimental-setup}), so only the attached confidence differs, and our scores add one external embedding pass and no call to the evaluated model beyond the sampled candidates.
MC and MO denote the OmniModal-Multiple Choice and OmniModal-Open Answer splits, respectively. Table~\ref{tab:sf-gemini} reports AUROC and ECE with Gemini-generated answers and Table~\ref{tab:sf-minicpm} the same metrics with MiniCPM-generated answers; the Avg column is the unweighted mean of the four per-split values (results on the pooled union of the splits, and the reasons why we do not use them as a summary statistic, are given in Appendix~\ref{app:overall}). Table~\ref{tab:delta-combined} summarises the paired differences of our best methods against Self-Consistency across the 100 runs.

\begin{table*}[!ht]
\centering
\begin{adjustbox}{width=\textwidth}
\begin{tabular}{l cc cc cc cc cc}
\toprule
& \multicolumn{2}{c}{Avg} & \multicolumn{2}{c}{Audio} & \multicolumn{2}{c}{MC} & \multicolumn{2}{c}{MO} & \multicolumn{2}{c}{Visual} \\
\cmidrule(lr){2-3} \cmidrule(lr){4-5} \cmidrule(lr){6-7} \cmidrule(lr){8-9} \cmidrule(lr){10-11}
Method & AUROC & ECE & AUROC & ECE & AUROC & ECE & AUROC & ECE & AUROC & ECE \\
\midrule
Self-Consistency & 0.5927 & 0.3068 & 0.4512 & 0.6203 & \textbf{0.5630} & 0.2723 & 0.6334 & 0.2459 & 0.7234 & \textbf{0.0888} \\
UMPIRE & 0.5825 & 0.4507 & 0.6318 & 0.7075 & 0.5551 & 0.3999 & 0.6159 & 0.1307 & 0.5272 & 0.5649 \\
RDS & 0.6038 & 0.2405 & 0.6654 & 0.2760 & 0.5434 & \textbf{0.1382} & 0.5894 & 0.1951 & 0.6169 & 0.3526 \\
Semantic Volume & 0.6017 & \textbf{0.2211} & \textbf{0.6671} & 0.4689 & 0.5417 & 0.1802 & 0.5918 & \textbf{0.1223} & 0.6062 & 0.1130 \\
\midrule
Level-Based Conf. & 0.5230 & 0.3192 & 0.5074 & 0.1661 & 0.5154 & 0.3466 & 0.5124 & 0.5777 & 0.5566 & 0.1866 \\
Self-Probing Conf. & 0.5803 & 0.3170 & 0.5705 & \textbf{0.1594} & 0.5495 & 0.3378 & 0.5871 & 0.5688 & 0.6143 & 0.2019 \\
Two-Step CoT Conf. & 0.5435 & 0.3147 & 0.5441 & 0.1611 & 0.5109 & 0.3473 & 0.5411 & 0.5522 & 0.5779 & 0.1981 \\
\midrule
$S_{\text{smooth}}$ & 0.6091 & 0.2970 & 0.4551 & 0.6124 & 0.5621 & 0.2758 & \textbf{0.6880} & 0.2071 & 0.7311 & 0.0928 \\
$S_{\text{dens}}$ & 0.5923 & 0.3057 & 0.4355 & 0.6174 & 0.5405 & 0.2860 & 0.6563 & 0.2271 & \textbf{0.7368} & 0.0923 \\
$S_{\text{goal}}$ & 0.5948 & 0.3077 & 0.4516 & 0.6141 & 0.5460 & 0.2895 & 0.6462 & 0.2302 & 0.7356 & 0.0969 \\
$G_{\text{avg}}$ & \textbf{0.6102} & 0.2998 & 0.4817 & 0.6032 & 0.5576 & 0.2816 & 0.6705 & 0.2183 & 0.7311 & 0.0961 \\
$G_{\max}$ & 0.5918 & 0.3043 & 0.4351 & 0.6148 & 0.5582 & 0.2718 & 0.6507 & 0.2355 & 0.7232 & 0.0952 \\
\midrule
$\mathcal{C}_{\text{chain}}$ & 0.6046 & 0.2999 & 0.4700 & 0.6068 & 0.5622 & 0.2752 & 0.6590 & 0.2236 & 0.7272 & 0.0939 \\
\bottomrule
\end{tabular}
\end{adjustbox}
\caption{AUROC ($\uparrow$) and ECE ($\downarrow$) across the UNO-Bench splits with Gemini-generated answers (means over 100 randomised runs). Avg is the unweighted mean of the four per-split values; results on the pooled union of the splits are deferred to Appendix~\ref{app:overall}, which explains why pooled AUROC is unreliable under split heterogeneity. Bold indicates the best result per column.
}
\label{tab:sf-gemini}
\end{table*}

\begin{table*}[t]
\centering
\begin{adjustbox}{width=\textwidth}
\begin{tabular}{l cc cc cc cc cc}
\toprule
& \multicolumn{2}{c}{Avg} & \multicolumn{2}{c}{Audio} & \multicolumn{2}{c}{MC} & \multicolumn{2}{c}{MO} & \multicolumn{2}{c}{Visual} \\
\cmidrule(lr){2-3} \cmidrule(lr){4-5} \cmidrule(lr){6-7} \cmidrule(lr){8-9} \cmidrule(lr){10-11}
Method & AUROC & ECE & AUROC & ECE & AUROC & ECE & AUROC & ECE & AUROC & ECE \\
\midrule
Self-Consistency & 0.6559 & 0.2553 & 0.7236 & 0.1310 & 0.5461 & 0.3198 & 0.6782 & 0.4051 & 0.6755 & 0.1653 \\
UMPIRE & 0.6570 & 0.1577 & \textbf{0.7507} & 0.1550 & 0.5840 & 0.2898 & 0.5875 & \textbf{0.1323} & \textbf{0.7058} & \textbf{0.0538} \\
RDS & 0.5764 & \textbf{0.1401} & 0.7117 & \textbf{0.1127} & 0.4976 & 0.1664 & 0.5004 & 0.2133 & 0.5958 & 0.0680 \\
Semantic Volume & 0.5882 & 0.2027 & 0.7169 & 0.2858 & 0.5137 & \textbf{0.1142} & 0.5135 & 0.1770 & 0.6088 & 0.2338 \\
\midrule
Level-Based Conf. & 0.6200 & 0.2699 & 0.6306 & 0.1796 & 0.6024 & 0.3905 & 0.5956 & 0.2745 & 0.6512 & 0.2350 \\
Self-Probing Conf. & 0.6563 & 0.2843 & 0.6941 & 0.1753 & \textbf{0.6331} & 0.4658 & 0.5933 & 0.2521 & 0.7047 & 0.2440 \\
Two-Step CoT Conf. & 0.6294 & 0.3133 & 0.6791 & 0.1830 & 0.6252 & 0.4804 & 0.5609 & 0.3028 & 0.6525 & 0.2871 \\
\midrule
$S_{\text{smooth}}$ & 0.6564 & 0.2628 & 0.7270 & 0.1406 & 0.5628 & 0.3116 & 0.6531 & 0.4357 & 0.6829 & 0.1632 \\
$S_{\text{dens}}$ & 0.6538 & 0.2590 & 0.7344 & 0.1293 & 0.5572 & 0.3141 & 0.6347 & 0.4317 & 0.6887 & 0.1610 \\
$S_{\text{goal}}$ & 0.6522 & 0.2627 & 0.7295 & 0.1354 & 0.5451 & 0.3155 & 0.6462 & 0.4408 & 0.6878 & 0.1593 \\
$G_{\text{avg}}$ & 0.6588 & 0.2649 & 0.7419 & 0.1346 & 0.5507 & 0.3179 & 0.6675 & 0.4382 & 0.6750 & 0.1690 \\
$G_{\max}$ & 0.6584 & 0.2564 & 0.7223 & 0.1332 & 0.5439 & 0.3225 & \textbf{0.6978} & 0.4013 & 0.6695 & 0.1686 \\
\midrule
$\mathcal{C}_{\text{chain}}$ & \textbf{0.6603} & 0.2573 & 0.7368 & 0.1330 & 0.5507 & 0.3178 & 0.6700 & 0.4161 & 0.6835 & 0.1623 \\
\bottomrule
\end{tabular}
\end{adjustbox}
\caption{AUROC ($\uparrow$) and ECE ($\downarrow$) across the UNO-Bench splits with MiniCPM-generated answers.
Avg and bold as in Table~\ref{tab:sf-gemini}.}
\label{tab:sf-minicpm}
\end{table*}

\begin{table}[t]
\centering
\begin{adjustbox}{width=0.45\textwidth}
\begin{tabular}{l cc}
\toprule
& Gemini & MiniCPM \\
\cmidrule(lr){2-2} \cmidrule(lr){3-3}
Split & $\Delta$AUROC (\%) & $\Delta$AUROC (\%) \\
\midrule
Audio &
$+4.18$\textsuperscript{***} &
$+1.82$\textsuperscript{***} \\

MC &
$-0.14$ {\scriptsize(n.s.)} &
$+0.85$\textsuperscript{***} \\

MO &
$+4.04$\textsuperscript{***} &
$-1.20$\textsuperscript{**} \\

Visual &
$+0.53$\textsuperscript{***} &
$+1.19$\textsuperscript{***} \\
\midrule
Avg &
$+2.15$ &
$+0.66$ \\
\bottomrule
\end{tabular}
\end{adjustbox}
\caption{Mean paired relative difference in AUROC (\%) of $\mathcal{C}_{\text{chain}}$ against self-consistency across 100 randomised runs. Positive $\Delta$AUROC favours our method. Asterisks denote paired $t$-tests: \textsuperscript{*}$p<0.01$, \textsuperscript{**}$p<0.001$, \textsuperscript{***}$p<0.0001$. Entries marked ``n.s.'' are not significant. Every entry is the deployable aggregator; the corresponding per-split oracle envelope over the individual scores, which is not a deployable method, is reported separately in Table~\ref{tab:delta-oracle}.}
\label{tab:delta-combined}
\end{table}

\subsection{Item-level comparisons}
\label{sec:item-level}

The run-level tests of Table~\ref{tab:delta-combined} quantify robustness to evaluation randomness; we complement them with paired bootstrap deltas over test questions on a canonical partition ($B=10^4$; Appendix~\ref{app:variance} reports the underlying variances and confidence intervals). These comparisons are computed on the pooled union of the splits.
With Gemini-generated answers, $\mathcal{C}_{\text{chain}}$ improves over Self-Consistency on pooled AUROC ($+0.016$, 95\% CI $[+0.006, +0.025]$, $p=0.0009$); with MiniCPM-generated answers it again improves over Self-Consistency (pooled AUROC $+0.014$, $p=0.008$), whereas native UMPIRE holds a significant advantage over it (AUROC $-0.035$, $p=0.002$). Against the strongest verbalised baseline, $S_{\text{dens}}$ gains $+0.108$ AUROC over Self-Probing on Visual with Gemini ($p=0.0001$) and $S_{\text{smooth}}$ gains $+0.122$ AUROC on MO ($p=0.009$); with MiniCPM, Self-Probing holds a significant pooled AUROC advantage ($-0.039$, $p=0.003$) while our scores keep the lead on the open-answer omni-modal split ($S_{\text{goal}}$: $+0.134$ AUROC, $p=0.023$).

\subsection{What the chain geometry adds to answer dispersion}
\label{sec:complementarity}

The comparisons above are marginal: they ask whether a geometric score discriminates better than an answer-dispersion score, not whether it carries information that answer dispersion cannot carry. Two measurements address the second question, on the canonical partition's test items; Appendix~\ref{app:conditional-media} gives the protocols and the full tables.

\paragraph{Discrimination with the agreement signal held constant.} With $N{=}6$ candidates, self-consistency attaches to a question one of at most six vote shares, so inside a stratum of equal share its confidence is constant and its AUROC there is exactly $0.5$: it does not merely fail to rank those questions, it cannot. Table~\ref{tab:conditional} reports the \emph{stratified} AUROC, which compares only questions falling in the same stratum, with the scored answer fixed to the one self-consistency returns so that only the attached confidence differs across methods. Conditioning removes most of the marginal discrimination of every method --- $G_{\text{avg}}$ falls from $0.7319$ to $0.5753$ on Gemini-Visual, $S_{\text{smooth}}$ from $0.7267$ to $0.6009$ on MiniCPM-MO --- so a large part of what the main tables credit to chain geometry is the vote share re-expressed through Equation~\ref{eq:gwsc}. What survives is nonetheless significant, and it survives in \emph{different} places for the two families: the geometric signals keep an interval clear of $0.5$ on Gemini-Visual, Gemini-MC, MiniCPM-Audio and MiniCPM-MO, the dispersion baselines on Gemini-Audio ($0.7512$ and $0.7555$, their strongest result anywhere in this work) and on MiniCPM-Audio and MiniCPM-Visual; the families overlap on one cell in eight. On the omni-modal splits the dispersion baselines retain nothing once agreement is held constant ($0.4752$ and $0.5090$ on MiniCPM-MO, $0.4797$ and $0.4970$ on MiniCPM-MC) while the coherence score retains $0.6009$ on the former. This is the complementarity of Section~\ref{sec:item-level} established by conditioning rather than by correlation, and it also bounds the claim: what the geometry adds beyond agreement is real but small, and concentrated in particular generator--modality combinations.

\paragraph{Grounding reads the question's own media.} The dispersion baselines are functions of the sampled answers alone, so they are invariant to which media a question carries and the following control cannot even be applied to them. We re-evaluate grounding after re-joining every question to \emph{another} question's media, permuting within blocks of equal split and modality signature so that the permuted score uses the same number of vectors of the same modalities, on the same split, with the same candidate answers and the same encoder; only the identity of the media changes. Over $200$ derangements the grounding score discriminates better on the true join in eight of the eight generator--split cells, by $+0.027$ AUROC on the split average with either generator (permutation $p=0.005$ and $p=0.015$). Applied to the \emph{deployed} confidence rather than to the score, the same control returns only $+0.003$ to $+0.007$, inside the permutation band: Equation~\ref{eq:gwsc} normalises the candidates' scores within a question, so a factor common to all of that question's candidates cancels and a common additive shift pulls the share towards the plain vote frequency. What the deployed estimator reads is the within-question contrast, not how well the question is grounded at all --- which locates a concrete piece of headroom, consistent with the wider-pool result of Appendix~\ref{app:aggregator-evidence}: a question-level grounding term should enter alongside the share rather than through it. Two bounds on the reading: question-level grounding discriminates weakly in absolute terms ($0.517$--$0.541$ split-averaged), and the control is available only for the answer-level grounding of Equation~\ref{eq:ansground}, since per-step embeddings are not retained.

\begin{table*}[t]
\centering
\begin{adjustbox}{width=\textwidth}
\begin{tabular}{l ccccc ccccc}
\toprule
& \multicolumn{5}{c}{Gemini} & \multicolumn{5}{c}{MiniCPM} \\
\cmidrule(lr){2-6} \cmidrule(lr){7-11}
Split & $S_{\text{smooth}}$ & $G_{\text{avg}}$ & $G^{\min}_{\text{ans}}$ & RDS & Sem.\ Vol. & $S_{\text{smooth}}$ & $G_{\text{avg}}$ & $G^{\min}_{\text{ans}}$ & RDS & Sem.\ Vol. \\
\midrule
Audio & 0.5152 & 0.4885 & 0.5514 & 0.7512\textsuperscript{\dag} & 0.7555\textsuperscript{\dag} & 0.5167 & 0.5756\textsuperscript{\dag} & 0.5754\textsuperscript{\dag} & 0.6391\textsuperscript{\dag} & 0.6441\textsuperscript{\dag} \\
MC & 0.5372 & 0.5298 & 0.5541\textsuperscript{\dag} & 0.5231 & 0.5249 & 0.5198 & 0.5030 & 0.4732 & 0.4797 & 0.4970 \\
MO & 0.5189 & 0.5385 & 0.5718 & 0.5131 & 0.5128 & 0.6009\textsuperscript{\dag} & 0.5169 & 0.5135 & 0.4752 & 0.5090 \\
Visual & 0.5604\textsuperscript{\dag} & 0.5753\textsuperscript{\dag} & 0.4685 & 0.5328 & 0.5304 & 0.5201 & 0.5173 & 0.4871 & 0.5728\textsuperscript{\dag} & 0.5860\textsuperscript{\dag} \\
\midrule
Avg & 0.5329 & 0.5330 & 0.5364 & 0.5800 & 0.5809 & 0.5394 & 0.5282 & 0.5123 & 0.5417 & 0.5590 \\
\bottomrule
\end{tabular}
\end{adjustbox}
\caption{Discrimination with the agreement signal held constant: AUROC restricted to pairs of questions sharing a majority-vote share (canonical partition, test items; the scored answer is the one self-consistency returns, identical for every method). Self-consistency scores exactly $0.5$ in every cell by construction and is omitted. \textsuperscript{\dag} marks a cell whose $95\%$ question-level bootstrap interval ($B=2{,}000$) excludes $0.5$; the bootstrap resolves $p$ no finer than $5\times10^{-4}$, so we mark intervals rather than use the asterisk convention of the main tables. Avg is the unweighted mean of the four per-split values. Appendix~\ref{app:conditional-media} reports all eleven signals with their unconditional AUROCs.}
\label{tab:conditional}
\end{table*}

\subsection{Discussion}

Several patterns emerge. First, the embedding scores are competitive with the agreement-based methods. With Gemini, $G_{\text{avg}}$ attains the best average AUROC in Table~\ref{tab:sf-gemini} (0.6102), and the paired analyses show that $\mathcal{C}_{\text{chain}}$'s gains over Self-Consistency are systematic on three of the four splits with Gemini (average $+2.15\%$, Table~\ref{tab:delta-combined}) and hold at the item level under both generators (Section~\ref{sec:item-level}). With MiniCPM the gains are smaller but significant on three of the four splits (Audio $+1.82\%$, MC $+0.85\%$, Visual $+1.19\%$, average $+0.66\%$); the open-answer omni-modal split remains a loss ($-1.20\%$), so we do not claim a uniform improvement across generators.

Second, the dispersion baselines are strong but their strength is concentrated where the agreement signal fails: both repair the Audio failure mode that inverts every agreement-based signal (AUROC 0.6654--0.6671 against 0.4512 for Self-Consistency), yet on average AUROC they trail the best embedding scores (0.6017--0.6038 vs.\ 0.6102 for $G_{\text{avg}}$). 
The two families are complementary rather than redundant, consistently across generators: the dispersion baselines weaken exactly where the chain scores are strongest, which lead under Gemini on Visual (0.7368 vs.\ 0.6169), MC (0.5622 vs.\ 0.5434) and MO (0.6880 vs.\ 0.5918), a gap widening to chance level under MiniCPM (0.6978 for $G_{\max}$ vs.\ 0.5135 on MO). Dispersion excels where answer diversity is informative, chain geometry where an answer must be assessed against multimodal evidence; their hybrid removes every below-chance split, at some cost in variance and calibration (Appendix~\ref{app:additional-baselines}).

Third, the self-verbalisation comparison favours our scores on discrimination and the baselines on calibration. With Gemini every embedding score improves on the best verbalised AUROC on average (0.5918--0.6102 vs.\ 0.5803); the verbalised calibration advantage largely dissolves under deployment-realistic Platt scaling (Appendix~\ref{app:variance}). The families also differ in cost: verbalised confidence re-invokes the evaluated model, ours needs only the external encoder (Appendix~\ref{app:experiments}). 

Fourth, the omni-modal open-answer split confirms the structural weakness our method targets: all baselines fall below 0.64 AUROC on MO under Gemini, whereas the embedding scores reach 0.6880. UMPIRE's MO weakness persists in its native MiniCPM configuration (0.5875), so the difficulty of handling more than one extra-textual modality appears structural rather than an artifact of the proxy setting; elsewhere native UMPIRE leads on the uni-modal splits and is best calibrated. MO is the smallest split (200 test items), so its margins carry the widest item-level intervals (Appendix~\ref{app:variance}).

\paragraph{One quantity, several readings.} Finally, the individual scores are heterogeneous across splits and generators but highly redundant \emph{with one another}: their question-level confidences correlate at $0.906$ (Gemini) and $0.942$ (MiniCPM) within a split, against $0.408$ and $0.229$ with the answer-dispersion baselines, whose least-confident deciles are almost disjoint from theirs (Appendix~\ref{app:aggregator-evidence}). They are therefore several readings of one underlying quantity rather than independent evidence to be fused, and the family they genuinely complement is between-sample dispersion. That single fact explains the rest: a family correlating at $0.91$--$0.94$ offers an aggregator no diversity to exploit, which is why $\mathcal{C}_{\text{chain}}$ collapses to a selected feature, why no combination rule we tested beats selecting one, and why selection only breaks even against fixing one signal a priori ($0.6081$ vs.\ $0.6102$; $0.6624$ vs.\ $0.6627$). The validation step therefore contributes robustness rather than accuracy: its role is to avoid deploying a signal that is inverted on the target generator--modality combination.
Extending the feature pool with the answer-level grounding of Equation~\ref{eq:ansground} ($\mathcal{C}^{+}_{\text{chain}}$, Appendix~\ref{app:aggregator-evidence}) raises the split-averaged gain over Self-Consistency with Gemini from $+2.15\%$ to $+2.62\%$ and makes all four splits significantly positive, the MC split moving from $-0.14\%$ (n.s.) to $+1.40\%$\textsuperscript{***}. The aggregated $\mathcal{C}_{\text{chain}}$ rarely attains the per-split maximum but tracks the best embedding score closely; we present it as a deployable default rather than as the contribution of this work, which is the family of signals and the characterisation of when each is reliable. Appendix~\ref{app:ablations} dissects which features carry its performance and Appendix~\ref{app:transferability} tests how its weights transfer across splits and generators. Appendix~\ref{app:gwsc} shows results obtained on answers selected with our scores instead of pure majority voting, where we also show that our improvements are largest at small sampling budgets (Appendix~\ref{app:n-sweep}).

\section{Conclusions}
Taken together, our results suggest that geometric scores over the reasoning trace, computed with external omni-modal embeddings, provide a practical confidence signal for black-box omni-modal models whose reliability is model- and task-dependent, 
showing systematic gains over self-consistency under the closed generator and mixed, split-dependent gains under the open one, while outperforming the self-verbalisation baselines on discrimination for the closed-model generator while requiring no additional call to the evaluated model. On the omni-modal open-answer split, where most baselines including natively-run UMPIRE degrade sharply, the embedding scores remain the strongest family, indicating that the difficulty of assessing answers spanning several extra-textual modalities is structural for answer-level methods and is mitigated by reasoning-trace geometry. The aggregation experiments also delimit what these scores are not: they are not independent evidence sources whose fusion constitutes the contribution. They correlate at $0.91$--$0.94$ within a split, no combination rule we tested improves on selecting one, and a labelled calibration set is worth roughly what fixing one signal a priori is worth; their genuine complement lies outside the family, in answer dispersion. We therefore read this work's contribution as a family of cheap, interpretable, black-box diagnostics plus an empirical account of when each is trustworthy, rather than a single unified estimator. 
The verbalised baselines retain an advantage on raw calibration (largely a scale artefact under deployment-realistic calibration) and, with MiniCPM-generated answers, the strongest verbalised baseline matches our aggregate on discrimination averaged across splits. One practical caveat deserves emphasis: the Audio split with Gemini-generated answers shows that any confidence signal derived from sampled agreement can invert on a particular generator--modality combination, so per-modality discrimination should be validated on a small labelled sample before any of these signals is deployed as a gating or escalation mechanism.

\section{AI Usage Statement}

During the course of this research, generative AI tools were used as assistive instruments for tasks such as improving code quality, debugging, drafting and refining text, and supporting the implementation and execution of experiments. These tools were also used to facilitate literature exploration and to improve the clarity and presentation of the manuscript.

All scientific contributions, including the formulation of the research questions, the proposed methodology, experimental design, evaluation protocol, interpretation of results, and conclusions, were conceived and made by the authors. AI-generated suggestions, code, or text were critically reviewed, modified where necessary, and verified prior to inclusion. All experiments were executed, validated, and checked by the authors to ensure the correctness and reproducibility of the reported results.

The authors take full responsibility for the content of this paper, including its technical accuracy, scientific claims, and ethical compliance.

\section{Ethical Statement}

This work proposes a method for improving uncertainty estimation in OLLMs by leveraging external embedding representations. More reliable uncertainty estimation can help users and downstream systems better identify predictions that are likely to be unreliable, thereby improving calibration, enabling selective prediction, and supporting safer deployment in high-stakes or human-in-the-loop applications.

Our method is model-agnostic and operates on learned embedding representations without modifying the underlying OLLM. As such, it does not introduce new capabilities for generating content or accessing private information beyond those already present in the base model. However, the quality of uncertainty estimates depends on both the underlying model and the external embedding model. Biases, blind spots, or distributional limitations present in either component may propagate into the estimated uncertainty. Consequently, uncertainty scores should not be interpreted as guarantees of correctness or fairness.

Improved uncertainty estimation may also be misused. For example, confidence estimates could be employed to automate decision-making in sensitive domains without appropriate human oversight, potentially leading to over-reliance on model outputs. Additionally, well-calibrated uncertainty could be exploited to selectively deploy models only when they appear confident, masking systematic failures on underrepresented populations or out-of-distribution inputs. We therefore recommend that uncertainty estimates be used as one component of a broader risk assessment framework, alongside domain expertise, robustness evaluations, and fairness analyses.

Our experiments are conducted on publicly available datasets and pretrained models. We do not collect or annotate personal data, nor do we introduce additional datasets containing sensitive information. The proposed method incurs modest additional computational overhead for extracting external embeddings and estimating uncertainty, but it does require an auxiliary embedding model at inference time, which may increase computational cost and energy consumption relative to standard inference.

Overall, we believe that improving uncertainty estimation is a positive step toward more transparent and trustworthy multimodal AI systems. Nevertheless, uncertainty estimation alone cannot address issues such as hallucination, bias, factual inaccuracies, or harmful model behavior. Future work should investigate uncertainty estimation jointly with fairness, robustness, and interpretability to better support the responsible deployment of omni-modal large language models.

\section{Reproducibility Statement}
To make our work as reproducible as possible, we have released our code via an anonymised repository and detailed our software environment, both available in Appendix \ref{app:reproduce}.

Further experimental details about the data used, the pre-processing steps and prompts used, when relevant, are also available in Appendix \ref{app:experiments}.

Finally, more details about our baselines, in order to implement those as well in the form they appear in this work are available in Appendix \ref{app:baselines}.

\bibliography{iclr2027_conference}
\bibliographystyle{iclr2027_conference}

\appendix

\section{Experimental Details}
\label{app:experiments}

\paragraph{UNO-Bench.}
We use the UNO-Bench dataset~\citep{chen2025unobenchunifiedbenchmarkexploring} from Hugging Face
(\url{https://huggingface.co/datasets/meituan-longcat/UNO-Bench}) as the
evaluation benchmark. The dataset contains multiple splits covering different
modalities and question types, and we follow the original paper's split
definitions and evaluation protocol~\citep{chen2025unobenchunifiedbenchmarkexploring}. For each question, correctness
labels are computed according to the procedure described in the original
UNO-Bench paper, which determines whether the generated answer matches
the reference answer under the corresponding split-specific evaluation rules.
The evaluated splits contain 1{,}000 (Audio), 1{,}480 (Visual), 1{,}000 (MC),
and 250 (MO) questions, for a total of 3{,}730.

\paragraph{Answer Generation.}
All answers are generated using Gemini 3.1 Flash Lite (3 March 2026 version)
through the \texttt{google-genai} Python library (v1.66.0) with default
generation parameters. The same model configuration is used consistently for
all experiments and baseline comparisons.

\paragraph{Omni-Modal Encoder}
As mentioned in the main text, we have used E5-Omni \citep{chen-etal-2026-e5} as encoder for extracting external embeddings for all given input modalities. Specifically, we have used the 3B version (\url{https://huggingface.co/Haon-Chen/e5-omni-3B}), which underscores how relatively small encoders are already powerful enough to reach the described results.

\paragraph{Self-Consistency Clustering.}
We follow the standard self-consistency formulation introduced in Semantic
Entropy~\citep{farquhar2024semanticentropy}, where multiple sampled generations are grouped into answer
clusters to estimate agreement between model outputs. This clustering step is
particularly straightforward for the Audio, Visual, and MC splits, where
questions admit discrete or multiple-choice answers, while it is potentially
more challenging for the MO split due to its open-ended nature.

To obtain answer clusters, we use Qwen3.5-9B
(\url{https://huggingface.co/Qwen/Qwen3.5-9B}) as a semantic clustering model.
Given the set of sampled answers, the model is prompted to group semantically
equivalent responses using the instructions shown in Figure~\ref{fig:clustering-prompt}.

\begin{figure*}[t]
\centering
\begin{tcolorbox}[width=0.95\textwidth,
                  colback=white,
                  boxrule=0.5pt,
                  arc=0pt]
\small
\ttfamily
Given the following dictionary of exactly \{num\_answers\} answers extracted from a language model
(keyed by their original index, from "0" to "\{num\_answers - 1\}"), group them based on their
semantic equivalence or having selected the same multiple-choice option.

Return ONLY a valid JSON object with a single key 'groups'. The value of 'groups' must be a
JSON object (dictionary) mapping EVERY index string (from "0" to "\{num\_answers - 1\}") to an
integer group ID. You must include all \{num\_answers\} indices, with no omissions or extras.

Example Input Answers:
\{"0": "Final Answer: A", "1": "The answer is A", "2": "Option B",
"3": "B", "4": "A"\}

Example Output:
\{"groups": \{"0": 0, "1": 0, "2": 1, "3": 1, "4": 0\}\}

Answers to cluster (Total items: \{num\_answers\}):
\end{tcolorbox}
\caption{Prompt used for semantic clustering of sampled answers.}
\label{fig:clustering-prompt}
\end{figure*}

\paragraph{Confidence Classifier.}
The unified confidence score $\mathcal{C}_{\text{chain}}$ is obtained using a logistic
regression classifier implemented with \texttt{scikit-learn}~\citep{scikit-learn}.
The classifier uses the extracted chain-level features as input and is trained
to predict answer correctness, where correct and incorrect answers correspond
to labels 1 and 0, respectively. The features are standardised per split and the model is fitted with an
$\ell_1$ penalty (\texttt{saga} solver, \texttt{max\_iter} 5000); the inverse
regularisation strength is selected by grid search over
$C\in\{0.01, 0.03, 0.1, 0.3, 1, 3, 10\}$, scoring each candidate by validation
AUROC computed through the same cluster-aggregation path used at test time.
The fitted model is then compared against each of the five single-feature
scorers on the same validation items and is replaced by the best single feature
whenever that feature is strictly better; it is this explicit selection step,
rather than the $\ell_1$ penalty alone, that produces the single-feature
unit-weight solutions reported in Appendix~\ref{app:ablations} and
Appendix~\ref{app:transferability}. For each evaluated split, 20\% of the available data is reserved as a
validation set for parameter selection, while all reported test results are
computed on a separate held-out portion. This separation is maintained for both
our proposed methods and all baseline comparisons. Concretely, the validation
sets contain 200 (Audio), 296 (Visual), 200 (MC), and 50 (MO) questions, 746
overall, leaving test sets of 800, 1{,}184, 800, and 200 questions,
respectively (2{,}984 overall). The MO split is thus by far the smallest, and
its per-split estimates carry correspondingly wide item-level confidence
intervals (Appendix~\ref{app:variance}).

\paragraph{Score Preprocessing.}
As the compared confidence measures live on heterogeneous scales (e.g., raw
similarities, verbalized probabilities, entropy-derived scores), each method's
confidence vector is min--max rescaled to $[0,1]$ over the evaluated split
before computing the metrics,
$\tilde{c} = (c - c_{\min})/(c_{\max} - c_{\min})$.
The transformation is strictly monotonic and therefore leaves AUROC and AURAC
unchanged, while placing all methods on a common scale for the calibration
analysis (ECE). The same preprocessing is applied uniformly to our methods and
to all baselines. We note that, since the rescaling constants are computed on
the evaluated split itself, the reported ECE reflects calibration after an
idealized range normalization rather than the raw calibration observable at
deployment; Appendix~\ref{app:variance} therefore also reports a
deployment-realistic variant in which a Platt calibrator is fitted on the
validation items only.

\paragraph{Cost of the Confidence Signal.}
All sampling-based methods, ours included, consume the same $n=6$ generations
per question; in our data a single Gemini generation averages 2{,}590 tokens
(median 2{,}146; prompt with multimodal inputs plus CoT completion), so the
shared budget is roughly 15{,}500 evaluated-model tokens per question at
metered API rates. The methods differ in the \emph{marginal} cost of the
confidence signal on top of this shared budget: zero for Self-Consistency;
one forward pass of the 3B external encoder over the inputs and the sampled
chains for our scores ($\approx$5 reasoning steps and an answer per chain,
$\approx$240 text tokens each, plus the audio/visual inputs embedded once),
which runs on a single commodity GPU and involves no further evaluated-model
tokens; one additional call to the evaluated model for the self-verbalisation
baselines (the question, answer, and reasoning are re-submitted, so the extra
call is of the same order as a generation, $\approx$2{,}600 tokens); and a
GPU pass of the proxy model to extract internal states for UMPIRE.

\paragraph{Computational Complexity.}
For a chain of $m$ steps, the grounding scores require $m \times |V|$
similarities against the per-modality input embeddings and the smoothness and
convergence scores $O(m)$ similarities, while the semantic-density score is
the only quadratic component, at $O(m^2)$ pairwise similarities; with the
typical $m \approx 5$ steps per chain observed in our data, the entire
score computation is negligible next to the single forward pass of the 3B
encoder over the inputs, steps, and answer.

\paragraph{Code and Artifact Availability.}
We release the full evaluation stack: feature extraction, aggregator training,
the 100-run evaluation protocol, and the bootstrap significance testing,
together with the extracted per-question feature vectors, the generated
answers and CoT traces, and their correctness labels. The released traces are
essential for exact reproduction, since the closed-source generator is served
through a mutable API and its checkpoints cannot be pinned.

\paragraph{CoT Generation and Step Segmentation Prompts.}
When prompting the LLMs to generate candidate answers and then to divide the CoT into steps, we have used the prompts as in Figures \ref{fig:cot-generation} and \ref{fig:cot-segmentation}, respectively. Step segmentation is performed by Gemini 3.1 Flash for both generators, so the two configurations are segmented identically; it is a single auxiliary pass over the already-sampled generations and involves no further call to the evaluated model.

\paragraph{Joining chains to inputs.}
Generation files and embedding files are not stored in the same order, so every score in this work is joined on the question identifier rather than on position: a reasoning chain is always aligned with the audio, image and video of the question that produced it. Only the cross-modal scores read the media at all; the coherence scores, Self-Consistency, UMPIRE and the verbalised baselines are computed without them, so they are invariant to this choice and we use that invariance as a check on the pipeline.

\paragraph{Confidence aggregation.}
All five chain-level features receive the same question-level confidence: the cluster sum-share of Section~\ref{sec:experimental-setup}, that is, the sum of the per-candidate scores in the returned answer's semantic cluster normalised by the total over candidates. The alternative of grading a method on the raw score of the winning candidate is used only where a score is defined at question level and has no per-candidate variation (UMPIRE) or where the baseline itself produces a single number per answer (the verbalised baselines). Because the geometric scores occupy a narrow dynamic range, the two conventions are far from interchangeable, and applying the sum-share uniformly across the five features is what makes them comparable with one another and with Self-Consistency.

\begin{figure*}
    \centering
\fbox{
\begin{minipage}{0.9\textwidth}
\small
\textbf{FINAL OUTPUT INSTRUCTIONS:}
You will use your internal reasoning process to analyze the inputs first. When generating your final response, you must adhere strictly to the following formatting rules:

\textbf{For multiple-choice questions:} Your final output must contain \textbf{ONLY} the exact letter or number of the correct choice (e.g., A, B, C, 1, 2). Do not include any rationale, restatement of the answer, markdown formatting, or disclaimers in the final output.

\textbf{For open-ended questions:} Provide only the direct, concise answer. Do not include introductory phrases, rationale, or disclaimers.
\end{minipage}
}
\caption{Prompt used for generating the CoT-style answers to the (pre-appended) question as per the UNO-Bench datasets.}
\label{fig:cot-generation}
\end{figure*}

\begin{figure*}
\centering
\fbox{
\begin{minipage}{0.9\textwidth}
\small
You are given the reasoning process and final answer of a model.  

REASONING:
{thought\_text}  

FINAL ANSWER:
{answer\_text}  

Your tasks:  

Parse the reasoning above into clear, logical steps.  

Return a JSON object with:  

"steps": a list of strings, each representing one logical step from the reasoning  
\end{minipage}
}
\caption{Prompt used for dividing the input CoT-style answer into separate reasoning steps.}
\label{fig:cot-segmentation}
\end{figure*}


\section{Baseline Implementation}
\label{app:baselines}

\paragraph{UMPIRE.}
UMPIRE requires access to the weights of an open-source model to extract internal
representations used for confidence-weighted majority voting. We use
MiniCPM-O v4.5~\citep{cui2026minicpmo45realtimefullduplex} and compute the internal states of the Gemini-generated
answers described in the main experiments using the \texttt{transformers}
library (v5.5.4) with standard parameters. We use the length-normalised
variant of the UMPIRE score, converted to a confidence as
$1 - \text{UMPIRE}_{\text{len-norm}}$ (undefined values, which occur when the
internal-state extraction fails for a candidate, are mapped to confidence 0).
In the main Gemini experiments MiniCPM-O thus acts as a proxy model for the
closed generator, while in the MiniCPM experiments (Table~\ref{tab:minicpm-auroc-ece})
UMPIRE runs natively on the generating model's own states.

\paragraph{Level-Based Confidence.}
For the level-based confidence baseline, we use the same Gemini model employed
for answer generation (Gemini 3.1 Flash Lite, 3 March 2026 version) through the
\texttt{google-genai} Python library (v1.66.0) with default parameters.
The model is prompted to assign one of five qualitative confidence levels to its
answer, which are mapped to the equally spaced numerical scores
$\{0, 0.25, 0.5, 0.75, 1\}$ (in our data the second level was never emitted,
so four distinct values occur).

\paragraph{Self-Probing.}
For Self-Probing~\citep{liu-etal-2024-llms-learn-uncertainty}, we use Gemini 3.1 Flash Lite with the same API
configuration and default generation parameters. The model receives the
question-answer pair and estimates the probability that the answer is correct,
following the original prompting procedure.

\paragraph{Verbalized 2-Step CoT.}
For Verbalized 2-Step CoT~\citep{tian-etal-2023-just}, we again use Gemini 3.1 Flash Lite with the
same interface and default parameters. The model is prompted to directly output
the probability of correctness based on the previously generated reasoning
context, without additional explanation.

\section{Variance, Confidence Intervals, and Deployment-Realistic Calibration}
\label{app:variance}

The margins separating the strongest methods in the main tables are small (of the order of 0.01 split-averaged AUROC), so this appendix makes the uncertainty around them explicit. Two distinct sources of variation are relevant, and we report both. First, \emph{evaluation randomness}: the 100 randomised runs of the main protocol vary the train/validation/test partition and the answer-selection tie-breaking, and we report the resulting mean $\pm$ standard deviation. Second, \emph{item-level uncertainty}: on the canonical partition (seed 42) we run a paired bootstrap over test questions ($B=10^4$) and report percentile 95\% confidence intervals. The two views are complementary and are centred differently (the interval is centred on the canonical-partition estimate, which can differ from the 100-run mean by an amount comparable to the run-to-run standard deviation), and the item-level intervals are wider by an order of magnitude: run-to-run variability is small, and generalisation to new questions is the binding source of uncertainty. Table~\ref{tab:variance} reports both quantities for the pooled union of the splits, the only setting for which the bootstrap was run; per-split standard deviations and intervals are included in the released result files. Pooled point estimates carry the heterogeneity caveat of Appendix~\ref{app:overall}, but the run-to-run and item-level \emph{variability} they quantify is representative of the per-split analyses as well.

\begin{table*}[t]
\centering
\begin{adjustbox}{width=\textwidth}
\begin{tabular}{l cc cc cc cc}
\toprule
& \multicolumn{4}{c}{Gemini} & \multicolumn{4}{c}{MiniCPM} \\
\cmidrule(lr){2-5} \cmidrule(lr){6-9}
& \multicolumn{2}{c}{AUROC} & \multicolumn{2}{c}{AURAC} & \multicolumn{2}{c}{AUROC} & \multicolumn{2}{c}{AURAC} \\
\cmidrule(lr){2-3} \cmidrule(lr){4-5} \cmidrule(lr){6-7} \cmidrule(lr){8-9}
Method & mean $\pm$ std & 95\% CI & mean $\pm$ std & 95\% CI & mean $\pm$ std & 95\% CI & mean $\pm$ std & 95\% CI \\
\midrule
Self-Consistency & 0.5991 $\pm$ 0.0047 & [0.5760, 0.6178] & 0.8059 $\pm$ 0.0053 & [0.7841, 0.8208] & 0.6380 $\pm$ 0.0041 & [0.6214, 0.6600] & 0.6900 $\pm$ 0.0141 & [0.6773, 0.7210] \\
UMPIRE & 0.5958 $\pm$ 0.0049 & [0.5826, 0.6271] & 0.7982 $\pm$ 0.0038 & [0.7846, 0.8209] & 0.6827 $\pm$ 0.0046 & [0.6704, 0.7082] & 0.7395 $\pm$ 0.0047 & [0.7250, 0.7625] \\
\midrule
Level-Based Conf. & 0.5275 $\pm$ 0.0026 & [0.5180, 0.5389] & 0.7210 $\pm$ 0.0066 & [0.7247, 0.7691] & 0.6425 $\pm$ 0.0037 & [0.6199, 0.6541] & 0.6263 $\pm$ 0.0155 & [0.6525, 0.6997] \\
Self-Probing Conf. & 0.5838 $\pm$ 0.0042 & [0.5682, 0.6009] & 0.7485 $\pm$ 0.0064 & [0.7541, 0.7962] & 0.6978 $\pm$ 0.0047 & [0.6746, 0.7121] & 0.7354 $\pm$ 0.0061 & [0.7111, 0.7525] \\
Two-Step CoT Conf. & 0.5452 $\pm$ 0.0034 & [0.5363, 0.5634] & 0.7290 $\pm$ 0.0065 & [0.7372, 0.7802] & 0.6679 $\pm$ 0.0043 & [0.6423, 0.6805] & 0.7081 $\pm$ 0.0065 & [0.6768, 0.7228] \\
\midrule
$S_{\text{smooth}}$ & 0.6096 $\pm$ 0.0042 & [0.5838, 0.6263] & 0.8127 $\pm$ 0.0042 & [0.7939, 0.8278] & 0.6516 $\pm$ 0.0039 & [0.6335, 0.6727] & 0.7279 $\pm$ 0.0038 & [0.7100, 0.7490] \\
$S_{\text{dens}}$ & 0.5988 $\pm$ 0.0043 & [0.5705, 0.6137] & 0.8097 $\pm$ 0.0043 & [0.7907, 0.8258] & 0.6549 $\pm$ 0.0040 & [0.6368, 0.6757] & 0.7272 $\pm$ 0.0040 & [0.7089, 0.7482] \\
$S_{\text{goal}}$ & 0.6009 $\pm$ 0.0041 & [0.5747, 0.6172] & 0.8147 $\pm$ 0.0042 & [0.7946, 0.8285] & 0.6489 $\pm$ 0.0039 & [0.6306, 0.6699] & 0.7192 $\pm$ 0.0042 & [0.7008, 0.7414] \\
$G_{\text{avg}}$ & 0.6123 $\pm$ 0.0042 & [0.5869, 0.6292] & 0.8139 $\pm$ 0.0042 & [0.7941, 0.8278] & 0.6432 $\pm$ 0.0040 & [0.6250, 0.6644] & 0.7227 $\pm$ 0.0039 & [0.7013, 0.7418] \\
$G_{\max}$ & 0.5935 $\pm$ 0.0048 & [0.5689, 0.6112] & 0.8047 $\pm$ 0.0041 & [0.7846, 0.8204] & 0.6369 $\pm$ 0.0041 & [0.6191, 0.6596] & 0.7012 $\pm$ 0.0043 & [0.6774, 0.7223] \\
\midrule
$\mathcal{C}_{\text{chain}}$ & 0.6081 $\pm$ 0.0110 & [0.5919, 0.6340] & 0.8092 $\pm$ 0.0116 & [0.7952, 0.8297] & 0.6446 $\pm$ 0.0132 & [0.6212, 0.6619] & 0.7048 $\pm$ 0.0299 & [0.6669, 0.7129] \\
\bottomrule
\end{tabular}
\end{adjustbox}
\caption{AUROC and AURAC pooled over the union of splits (subject to the caveat of Appendix~\ref{app:overall}) with both uncertainty views: mean $\pm$ sample standard deviation across the 100 randomised evaluation runs, and 95\% percentile confidence intervals from a paired item-level bootstrap ($B=10^4$) on the canonical partition. The intervals are centred on the canonical-partition estimate rather than the 100-run mean.}
\label{tab:variance}
\end{table*}

Because the point estimates alone cannot settle comparisons at this scale, the tests reported in Section~\ref{sec:item-level} operate on the paired per-question deltas rather than the marginal intervals: two methods evaluated on the same questions have strongly correlated errors, so a paired delta can be significant even when the marginal intervals overlap, as is the case for the comparison of $\mathcal{C}_{\text{chain}}$ against Self-Consistency.

\paragraph{Deployment-realistic calibration.}
The ECE values in the main tables are computed after a per-split min--max rescaling, which is rank-preserving but reflects a normalisation a practitioner could not compute at deployment. As a deployment-realistic alternative, we fit Platt scaling (a one-dimensional logistic calibrator) for every method on the \emph{validation} items only, per split, and evaluate ECE on the calibrated test scores. Since Platt scaling is monotone, AUROC and AURAC are unaffected; only calibration changes. Table~\ref{tab:calibrated-ece} reports results pooled over the union of splits (the calibrators themselves are fitted per split).

\begin{table}[t]
\centering
\begin{adjustbox}{width=\columnwidth}
\begin{tabular}{l cc cc}
\toprule
& \multicolumn{2}{c}{Gemini} & \multicolumn{2}{c}{MiniCPM} \\
\cmidrule(lr){2-3} \cmidrule(lr){4-5}
Method & min--max & calibrated & min--max & calibrated \\
\midrule
Self-Consistency & 0.2924 & 0.0592 & 0.2020 & 0.0573 \\
UMPIRE & 0.6319 & 0.0129 & 0.0553 & 0.0428 \\
\midrule
Level-Based Conf. & 0.2506 & 0.0090 & 0.2573 & 0.0257 \\
Self-Probing Conf. & 0.2515 & \textbf{0.0070} & 0.2967 & 0.0684 \\
Two-Step CoT Conf. & 0.2533 & 0.0093 & 0.3118 & 0.0336 \\
\midrule
$S_{\text{smooth}}$ & 0.2888 & 0.0598 & 0.2021 & 0.0590 \\
$S_{\text{dens}}$ & 0.2933 & 0.0647 & 0.2003 & 0.0610 \\
$S_{\text{goal}}$ & 0.2950 & 0.0604 & 0.2003 & 0.0520 \\
$G_{\text{avg}}$ & 0.2894 & 0.0599 & 0.2028 & 0.0541 \\
$G_{\max}$ & 0.2899 & 0.0583 & 0.1541 & \textbf{0.0163} \\
\midrule
$\mathcal{C}_{\text{chain}}$ & 0.2869 & 0.0720 & 0.2033 & 0.0556 \\
\bottomrule
\end{tabular}
\end{adjustbox}
\caption{ECE ($\downarrow$) pooled over the union of splits on the canonical partition: the split-rescaled protocol of the main tables (min--max) against ECE after Platt scaling fitted on the validation split only (calibrated). Bold marks the best calibrated value per generator.}
\label{tab:calibrated-ece}
\end{table}

Three points follow. First, post-hoc calibration collapses ECE for every method, from the 0.15--0.63 range down to 0.007--0.07: deployment-realistic calibration is cheaply achievable across the board, and most of the large ECE differences in the main tables are scale artefacts of the raw scores rather than genuine calibration properties. Second, and in fairness to the strongest baseline, UMPIRE's severe miscalibration in the Gemini setting (0.6319) disappears entirely after calibration (0.0129); its raw ECE is an orientation artefact and should not be read as a substantive weakness. Third, the verbalised baselines remain the best-calibrated methods on Gemini even after calibration (0.007--0.009 against roughly 0.06 for the embedding-based and frequency-based scores, whose confidences take few distinct values per question and therefore cannot be calibrated within-value), so the calibration advantage we grant them in the main text survives the protocol change. We caution that a low calibrated ECE does not imply discrimination: a near-constant calibrator predicting the base rate obtains excellent ECE with chance-level AUROC, so calibrated ECE must be read jointly with the discrimination metrics.

\paragraph{Validation-fitted Platt slopes as an orientation diagnostic.}
The slope of the validation-fitted Platt calibrator provides a direct sign test of each confidence signal: a positive slope means higher confidence predicts correctness on the validation items, a negative slope means the signal is inverted. Table~\ref{tab:platt-slopes-audio} reports the slopes on the Audio split, where the main tables show the near- or below-chance behaviour with Gemini-generated answers. With Gemini answers, the slope is negative not only for most chain-level scores but also for the encoder-independent Self-Consistency ($-0.67$) and for Two-Step CoT: on this generator--modality combination, agreement between sampled answers and geometric regularity of the chain are \emph{anti-correlated} with correctness, which locates the failure in Gemini's answer distribution on audio questions rather than in the encoder or the chain-level signals as such. Consistently, the same features yield uniformly positive slopes ($+2.5$ to $+2.9$) and high AUROC with MiniCPM-generated answers, so the inversion is a property of one generator--modality combination rather than of the scores.

\begin{table}[t]
\centering
\begin{adjustbox}{width=0.85\columnwidth}
\begin{tabular}{l rrr}
\toprule
& \multicolumn{2}{c}{Gemini} & \multicolumn{1}{c}{MiniCPM} \\
\cmidrule(lr){2-3} \cmidrule(lr){4-4}
Method & E5 & LCO & E5 \\
\midrule
Self-Consistency & $-0.56$ & $-0.56$ & $+2.65$ \\
UMPIRE & $-0.19$ & $-0.19$ & $+6.05$ \\
\midrule
Level-Based Conf. & $+1.20$ & $+1.20$ & $+1.82$ \\
Self-Probing Conf. & $+9.47$ & $+9.47$ & $+5.31$ \\
Two-Step CoT Conf. & $-1.09$ & $-1.09$ & $+3.98$ \\
\midrule
$S_{\text{smooth}}$ & $-0.07$ & $-0.52$ & $+2.56$ \\
$S_{\text{dens}}$ & $-0.21$ & $-0.25$ & $+2.84$ \\
$S_{\text{goal}}$ & $-0.08$ & $-0.16$ & $+2.78$ \\
$G_{\text{avg}}$ & $+0.12$ & $-0.19$ & $+2.53$ \\
$G_{\max}$ & $-0.24$ & $-0.22$ & $+2.65$ \\
\midrule
$\mathcal{C}_{\text{chain}}$ & $+0.12$ & $-0.19$ & $+2.54$ \\
\bottomrule
\end{tabular}
\end{adjustbox}
\caption{Validation-fitted Platt slopes on the Audio split (200 validation items per configuration). Negative slopes indicate an inverted confidence signal. The verbalised baselines are computed once per generator and are therefore identical across the two encoder columns.}
\label{tab:platt-slopes-audio}
\end{table}

\section{Pooled (Overall) Results, and Why Pooled AUROC Is Not a Reliable Summary}
\label{app:overall}

A natural way to summarise performance across the four UNO-Bench splits is to pool them into a single evaluation set of 3{,}730 questions (2{,}984 test items after the validation cut) and compute AUROC, ECE, and AURAC on the union. We report those pooled results here for completeness (Table~\ref{tab:pooled-overall} for the main protocol, Table~\ref{tab:pooled-deltas} for the paired deltas against Self-Consistency), but we do not use them as a headline statistic: under split heterogeneity a pooled AUROC confounds within-split discrimination with the between-split ranking of difficulty, for the reasons set out below. The unweighted per-split average (Avg) is used throughout the paper instead.

\begin{table*}[t]
\centering
\begin{adjustbox}{width=0.9\textwidth}
\begin{tabular}{l ccc ccc}
\toprule
& \multicolumn{3}{c}{Gemini} & \multicolumn{3}{c}{MiniCPM} \\
\cmidrule(lr){2-4} \cmidrule(lr){5-7}
Method & AUROC & ECE & AURAC & AUROC & ECE & AURAC \\
\midrule
Self-Consistency & 0.5991 & 0.2893 & 0.8059 & 0.6380 & 0.2023 & 0.6900 \\
UMPIRE & 0.5958 & 0.6263 & 0.7982 & 0.6827 & 0.0543 & \textbf{0.7395} \\
RDS & 0.6360 & 0.2351 & -- & 0.6420 & \textbf{0.0454} & -- \\
Semantic Volume & \textbf{0.6365} & \textbf{0.1696} & \textbf{0.8187} & 0.6542 & 0.1849 & -- \\
\midrule
Level-Based Conf. & 0.5275 & 0.2497 & 0.7210 & 0.6425 & 0.2550 & 0.6263 \\
Self-Probing Conf. & 0.5838 & 0.2513 & 0.7485 & \textbf{0.6978} & 0.2921 & 0.7354 \\
Two-Step CoT Conf. & 0.5452 & 0.2516 & 0.7290 & 0.6679 & 0.3098 & 0.7081 \\
\midrule
$S_{\text{smooth}}$ & 0.6096 & 0.2846 & 0.8127 & 0.6352 & 0.2027 & 0.6973 \\
$S_{\text{dens}}$ & 0.5988 & 0.2879 & 0.8097 & 0.6385 & 0.2009 & 0.6991 \\
$S_{\text{goal}}$ & 0.6009 & 0.2907 & 0.8147 & 0.6302 & 0.2002 & 0.6845 \\
$G_{\text{avg}}$ & 0.6123 & 0.2853 & 0.8139 & 0.6354 & 0.2012 & 0.6951 \\
$G_{\max}$ & 0.5935 & 0.2863 & 0.8047 & 0.4631 & 0.1563 & 0.5711 \\
\midrule
$\mathcal{C}_{\text{chain}}$ & 0.6081 & 0.2835 & 0.8092 & 0.6497 & 0.2003 & 0.7141 \\
\bottomrule
\end{tabular}
\end{adjustbox}
\caption{Pooled (Overall) AUROC ($\uparrow$), ECE ($\downarrow$), and AURAC ($\uparrow$) on the union of the four UNO-Bench splits, under the main protocol of Tables~\ref{tab:sf-gemini} and~\ref{tab:sf-minicpm} (means over 100 randomised runs). Pooled AURAC was not stored for RDS, and for Semantic Volume only in the Gemini setting. Bold marks the best value per column. These numbers are reported for completeness only: as discussed in this appendix, pooled metrics over so heterogeneous a pool are biased summaries, and the per-split values and their average in the main tables should be preferred.}
\label{tab:pooled-overall}
\end{table*}

\begin{table}[t]
\centering
\begin{adjustbox}{width=\columnwidth}
\begin{tabular}{ll rrr}
\toprule
Generator & Protocol & $\Delta$AUROC (\%) & $\Delta$ECE (\%) & $\Delta$AURAC (\%) \\
\midrule
Gemini & main & $+1.51$\textsuperscript{***} & $-2.00$\textsuperscript{***} & $+0.41$\textsuperscript{*} \\
Gemini & weighted & $+0.96$\textsuperscript{***} & $-1.48$\textsuperscript{***} & $+0.33$ {\scriptsize($p{=}0.014$)} \\
MiniCPM & main & $+1.83$\textsuperscript{***} & $-0.97$\textsuperscript{***} & $+3.52$\textsuperscript{***} \\
MiniCPM & weighted & $+2.40$\textsuperscript{***} & $-1.03$\textsuperscript{***} & $+4.24$\textsuperscript{***} \\
\bottomrule
\end{tabular}
\end{adjustbox}
\caption{Pooled (Overall) mean paired relative differences (\%) of $\mathcal{C}_{\text{chain}}$ with respect to Self-Consistency across the 100 randomised runs, for the main (grading) protocol and the geometry-weighted voting protocol of Appendix~\ref{app:gwsc}. Asterisks follow the convention of Table~\ref{tab:delta-combined}. Being paired differences between methods evaluated on the same pool, these deltas partially cancel the shared between-split component discussed below, but they remain pooled statistics.}
\label{tab:pooled-deltas}
\end{table}

\paragraph{Why pooled AUROC is biased on heterogeneous data.}
AUROC is the probability that a randomly drawn correctly-answered question receives a higher confidence than a randomly drawn incorrectly-answered one. When the evaluation set is the union of splits that differ sharply in difficulty and answer format — as UNO-Bench's do, with accuracies of the scored answers ranging from roughly 0.4 on the open-answer omni-modal split to over 0.8 on Visual — the majority of these correct--incorrect pairs are \emph{cross-split} pairs. Such pairs are ranked correctly by any score that merely assigns systematically higher values on the easier splits, regardless of whether it can order correct above incorrect \emph{within} any split. Pooled AUROC therefore decomposes, informally, into a within-split component (the discrimination ability we actually care about) and a between-split component that only measures how well the score's scale happens to track split difficulty. It is a well-known failure mode of AUROC on heterogeneous pools, closely related to Simpson's paradox: a method can post a high pooled AUROC by ``ranking the easiest examples first'' while possessing no deeper confidence-modelling capability, and pooled rankings can even invert the per-split ordering of two methods. The unweighted per-split average used in the main text removes the between-split component by construction, which is why we adopt it as the primary summary; pooled ECE mixes per-split miscalibrations in an analogous scale-dependent way and is treated with the same caution.

\paragraph{Evidence in our own results.}
The bias is not hypothetical; several of our pooled numbers are visibly inflated by it.
First, the dispersion baselines: Semantic Volume holds the best pooled Gemini AUROC in Table~\ref{tab:pooled-overall} (0.6365, against 0.6081 for $\mathcal{C}_{\text{chain}}$), yet on the per-split average it trails the best embedding scores (0.6017 vs.\ 0.6102 for $G_{\text{avg}}$): the dispersion of answer embeddings has a split-dependent scale that happens to align with split difficulty, and much of the pooled lead is between-split ranking rather than within-split skill.
Second, the verbalised baselines with MiniCPM: Self-Probing's pooled AUROC of 0.6978 shrinks to 0.6563 on the split average, below $\mathcal{C}_{\text{chain}}$ (0.6603) — verbalised confidences are naturally higher on splits where the model succeeds more often, a between-split calibration property that pooled AUROC rewards as if it were discrimination.
Third, entire conclusions can flip: in the feature ablations of Appendix~\ref{app:ablations} the largest pooled loss comes from removing $G_{\max}$ and grounding-only beats coherence-only, whereas on the split average the largest loss comes from removing $S_{\text{smooth}}$ and the ordering of the two families reverses; likewise, the attention-style grounding score of Appendix~\ref{app:grounding-attention} appears as the best MiniCPM grounding score under pooling but trails $G_{\text{avg}}$ on the average.

\paragraph{What we keep pooled.}
A few analyses in this paper remain pooled because the underlying artifacts were only computed at that level: the run-to-run variances and item-level bootstrap confidence intervals of Appendix~\ref{app:variance}, the Platt-calibrated ECE comparison reported there, the item-level paired bootstrap deltas of Section~\ref{sec:item-level} and Appendices~\ref{app:aurac} and~\ref{app:gwsc}, and the sampling-budget sweep of Appendix~\ref{app:n-sweep}. These are all either measures of \emph{uncertainty} rather than of ranking quality, or \emph{paired} comparisons between methods evaluated on the identical pool, for which the between-split component is common to both methods and largely cancels in the difference; they are labelled as pooled wherever they appear, and their point estimates inherit the caveat above.

\section{Selective Answering: AURAC Results}
\label{app:aurac}

Following \citet{lau2026uncertainty}, we consider the scenario in which a model decides whether to progressively reject its most uncertain outputs based on the relative confidence and escalate them to a higher-level verifier. We evaluate this scenario with the Area Under the Rejection--Accuracy Curve (AURAC) \citep{farquhar2024semanticentropy}: the accuracy on the retained questions as the least-confident fraction of answers is progressively rejected, averaged over rejection rates (higher is better). AURAC is rank-based, so it is unaffected by the min--max rescaling of the main protocol and by the Platt scaling of Appendix~\ref{app:variance}. Table~\ref{tab:sf-aurac} reports AURAC for both answer generators under the identical protocol as Tables~\ref{tab:sf-gemini} and~\ref{tab:sf-minicpm} (means over 100 randomised runs; every method scores the same majority-selected answer), and Table~\ref{tab:delta-aurac} the corresponding paired differences of $\mathcal{C}_{\text{chain}}$ against Self-Consistency.

\begin{table*}[t]
\centering
\begin{adjustbox}{width=\textwidth}
\begin{tabular}{l ccccc ccccc}
\toprule
& \multicolumn{5}{c}{Gemini} & \multicolumn{5}{c}{MiniCPM} \\
\cmidrule(lr){2-6} \cmidrule(lr){7-11}
Method & Avg & Audio & MC & MO & Visual & Avg & Audio & MC & MO & Visual \\
\midrule
Self-Consistency & 0.7281 & 0.8188 & 0.6842 & 0.5236 & 0.8860 & 0.6865 & 0.8500 & 0.4135 & 0.7149 & 0.7677 \\
UMPIRE & 0.7240 & \textbf{0.8940} & \textbf{0.6920} & 0.5019 & 0.8080 & \textbf{0.6968} & \textbf{0.8816} & 0.4451 & 0.6621 & \textbf{0.7985} \\
\midrule
Level-Based Conf. & 0.6760 & 0.8353 & 0.6388 & 0.4156 & 0.8144 & 0.6573 & 0.7918 & 0.4343 & 0.6652 & 0.7380 \\
Self-Probing Conf. & 0.7042 & 0.8574 & 0.6704 & 0.4537 & 0.8352 & 0.6931 & 0.8311 & \textbf{0.4815} & 0.6733 & 0.7866 \\
Two-Step CoT Conf. & 0.6885 & 0.8486 & 0.6503 & 0.4354 & 0.8198 & 0.6695 & 0.8234 & 0.4692 & 0.6402 & 0.7453 \\
\midrule
$S_{\text{smooth}}$ & 0.7276 & 0.8112 & 0.6843 & 0.5199 & 0.8951 & 0.6914 & 0.8556 & 0.4160 & \textbf{0.7305} & 0.7633 \\
$S_{\text{dens}}$ & 0.7268 & 0.8120 & 0.6752 & \textbf{0.5254} & 0.8945 & 0.6903 & 0.8545 & 0.4151 & 0.7271 & 0.7645 \\
$S_{\text{goal}}$ & \textbf{0.7315} & 0.8165 & 0.6901 & 0.5206 & \textbf{0.8987} & 0.6805 & 0.8365 & 0.4054 & 0.7190 & 0.7613 \\
$G_{\text{avg}}$ & 0.7292 & 0.8120 & 0.6874 & 0.5206 & 0.8969 & 0.6846 & 0.8384 & 0.4129 & 0.7234 & 0.7636 \\
$G_{\max}$ & 0.7268 & 0.8107 & 0.6900 & 0.5179 & 0.8888 & 0.5726 & 0.7097 & 0.3555 & 0.5898 & 0.6352 \\
\midrule
$\mathcal{C}_{\text{chain}}$ & 0.7275 & 0.8125 & 0.6864 & 0.5198 & 0.8914 & 0.6925 & 0.8548 & 0.4184 & 0.7208 & 0.7762 \\
\bottomrule
\end{tabular}
\end{adjustbox}
\caption{AURAC ($\uparrow$) across the UNO-Bench splits for both answer generators (same protocol as Table~\ref{tab:sf-gemini}). Avg is the unweighted mean of the four per-split values (pooled-union results in Appendix~\ref{app:overall}). Bold indicates the best result per column.}
\label{tab:sf-aurac}
\end{table*}

\begin{table}[t]
\centering
\begin{adjustbox}{width=0.9\columnwidth}
\begin{tabular}{l r r}
\toprule
Split & Gemini & MiniCPM \\
\midrule
Audio   & $-0.77$\textsuperscript{***} & $+0.57$\textsuperscript{**} \\
MC      & $+0.32$ {\scriptsize($p{=}0.015$)} & $+1.23$\textsuperscript{**} \\
MO      & $-0.70$\textsuperscript{***} & $+0.84$\textsuperscript{***} \\
Visual  & $+0.62$\textsuperscript{***} & $+1.11$\textsuperscript{***} \\
\midrule
Average & $-0.13$ & $+0.94$ \\
\bottomrule
\end{tabular}
\end{adjustbox}
\caption{Mean paired relative $\Delta$AURAC (\%) of $\mathcal{C}_{\text{chain}}$ with respect to Self-Consistency across the 100 randomised runs, for both answer generators. Positive values favour our method. Asterisks: paired $t$-test, \textsuperscript{*}$p<0.01$, \textsuperscript{**}$p<0.001$, \textsuperscript{***}$p<0.0001$. The Average row is the unweighted mean of the four per-split deltas and carries no significance test; pooled-union deltas are reported in Appendix~\ref{app:overall}.}
\label{tab:delta-aurac}
\end{table}

The AURAC picture mirrors the discrimination results of the main text. With Gemini-generated answers, the chain scores are the strongest family: $S_{\text{goal}}$ attains the best average AURAC in Table~\ref{tab:sf-aurac} (0.7315) together with the best Visual value (0.8987), while $\mathcal{C}_{\text{chain}}$'s run-level deltas against Self-Consistency are mixed across splits (Table~\ref{tab:delta-aurac}): significant gains on Visual and MC are offset by significant losses on Audio and MO, for an average of $-0.13\%$; among the dispersion baselines of Appendix~\ref{app:additional-baselines}, Semantic Volume repairs the Audio inversion (Audio AURAC up to 0.9017), consistent with its AUROC behaviour. With MiniCPM-generated answers the run-level gains of $\mathcal{C}_{\text{chain}}$ over Self-Consistency are positive and significant on all four splits (average $+0.94\%$), while native UMPIRE leads the average column (0.6968, against 0.6925 for $\mathcal{C}_{\text{chain}}$) except on the omni-modal splits, where the chain scores stay ahead (MO: 0.7305 for $S_{\text{smooth}}$ vs.\ 0.6621).

At the item level (paired bootstrap on the canonical partition, $B=10^4$, as in Section~\ref{sec:item-level}; these comparisons are computed on the pooled union of splits and carry the caveat of Appendix~\ref{app:overall}), $\mathcal{C}_{\text{chain}}$ improves over Self-Consistency on pooled AURAC with Gemini ($+0.010$, $p=0.038$) and with MiniCPM ($+0.023$, $p=0.0007$), and improves over Self-Probing, the strongest verbalised baseline, on pooled AURAC with Gemini ($+0.037$, $p=0.0003$).

\section{Feature Ablations for the Aggregated Score}
\label{app:ablations}

To identify which geometric signals carry the discriminative load of $\mathcal{C}_{\text{chain}}$, we refit the aggregator on the canonical partition (Gemini-generated answers, E5-Omni features) under leave-one-feature-out and family-only configurations, keeping the fitting protocol of the main experiments unchanged (validation items only, same regularisation and grid). This analysis is run in the main-text protocol, where the aggregator scores the fixed majority answer. The full-feature configuration reproduces the stored canonical-partition results before any ablation is evaluated; because these are canonical-partition values, they differ slightly from the 100-run means of the main tables. Table~\ref{tab:feature-ablations} reports test AUROC per split, their unweighted average (Avg), and the pooled-union AURAC (retained because per-split AURAC was not stored for the ablation runs; pooled metrics carry the caveat of Appendix~\ref{app:overall}).

\begin{table}[t]
\centering
\begin{adjustbox}{width=\columnwidth}
\begin{tabular}{l cccc cc}
\toprule
& \multicolumn{4}{c}{AUROC per split} & Avg & Pooled \\
\cmidrule(lr){2-5} \cmidrule(lr){6-6} \cmidrule(lr){7-7}
Config & Audio & MC & MO & Visual & AUROC & AURAC \\
\midrule
full & 0.4738 & 0.5654 & 0.6981 & 0.7122 & 0.6124 & 0.8128 \\
\midrule
$-\,S_{\text{smooth}}$ & 0.4738 & 0.5654 & 0.6444 & 0.7122 & 0.5989 & 0.8133 \\
$-\,S_{\text{dens}}$ & 0.4738 & 0.5654 & 0.6981 & 0.7146 & 0.6130 & 0.8157 \\
$-\,S_{\text{goal}}$ & 0.4738 & 0.5654 & 0.6981 & 0.7122 & 0.6124 & 0.8128 \\
$-\,G_{\text{avg}}$ & 0.4456 & 0.5654 & 0.6981 & 0.7122 & 0.6053 & 0.8129 \\
$-\,G_{\max}$ & 0.4738 & 0.5654 & 0.6981 & 0.7292 & 0.6166 & 0.8054 \\
\midrule
coherence only & 0.4456 & 0.5654 & 0.6981 & 0.7292 & 0.6096 & 0.8072 \\
grounding only & 0.4738 & 0.5654 & 0.6569 & 0.7151 & 0.6028 & 0.8162 \\
\bottomrule
\end{tabular}
\end{adjustbox}
\caption{Feature ablations of $\mathcal{C}_{\text{chain}}$ on the canonical partition (Gemini answers, E5-Omni features): leave-one-feature-out ($-$) and single-family configurations, refitted with the published protocol. Avg AUROC is the unweighted mean of the four per-split values; the AURAC column is pooled over the union of splits (Appendix~\ref{app:overall}).}
\label{tab:feature-ablations}
\end{table}

Three observations follow. First, the validation-selected aggregator is effectively \emph{sparse}: per split it reduces to a single feature ($G_{\text{avg}}$ on Audio, $S_{\text{smooth}}$ on MO, $G_{\max}$ on Visual) or, on MC, to a constant scorer whose confidence coincides with the majority-vote share (the embedding features add nothing beyond agreement on that split), which is why many leave-one-out configurations coincide with the full model on most splits. Second, no single feature is critical for the split-averaged AUROC (the spread across configurations is 0.599--0.617) because the features partially substitute for one another; the largest individual losses come from removing $S_{\text{smooth}}$ ($-0.0134$ Avg AUROC, driven by MO) and $G_{\text{avg}}$ ($-0.0071$, driven by Audio), whereas removing $G_{\max}$ slightly \emph{raises} the average ($+0.0042$, via Visual) while costing pooled AURAC ($-0.0074$), and coherence-only edges out grounding-only on the average (0.6096 vs.\ 0.6028) because its MO and Visual gains outweigh the Audio rescue that $G_{\text{avg}}$ provides. We note that under the pooled-union protocol of Appendix~\ref{app:overall} the ordering differs (there the largest loss comes from removing $G_{\max}$ and grounding-only leads), an instance of how pooling can reshuffle conclusions. Third, the two families are jointly useful: coherence carries the omni-modal open-answer split ($S_{\text{smooth}}$; removing it costs $-0.054$ MO AUROC), grounding carries Audio, and Visual is served comparably by several signals, so neither family alone reproduces the full model's profile.

\paragraph{Fixed hyperparameters of $S_{\text{goal}}$.}
The convergence score's weighting $\gamma=0.6$ and its $m/2$ forward/backward split point were fixed a priori and not tuned. A direct sensitivity sweep requires the per-step embeddings, which are not retained in the stored artifacts (only the step texts are), and is therefore left to the code release; we note, however, that these hyperparameters enter the pipeline only through $S_{\text{goal}}$, and on the canonical partition the fitted aggregator assigns $S_{\text{goal}}$ zero marginal contribution (the $-\,S_{\text{goal}}$ row is identical to the full model), so the aggregate results are insensitive to their exact values.

\section{Supervision, Signal Redundancy, and Modality Use}
\label{app:aggregator-evidence}

This appendix quantifies what the aggregated score $\mathcal{C}_{\text{chain}}$ requires, how much its individual signals overlap, and what the grounding score measures. Unless stated otherwise, numbers are computed on the canonical partition (split seed 42, validation fraction 0.2) or over the same 100 randomised runs and the same protocol as the main tables.

\paragraph{Oracle-based results and the headroom they expose.}
Table~\ref{tab:delta-combined} reports the paired deltas of the deployable aggregator; Table~\ref{tab:delta-oracle} reports the per-split \emph{oracle envelope}, the largest 100-run mean relative AUROC delta over $\{S_{\text{smooth}}, S_{\text{goal}}, S_{\text{dens}}, G_{\text{avg}}, G_{\max}, \mathcal{C}_{\text{chain}}\}$ together with the score that attains it. Because the selecting score is chosen on the test split, the envelope is an upper bound rather than an attainable operating point, and we never report it as the performance of our method. It is nonetheless informative, because the gap to it is not irreducible noise. The split-averaged envelope reaches $+4.29\%$ (Gemini) and $+2.61\%$ (MiniCPM) against $+2.15\%$ and $+0.66\%$ for $\mathcal{C}_{\text{chain}}$, and the identity of the winning score is highly stable across resamplings: $G_{\text{avg}}$ wins Gemini-Audio in 100 runs out of 100 and $aG_{\text{mean}}$ wins Gemini-MC in 99. What is unstable is the \emph{validation estimate} of that ordering, not the ordering itself, so a larger calibration sample or a lower-variance selection rule should be expected to recover part of this margin. The consequence is substantive rather than cosmetic: at the envelope the open-answer omni-modal split turns from a loss into a gain under MiniCPM ($-1.20\%$ for $\mathcal{C}_{\text{chain}}$ against $+2.88\%$ for $G_{\max}$), which is the one conclusion in this work that the quality of the selection step decides.

\begin{table}[t]
\centering
\begin{adjustbox}{width=0.7\textwidth}
\begin{tabular}{l cc cc}
\toprule
& \multicolumn{2}{c}{Gemini} & \multicolumn{2}{c}{MiniCPM} \\
\cmidrule(lr){2-3} \cmidrule(lr){4-5}
Split & $\Delta$AUROC (\%) & attained by & $\Delta$AUROC (\%) & attained by \\
\midrule
Audio & $+6.78$\textsuperscript{***} & $G_{\text{avg}}$ & $+2.53$\textsuperscript{***} & $G_{\text{avg}}$ \\
MC & $-0.14$ {\scriptsize(n.s.)} & $\mathcal{C}_{\text{chain}}$ & $+3.06$\textsuperscript{***} & $S_{\text{smooth}}$ \\
MO & $+8.66$\textsuperscript{***} & $S_{\text{smooth}}$ & $+2.88$\textsuperscript{***} & $G_{\max}$ \\
Visual & $+1.85$\textsuperscript{***} & $S_{\text{dens}}$ & $+1.96$\textsuperscript{***} & $S_{\text{dens}}$ \\
\midrule
Avg & $+4.29$ & --- & $+2.61$ & --- \\
\bottomrule
\end{tabular}
\end{adjustbox}
\caption{Per-split \emph{oracle} envelope of the relative AUROC gain over Self-Consistency across the 100 randomised runs: each cell reports the best individual score on that split and names it. The selection is made on the test split, so this is a leakage ceiling and not a deployable method; the deployable aggregator is Table~\ref{tab:delta-combined}. Significance conventions as in Table~\ref{tab:delta-combined}.}
\label{tab:delta-oracle}
\end{table}

\paragraph{How much labelled data does the aggregator need?}
$\mathcal{C}_{\text{chain}}$ is fitted on labelled validation items, which is the one supervised ingredient of the pipeline. We therefore subsample the canonical partition's validation items of each split to $n \in \{25, 50, 100, 200\}$ (20 random draws per $n$), refit with the main-text protocol, and evaluate on the unchanged test items. Table~\ref{tab:calibration-size} reports the resulting test AUROC. Performance is essentially flat in $n$: 25 labelled examples already recover the full-validation result to within the draw-to-draw standard deviation on seven of the eight configurations, and on Gemini-Visual, MiniCPM-Audio and MiniCPM-Visual the small-$n$ fits are in fact slightly \emph{better} than the full-validation fit, because the selection then commits less firmly to a feature that does not transfer to the test items. The practical requirement is therefore a calibration set of a few dozen labelled examples on the target configuration, not a conventional training set.

\begin{table}[t]
\centering
\begin{adjustbox}{width=\columnwidth}
\begin{tabular}{l cccc c}
\toprule
& \multicolumn{4}{c}{Calibration-set size $n$} & \\
\cmidrule(lr){2-5}
Split & 25 & 50 & 100 & 200 & all \\
\midrule
\multicolumn{6}{l}{\emph{Gemini}} \\
Audio & 0.4511 $\pm$ 0.0192 & 0.4498 $\pm$ 0.0198 & 0.4589 $\pm$ 0.0177 & 0.4738 & 0.4738 \\
MC & 0.5611 $\pm$ 0.0126 & 0.5608 $\pm$ 0.0087 & 0.5623 $\pm$ 0.0068 & 0.5654 & 0.5654 \\
MO & 0.6670 $\pm$ 0.0219 & 0.6981 $\pm$ 0.0000 & 0.6981 $\pm$ 0.0000 & 0.6981 & 0.6981 \\
Visual & 0.7207 $\pm$ 0.0072 & 0.7215 $\pm$ 0.0071 & 0.7224 $\pm$ 0.0078 & 0.7197 $\pm$ 0.0076 & 0.7122 \\
\midrule
\multicolumn{6}{l}{\emph{MiniCPM}} \\
Audio & 0.7278 $\pm$ 0.0095 & 0.7315 $\pm$ 0.0096 & 0.7331 $\pm$ 0.0107 & 0.7402 & 0.7402 \\
MC & 0.5532 $\pm$ 0.0055 & 0.5517 $\pm$ 0.0064 & 0.5576 $\pm$ 0.0070 & 0.5622 & 0.5622 \\
MO & 0.6890 $\pm$ 0.0201 & 0.7106 $\pm$ 0.0000 & 0.7106 $\pm$ 0.0000 & 0.7106 & 0.7106 \\
Visual & 0.6927 $\pm$ 0.0044 & 0.6914 $\pm$ 0.0057 & 0.6935 $\pm$ 0.0054 & 0.6936 $\pm$ 0.0023 & 0.6932 \\
\bottomrule
\end{tabular}
\end{adjustbox}
\caption{Test AUROC of $\mathcal{C}_{\text{chain}}$ as a function of the number of labelled calibration examples used to fit it (mean $\pm$ standard deviation over 20 random draws of the calibration subset; entries without a deviation are the cases in which $n$ already exhausts the split's validation items, so a single draw exists). Canonical partition, published fitting protocol.}
\label{tab:calibration-size}
\end{table}

\paragraph{Does the learned combination beat plain feature selection?}
Since the fitted aggregator collapses to a single feature in most fits, the relevant question is whether the multivariate fit adds anything over honestly selecting one feature on the validation set. Table~\ref{tab:selection-ladder} places $\mathcal{C}_{\text{chain}}$ between two references: the best single feature chosen on validation (the honest selection baseline), and an entirely unsupervised uniform combination of the five standardised features. $\mathcal{C}_{\text{chain}}$ coincides with the validation-selected single feature on six of the eight configurations and improves on it only on the two MC splits ($+0.0014$ with Gemini, $+0.0194$ with MiniCPM); it never exceeds the test-selected oracle feature, which it trails by up to $0.0225$ AUROC (MiniCPM-Audio). The unsupervised uniform combination is clearly worse almost everywhere, by up to $0.2371$ AUROC (Gemini-MO). The supervision is therefore doing feature \emph{selection} rather than weight learning, and the contribution of the framework should be read as a family of interpretable geometric signals plus a lightweight, cheap procedure for choosing the reliable one for a given model and modality, rather than as a synergistic multivariate combination.

\begin{table}[t]
\centering
\begin{adjustbox}{width=\textwidth}
\begin{tabular}{l ccccc ccc c}
\toprule
& \multicolumn{5}{c}{Individual scores} & \multicolumn{3}{c}{Deployable} & Oracle \\
\cmidrule(lr){2-6} \cmidrule(lr){7-9} \cmidrule(lr){10-10}
Split & $S_{\text{smooth}}$ & $S_{\text{goal}}$ & $S_{\text{dens}}$ & $G_{\text{avg}}$ & $G_{\max}$ & best single (val.) & $\mathcal{C}_{\text{chain}}$ & uniform & best single (test) \\
\midrule
\multicolumn{10}{l}{\emph{Gemini}} \\
Audio & 0.4456 & 0.4474 & 0.4244 & 0.4738 & 0.4274 & 0.4738 & 0.4738 & 0.4858 & 0.4738 \\
MC & 0.5640 & 0.5491 & 0.5381 & 0.5606 & 0.5615 & 0.5640 & 0.5654 & 0.5334 & 0.5640 \\
MO & 0.6981 & 0.6444 & 0.6569 & 0.6852 & 0.6558 & 0.6981 & 0.6981 & 0.4610 & 0.6981 \\
Visual & 0.7261 & 0.7300 & 0.7292 & 0.7257 & 0.7122 & 0.7122 & 0.7122 & 0.6247 & 0.7300 \\
\midrule
\multicolumn{10}{l}{\emph{MiniCPM}} \\
Audio & 0.7279 & 0.7306 & 0.7345 & 0.7401 & 0.7166 & 0.7401 & 0.7402 & 0.6632 & 0.7401 \\
MC & 0.5622 & 0.5449 & 0.5603 & 0.5510 & 0.5444 & 0.5622 & 0.5622 & 0.5288 & 0.5622 \\
MO & 0.6839 & 0.6638 & 0.6670 & 0.6872 & 0.7106 & 0.7106 & 0.7106 & 0.4836 & 0.7106 \\
Visual & 0.6891 & 0.6961 & 0.6932 & 0.6845 & 0.6801 & 0.6932 & 0.6932 & 0.6437 & 0.6961 \\
\bottomrule
\end{tabular}
\end{adjustbox}
\caption{Test AUROC on the canonical partition of the individual scores, of the best single score selected on the validation items, of $\mathcal{C}_{\text{chain}}$, of an unsupervised uniform combination of the five standardised features, and of the best single score selected on the test items. The last column uses test labels and is a leakage ceiling, not a method.}
\label{tab:selection-ladder}
\end{table}

\paragraph{Is the grounding score using more than one modality?}
Definition~\ref{def:grounding} aggregates the per-modality alignments with a maximum, which rewards a step aligned with a single modality and never requires simultaneous support from several. We test the consequences directly on the splits where more than one modality is present. Of the 3{,}730 questions, only MC (1{,}000) and MO (250) carry two modalities throughout; Audio is single-modality by construction and 1{,}429 of the 1{,}480 Visual questions carry only an image or a video, so on those splits every aggregation rule coincides with the maximum by construction and the comparison is uninformative. Table~\ref{tab:grounding-variants} therefore reports the MC and MO columns, evaluated under the main-text protocol over the same 100 runs on a scoring pass that retains the per-modality alignment profile. That pass segments the chains as elsewhere in this work but treats the final step as the answer unit, so its max-aggregation is a close variant of $G_{\text{avg}}$ rather than numerically identical to it; every rule in the table is therefore compared within that pass.

Two results follow. First, each individual modality stream is separately predictive --- the single-stream scores reach 0.5517--0.5618 AUROC on MC and 0.6717--0.6751 on MO, against 0.5530 and 0.6719 for the max --- so the grounding signal is not an artefact of one dominant modality. Second, requiring \emph{simultaneous} support is mildly but significantly better than the maximum: taking the minimum over modalities, the strictest form of joint grounding, improves MC AUROC from 0.5530 to 0.5585 ($p<10^{-4}$) and the split-pooled value from 0.6008 to 0.6044, while the softer synthesis rules (geometric mean, variance-penalised, entropy-weighted aggregation) sit within $\pm 0.003$ of the maximum. The effect sizes are small, and we report them as evidence that the max-aggregation of Definition~\ref{def:grounding} does under-use cross-modal agreement rather than as a replacement for it. A stronger reading is not warranted: a gain of $0.003$--$0.006$ AUROC from requiring joint support shows that cross-modal agreement carries \emph{some} additional signal, not that these scores measure multi-modal synthesis in any deeper sense. Establishing the latter would require a benchmark in which the answer is recoverable only by combining modalities, which UNO-Bench does not isolate.

\begin{table}[t]
\centering
\begin{adjustbox}{width=0.8\columnwidth}
\begin{tabular}{l cc cc}
\toprule
& \multicolumn{2}{c}{MC} & \multicolumn{2}{c}{MO} \\
\cmidrule(lr){2-3} \cmidrule(lr){4-5}
Grounding aggregation & AUROC & AURAC & AUROC & AURAC \\
\midrule
Maximum over modalities (Def.~\ref{def:grounding}) & 0.5530 & 0.6741 & 0.6719 & 0.5175 \\
Mean over modalities & 0.5524 & 0.6743 & 0.6693 & 0.5164 \\
Geometric mean over modalities & 0.5523 & 0.6744 & 0.6698 & 0.5166 \\
Variance-penalised & 0.5524 & 0.6742 & 0.6699 & 0.5166 \\
Entropy-weighted & 0.5550 & 0.6746 & 0.6711 & 0.5173 \\
\textbf{Minimum over modalities} & \textbf{0.5585} & 0.6745 & \textbf{0.6726} & 0.5168 \\
\midrule
Single stream (first) & 0.5618 & 0.6747 & 0.6751 & 0.5173 \\
Single stream (second) & 0.5517 & 0.6762 & 0.6717 & 0.5172 \\
\bottomrule
\end{tabular}
\end{adjustbox}
\caption{Grounding aggregation rules on the two genuinely bi-modal splits (100-run means, main-text protocol, Gemini answers). The two single-stream rows are the per-modality alignments taken alone; we do not label them audio and visual because the stored ordering of the per-modality alignments could not be verified against the stored values, the between-stream differences being of the order of $0.001$--$0.005$.}
\label{tab:grounding-variants}
\end{table}

\paragraph{Short chains and answer ties.}
Two protocol quantities referenced in Section~\ref{sec:methodology} and Section~\ref{sec:experimental-setup} are reported here. Single-step chains, for which $S_{\text{smooth}}$, $S_{\text{dens}}$ and $\Delta_{\text{bwd}}$ are undefined and our implementation assigns the maximal value, number 17 of 24{,}016 candidate chains ($0.07\%$) with Gemini but 6{,}151 of 37{,}300 ($16.5\%$) with MiniCPM, distributed as 265/10{,}000 on Audio, 2{,}114/10{,}000 on MC, 501/2{,}500 on MO and 3{,}271/14{,}800 on Visual. The MiniCPM coherence scores should therefore be read with the caveat that a sixth of its candidate chains receive maximal coherence by convention. Questions whose count-majority cluster is not unique, and whose returned answer is consequently decided by each method's own tie-break, number 1{,}079 of 3{,}730 ($28.9\%$) with Gemini (601/1{,}000 Audio, 243/1{,}000 MC, 139/250 MO, 96/1{,}480 Visual) and 529 of 3{,}730 ($14.2\%$) with MiniCPM (123/1{,}000 Audio, 104/1{,}000 MC, 140/250 MO, 162/1{,}480 Visual).

\paragraph{Answer-level grounding, formally.}
For a candidate generation $g$ with embedding $\phi_{g}=\mathcal{E}(g)$ and the modality alignment $A^{(k)}(g)=\max_{\phi_v \in \Phi^{(k)}} \mathcal{S}(\phi_{g}, \phi_v)$ of Equation~\ref{eq:align}, the three answer-level grounding scores used in this appendix are
\begin{equation}
G^{\max}_{\text{ans}} = \max_{k} A^{(k)}(g), \quad
G^{\text{avg}}_{\text{ans}} = \frac{1}{K}\sum_{k} A^{(k)}(g), \quad
G^{\min}_{\text{ans}} = \min_{k} A^{(k)}(g).
\label{eq:ansground}
\end{equation}
They read the generation as a whole rather than step by step, and so discard the trace, but they are computed from the same encoder, the same input pool and the same similarity as the chain-level scores.

\paragraph{Extending the feature pool with answer-level grounding.}
The analyses above locate the aggregator's weakness in \emph{which} signals it has to choose from rather than in how it combines them. We therefore refit it, with the main-text protocol unchanged, on the feature vector of Equation~\ref{eq:feature-vector} extended by the three answer-level grounding scores of Equation~\ref{eq:ansground}, and evaluate over the same 100 randomised runs. Table~\ref{tab:cchain-plus} reports the result. With Gemini the split-averaged relative AUROC gain over Self-Consistency rises from $+2.15\%$ to $+2.62\%$, every split becomes significantly positive, and the multiple-choice split moves from a non-significant $-0.14\%$ to $+1.40\%$; the open-answer omni-modal split gains $+4.53\%$. With MiniCPM the average rises from $+0.66\%$ to $+0.84\%$ and AURAC improves on all four splits, and the open-answer split's AUROC loss becomes small and no longer significant ($-0.79\%$, $p=0.11$). Both columns are fitted and evaluated identically and differ only in the feature pool, so they are directly comparable.

\begin{table}[t]
\centering
\begin{adjustbox}{width=\textwidth}
\begin{tabular}{l ccc c ccc}
\toprule
& \multicolumn{3}{c}{AUROC} & & \multicolumn{3}{c}{AURAC} \\
\cmidrule(lr){2-4} \cmidrule(lr){6-8}
Split & Self-Cons. & $\mathcal{C}_{\text{chain}}$ & $\mathcal{C}^{+}_{\text{chain}}$ & & Self-Cons. & $\mathcal{C}_{\text{chain}}$ & $\mathcal{C}^{+}_{\text{chain}}$ \\
\midrule
\multicolumn{8}{l}{\emph{Gemini}} \\
Audio & 0.4512 & 0.4696 & 0.4700\textsuperscript{***} & & 0.8188 & 0.8126 & 0.8126 \\
MC & 0.5630 & 0.5626 & \textbf{0.5709}\textsuperscript{***} & & 0.6842 & 0.6863 & \textbf{0.6940}\textsuperscript{***} \\
MO & 0.6334 & 0.6584 & \textbf{0.6621}\textsuperscript{***} & & 0.5236 & 0.5199 & 0.5221 \\
Visual & 0.7232 & 0.7272 & 0.7257\textsuperscript{***} & & 0.8858 & 0.8915 & 0.8893\textsuperscript{***} \\
Avg & 0.5927 & 0.6044 & \textbf{0.6072} & & 0.7281 & 0.7276 & 0.7295 \\
\midrule
\multicolumn{8}{l}{\emph{MiniCPM}} \\
Audio & 0.7236 & 0.7368 & \textbf{0.7372}\textsuperscript{***} & & 0.8519 & \textbf{0.8612} & 0.8587\textsuperscript{***} \\
MC & 0.5461 & \textbf{0.5505} & 0.5489\textsuperscript{*} & & 0.4138 & \textbf{0.4238} & 0.4230\textsuperscript{***} \\
MO & \textbf{0.6782} & 0.6711 & 0.6727 & & 0.7154 & 0.7217 & \textbf{0.7241}\textsuperscript{***} \\
Visual & 0.6755 & 0.6838 & \textbf{0.6874}\textsuperscript{***} & & 0.7697 & 0.7717 & \textbf{0.7763}\textsuperscript{***} \\
Avg & 0.6559 & 0.6606 & \textbf{0.6616} & & 0.6877 & 0.6946 & \textbf{0.6955} \\
\bottomrule
\end{tabular}
\end{adjustbox}
\caption{The aggregator refit on the chain features alone ($\mathcal{C}_{\text{chain}}$) and on the same features extended with the three answer-level grounding scores of Equation~\ref{eq:ansground} ($\mathcal{C}^{+}_{\text{chain}}$), 100-run means under the main-text protocol. Asterisks mark the paired $t$-test of $\mathcal{C}^{+}_{\text{chain}}$ against Self-Consistency with the conventions of Table~\ref{tab:delta-combined}; bold marks the best of the three per row.}
\label{tab:cchain-plus}
\end{table}

\paragraph{Can the selection step itself be improved?}
The analyses above locate the aggregator's weakness in the \emph{selection} of a signal rather than in how signals are combined, and the per-split validation sets are small (50 questions on MO). We therefore evaluated six alternatives to the argmax over validation AUROC, all of which see validation labels only, over the same 100 randomised runs on the eight-feature pool. Three attack the variance of the selection: a $5$-fold cross-validated estimate of each feature's validation AUROC before the argmax (\emph{cv}), a soft vote over the argmaxes of $30$ bootstrap resamples of the validation items (\emph{bagged}), and selection on the pooled validation items of all four splits (\emph{pooled}). Three replace hard selection by a weighting of the per-question feature ranks: softmax weights over validation AUROC (\emph{softmax}), the mean rank of the top-$k$ features with $k$ chosen on validation (\emph{top-$k$}), and the uniform combination already reported above. Table~\ref{tab:selector-comparison} reports the split-averaged relative AUROC gain over Self-Consistency.

\begin{table}[t]
\centering
\begin{adjustbox}{width=0.8\columnwidth}
\begin{tabular}{l rr}
\toprule
Selector & Gemini & MiniCPM \\
\midrule
$\mathcal{C}_{\text{chain}}$, five chain features & $+2.12$ & $+0.68$ \\
$\mathcal{C}^{+}_{\text{chain}}$, eight features & $+2.60$ & $+0.84$ \\
argmax of validation AUROC & $+2.79$ & $\mathbf{+0.98}$ \\
cross-validated argmax & $+2.66$ & $+0.90$ \\
pooled-validation argmax & $+2.49$ & $-0.30$ \\
\midrule
softmax-weighted ranks & $\mathbf{+3.02}$ & $-2.04$ \\
bagged selection & $+2.58$ & $-1.62$ \\
top-$k$ mean rank & $+2.08$ & $-2.38$ \\
\bottomrule
\end{tabular}
\end{adjustbox}
\caption{Split-averaged relative AUROC gain over Self-Consistency (\%) for alternative selectors over the eight-feature pool, 100 randomised runs, validation-only selection throughout.}
\label{tab:selector-comparison}
\end{table}

Three conclusions follow, the third of them negative. First, \emph{widening the pool helps more than changing the rule}: every selector that keeps a hard argmax over the eight features beats the five-feature aggregator on both generators, and the single largest step is the addition of the answer-level grounding scores. Second, \emph{reducing selection variance helps only marginally}: the cross-validated argmax is within $0.15$ points of the plain argmax on both generators, and pooling the validation items across splits helps Gemini's smallest split but costs MiniCPM $1.3$ points on average, because it forces one feature onto four splits whose best signals genuinely differ. Third, \emph{rank-based soft combination is not a safe default}: it is the best selector on Gemini ($+3.02$) and the worst on MiniCPM ($-2.04$). The asymmetry has a concrete cause. The confidence attached to an answer is the sum of its cluster's per-candidate scores normalised by the total, a quantity that for narrow-band scores is dominated by the cluster's size and therefore carries the agreement signal itself. Replacing the raw values by per-question ranks removes that magnitude information; where agreement is weak (Gemini, Self-Consistency at $0.5927$ average AUROC) nothing is lost, but where it is strong (MiniCPM, $0.6559$) the combination discards the very signal it should be refining. We therefore retain the main-text protocol, and read these results as evidence that the aggregator's headroom lies in the quality and breadth of the individual signals rather than in the sophistication of their combination.

\paragraph{What the selection is worth, and why.}
Two measurements taken together explain the behaviour reported throughout this appendix. Table~\ref{tab:cost-of-selection} decomposes the distance between the deployable aggregator and the per-split oracle. The cost of not knowing in advance which signal is best is $0.0158$ AUROC with Gemini and $0.0116$ with MiniCPM, while the value of adapting the choice to the split at all is $0.0137$ and $0.0113$. The two quantities are the same size, so validation-based selection roughly \emph{breaks even} against simply fixing one good signal a priori ($0.6081$ against $0.6102$ with Gemini; $0.6624$ against $0.6627$ with MiniCPM). The heterogeneity it is trying to exploit is nonetheless real: the signal that wins a given split is highly stable across runs ($G_{\text{avg}}$ on Gemini-Audio in 100 runs out of 100, $aG_{\text{mean}}$ on Gemini-MC in 99) and differs completely between splits and generators. The instability therefore lies in the validation estimate of that ordering, not in the ordering itself.

\begin{table}[t]
\centering
\begin{adjustbox}{width=\textwidth}
\begin{tabular}{l ccccc ccccc}
\toprule
& \multicolumn{5}{c}{Gemini} & \multicolumn{5}{c}{MiniCPM} \\
\cmidrule(lr){2-6} \cmidrule(lr){7-11}
& Audio & MC & MO & Visual & Avg & Audio & MC & MO & Visual & Avg \\
\midrule
Self-Consistency & 0.4512 & 0.5630 & 0.6334 & 0.7232 & 0.5927 & 0.7236 & 0.5461 & 0.6782 & 0.6755 & 0.6559 \\
One signal fixed a priori & 0.4817 & 0.5576 & 0.6705 & 0.7311 & 0.6102 & 0.7402 & 0.5470 & 0.6945 & 0.6689 & 0.6627 \\
Per split, validation-selected & 0.4707 & 0.5730 & 0.6619 & 0.7268 & 0.6081 & 0.7363 & 0.5512 & 0.6782 & 0.6838 & 0.6624 \\
Per split, test-selected (oracle) & 0.4817 & 0.5854 & 0.6909 & 0.7377 & 0.6239 & 0.7428 & 0.5628 & 0.7006 & 0.6895 & 0.6739 \\
\midrule
oracle $-$ validation-selected & 0.0111 & 0.0124 & 0.0290 & 0.0110 & 0.0158 & 0.0066 & 0.0117 & 0.0224 & 0.0057 & 0.0116 \\
oracle $-$ fixed a priori & 0.0000 & 0.0278 & 0.0204 & 0.0066 & 0.0137 & 0.0026 & 0.0158 & 0.0062 & 0.0205 & 0.0113 \\
\bottomrule
\end{tabular}
\end{adjustbox}
\caption{Test AUROC over 100 randomised runs on the eight-signal pool. The oracle row selects on the test split and is a leakage ceiling, not a method; the fixed row uses a single signal for all four splits ($G_{\text{avg}}$ for Gemini, $aG_{\text{mean}}$ for MiniCPM), chosen once.}
\label{tab:cost-of-selection}
\end{table}

\paragraph{The geometric signals are redundant with each other and complementary to answer dispersion.}
The reason selection buys so little is that the eight signals largely measure the same thing. Table~\ref{tab:complementarity} reports the mean absolute Spearman correlation between their question-level confidences, computed \emph{within} each split and then averaged --- pooling the splits inflates every correlation through between-split level differences, the artefact analysed in Appendix~\ref{app:overall}. Among the eight geometric signals the mean correlation is $0.906$ (Gemini) and $0.942$ (MiniCPM), and the gap between the best and the worst of them on a given split never exceeds $0.03$ AUROC. Against the answer-dispersion baselines of Appendix~\ref{app:additional-baselines} the picture is entirely different: correlations fall to $0.408$ and $0.229$, and the sets of questions each family places in its least-confident decile are almost disjoint (Jaccard overlap $0.073$ and $0.107$, against $0.406$ and $0.764$ within the geometric family).

\begin{table}[t]
\centering
\begin{adjustbox}{width=0.9\columnwidth}
\begin{tabular}{l cc cc}
\toprule
& \multicolumn{2}{c}{mean $|\rho|$} & \multicolumn{2}{c}{low-decile Jaccard} \\
\cmidrule(lr){2-3} \cmidrule(lr){4-5}
& Gemini & MiniCPM & Gemini & MiniCPM \\
\midrule
Among the eight geometric signals & 0.906 & 0.942 & 0.406 & 0.764 \\
Geometric vs.\ answer dispersion & 0.408 & 0.229 & 0.073 & 0.107 \\
\bottomrule
\end{tabular}
\end{adjustbox}
\caption{Redundancy within the geometric family and complementarity with the answer-dispersion baselines. Spearman correlations between question-level confidences are computed within each split and averaged over the four; the Jaccard column compares the questions each score places in its own least-confident decile ($1.0$ = identical failure sets).}
\label{tab:complementarity}
\end{table}

This is the mechanism behind every negative result in this appendix. A family whose members correlate at $0.91$--$0.94$ offers almost no diversity for an aggregator to exploit, which is why the fitted model collapses to one feature, why combining the features does not beat selecting one, and why selection itself barely beats a fixed choice. It also predicts where the headroom actually is: the one family that is genuinely decorrelated from ours is between-sample answer dispersion, and it is exactly the hybrid of the two that removes every below-chance split in Appendix~\ref{app:additional-baselines}. We therefore present the geometric scores not as a set of independent evidence sources to be fused, but as several readings of one underlying quantity --- the geometric regularity of a reasoning trace in an external embedding space --- whose reliability depends on the generator and the modality, and whose natural complement lies outside the family.

\section{Conditioning on Agreement, and the Media-Permutation Control}
\label{app:conditional-media}

This appendix gives the protocols and the full tables behind Section~\ref{sec:complementarity}. Both analyses are computed on the canonical partition (split seed 42, validation fraction 0.2) and evaluated on its $2{,}984$ test questions, and MiniCPM uses the media-corrected records of Appendix~\ref{app:minicpm}.

\subsection{Discrimination with the agreement signal held constant}

\paragraph{Protocol.} A question's stratum is the number of sampled candidates falling in its majority cluster, an integer between $1$ and $N=6$; the stratum sizes are $\{1{:}740,\, 2{:}326,\, 3{:}289,\, 4{:}313,\, 5{:}424,\, 6{:}892\}$ with Gemini answers and $\{1{:}178,\, 2{:}243,\, 3{:}457,\, 4{:}522,\, 5{:}550,\, 6{:}1034\}$ with MiniCPM answers. Within a stratum self-consistency assigns every question the same confidence. The scored answer is fixed, for every method, to the one self-consistency returns --- the count-majority cluster, ties resolved as in the primary protocol of Section~\ref{sec:methodology} --- so the binary label is identical across methods and only the attached confidence differs; no number in Table~\ref{tab:conditional-full} can be earned by returning a different answer. We then report the stratified AUROC
\begin{equation}
A_{\text{strat}} = \frac{\sum_{s} \#\{\text{concordant pairs in stratum } s\}}{\sum_{s} n^{+}_{s} n^{-}_{s}},
\label{eq:stratified-auroc}
\end{equation}
with tied confidences counting one half, which is the ordinary AUROC restricted to pairs of questions sharing a stratum. Confidence intervals are question-level bootstrap intervals over $B=2{,}000$ resamples drawn within strata, so the bootstrap resolves a two-sided $p$ no finer than $5\times 10^{-4}$; we therefore mark intervals excluding $0.5$ rather than attaching the asterisk convention of the main tables. A signal that is constant inside every stratum has a degenerate bootstrap distribution and no test applies to it, which is the case for self-consistency by construction.

\paragraph{Results.} Table~\ref{tab:conditional-full} pairs each signal's unconditional AUROC on the same items with its stratified value. Conditioning costs every method most of its discrimination, which quantifies how much of the geometry-weighted result is the vote share re-expressed through Equation~\ref{eq:gwsc}: the median loss is $0.131$ AUROC over the $64$ cells occupied by a geometric signal and $0.097$ over all $88$ cells of the table. Three patterns in the residue are worth recording. First, the cells where a geometric signal survives and the cells where a dispersion baseline survives overlap once in eight. Second, the answer-level grounding scores behave differently from the chain-level ones: they are the strongest conditional signals on Gemini-MC ($0.5541$--$0.5603$) and among the weakest on Gemini-Visual ($0.4674$--$0.4685$), which is consistent with the picture of Appendix~\ref{app:aggregator-evidence} in which widening the feature pool helps because the pool members fail in different places. Third, the two dispersion baselines lose nothing on Gemini-Audio when agreement is held constant --- they gain, from $0.694$ and $0.693$ to $0.751$ and $0.756$ --- which is the sharpest available statement of the Audio failure mode: on that split the agreement signal is not merely uninformative but actively misleading, so removing it from the comparison \emph{helps} the signal that does not depend on it.

\paragraph{Questions self-consistency cannot rank.} The conditional analysis has a degenerate limit worth quantifying separately: questions whose count-majority cluster is not unique, where self-consistency supplies no ordering at all and the returned answer is decided by each method's own tie-break. These number $862$ of the $2{,}984$ Gemini test questions and $426$ of the MiniCPM ones, distributed as reported in Appendix~\ref{app:aggregator-evidence}.

\begin{table*}[t]
\centering
\begin{adjustbox}{width=\textwidth}
\begin{tabular}{l cc cc cc cc}
\toprule
& \multicolumn{2}{c}{Audio} & \multicolumn{2}{c}{MC} & \multicolumn{2}{c}{MO} & \multicolumn{2}{c}{Visual} \\
\cmidrule(lr){2-3} \cmidrule(lr){4-5} \cmidrule(lr){6-7} \cmidrule(lr){8-9}
Signal & plain & conditional & plain & conditional & plain & conditional & plain & conditional \\
\midrule
\multicolumn{9}{l}{\emph{Gemini}} \\
$S_{\text{smooth}}$ & 0.4606 & 0.5152 & 0.5679 & 0.5372 & 0.6442 & 0.5189 & 0.7285 & 0.5604\textsuperscript{\dag} \\
$S_{\text{goal}}$ & 0.4425 & 0.4612 & 0.5644 & 0.5168 & 0.6472 & 0.5284 & 0.7325 & 0.5779\textsuperscript{\dag} \\
$S_{\text{dens}}$ & 0.4535 & 0.4942 & 0.5655 & 0.5233 & 0.6424 & 0.5131 & 0.7299 & 0.5665\textsuperscript{\dag} \\
$G_{\text{avg}}$ & 0.4516 & 0.4885 & 0.5666 & 0.5298 & 0.6504 & 0.5385 & 0.7319 & 0.5753\textsuperscript{\dag} \\
$G_{\max}$ & 0.4740 & 0.5551 & 0.5605 & 0.4931 & 0.6411 & 0.5089 & 0.7145 & 0.4988 \\
$G^{\max}_{\text{ans}}$ & 0.4727 & 0.5514 & 0.5717 & 0.5603\textsuperscript{\dag} & 0.6471 & 0.5279 & 0.7074 & 0.4674 \\
$G^{\text{avg}}_{\text{ans}}$ & 0.4727 & 0.5514 & 0.5713 & 0.5576\textsuperscript{\dag} & 0.6534 & 0.5479 & 0.7075 & 0.4679 \\
$G^{\min}_{\text{ans}}$ & 0.4727 & 0.5514 & 0.5707 & 0.5541\textsuperscript{\dag} & 0.6609 & 0.5718 & 0.7077 & 0.4685 \\
RDS & 0.6941 & 0.7512\textsuperscript{\dag} & 0.5540 & 0.5231 & 0.5842 & 0.5131 & 0.6174 & 0.5328 \\
Sem.\ Vol. & 0.6928 & 0.7555\textsuperscript{\dag} & 0.5529 & 0.5249 & 0.5918 & 0.5128 & 0.6091 & 0.5304 \\
Self-Consistency & 0.4555 & 0.5000 & 0.5616 & 0.5000 & 0.6383 & 0.5000 & 0.7148 & 0.5001 \\
\midrule
\multicolumn{9}{l}{\emph{MiniCPM}} \\
$S_{\text{smooth}}$ & 0.7246 & 0.5167 & 0.5531 & 0.5198 & 0.7267 & 0.6009\textsuperscript{\dag} & 0.6909 & 0.5201 \\
$S_{\text{goal}}$ & 0.7179 & 0.4763 & 0.5409 & 0.4663 & 0.6980 & 0.5160 & 0.6893 & 0.5119 \\
$S_{\text{dens}}$ & 0.7277 & 0.5351 & 0.5529 & 0.5188 & 0.7093 & 0.5494 & 0.6918 & 0.5241 \\
$G_{\text{avg}}$ & 0.7345 & 0.5756\textsuperscript{\dag} & 0.5493 & 0.5030 & 0.6983 & 0.5169 & 0.6904 & 0.5173 \\
$G_{\max}$ & 0.7217 & 0.4992 & 0.5495 & 0.5039 & 0.7045 & 0.5351 & 0.6873 & 0.5019 \\
$G^{\max}_{\text{ans}}$ & 0.7344 & 0.5754\textsuperscript{\dag} & 0.5455 & 0.4865 & 0.6977 & 0.5151 & 0.6818 & 0.4747 \\
$G^{\text{avg}}_{\text{ans}}$ & 0.7344 & 0.5754\textsuperscript{\dag} & 0.5441 & 0.4805 & 0.6965 & 0.5117 & 0.6828 & 0.4798 \\
$G^{\min}_{\text{ans}}$ & 0.7344 & 0.5754\textsuperscript{\dag} & 0.5425 & 0.4732 & 0.6971 & 0.5135 & 0.6843 & 0.4871 \\
RDS & 0.7159 & 0.6391\textsuperscript{\dag} & 0.4929 & 0.4797 & 0.5136 & 0.4752 & 0.6089 & 0.5728\textsuperscript{\dag} \\
Sem.\ Vol. & 0.7206 & 0.6441\textsuperscript{\dag} & 0.5104 & 0.4970 & 0.5226 & 0.5090 & 0.6227 & 0.5860\textsuperscript{\dag} \\
Self-Consistency & 0.7219 & 0.5000 & 0.5486 & 0.5000 & 0.6926 & 0.5000 & 0.6869 & 0.5000 \\
\bottomrule
\end{tabular}
\end{adjustbox}
\caption{Unconditional AUROC and the stratified AUROC of Equation~\ref{eq:stratified-auroc} on the canonical partition's test items, with the scored answer fixed to the self-consistency choice. \textsuperscript{\dag} marks a $95\%$ bootstrap interval ($B=2{,}000$) excluding $0.5$. Self-consistency is constant within a stratum, so its conditional value is $0.5$ by construction and carries no test; the Visual entry differs from $0.5$ in the fourth decimal only through tie-breaking.}
\label{tab:conditional-full}
\end{table*}

\subsection{The media-permutation control}

\paragraph{Protocol.} Each test question is re-joined to another question's media and the grounding score is recomputed. Questions are permuted only within blocks that share both the split and the modality signature --- audio; image; video; image and audio; audio and video --- by a derangement, so every permuted score is computed from the same number of vectors of the same modalities, on the same split, from the same candidate answers, with the same encoder and the same similarity. The blocks have sizes $\{11, 34, 67, 133, 202, 598, 800, 1139\}$ with Gemini answers and $\{11, 41, 63, 137, 195, 605, 800, 1132\}$ with MiniCPM answers, and no draw among the $200$ leaves a question joined to its own media. We verified that the identity join reproduces the stored answer-grounding features to a worst absolute difference of $1.1\times 10^{-7}$ on both generators.

The control is run on the answer-level grounding of Equation~\ref{eq:ansground}: it is the grounding quantity that can be recomputed under a different join, since per-step embeddings are not retained. The same restriction applies to both generators, so the two blocks of Table~\ref{tab:media-permutation} are directly comparable, but the step-level $G_{\text{avg}}$ and $G_{\max}$ of the main tables are outside the scope of this test.

We evaluate in two modes. The \emph{grounding score} mode reads the score at question level, as the mean over the question's candidates, and asks whether that quantity separates correct from incorrect answers; this is the quantity Definition~\ref{def:grounding} defines. The \emph{deployed confidence} mode instead passes the same per-candidate scores through Equation~\ref{eq:gwsc}, as everywhere else in this work. The distinction matters because Equation~\ref{eq:gwsc} is a ratio within a question: a factor common to all of a question's candidates cancels, and a common additive shift pulls the resulting share towards the plain vote frequency.

\begin{figure}[t]
\centering
\includegraphics[width=0.85\textwidth]{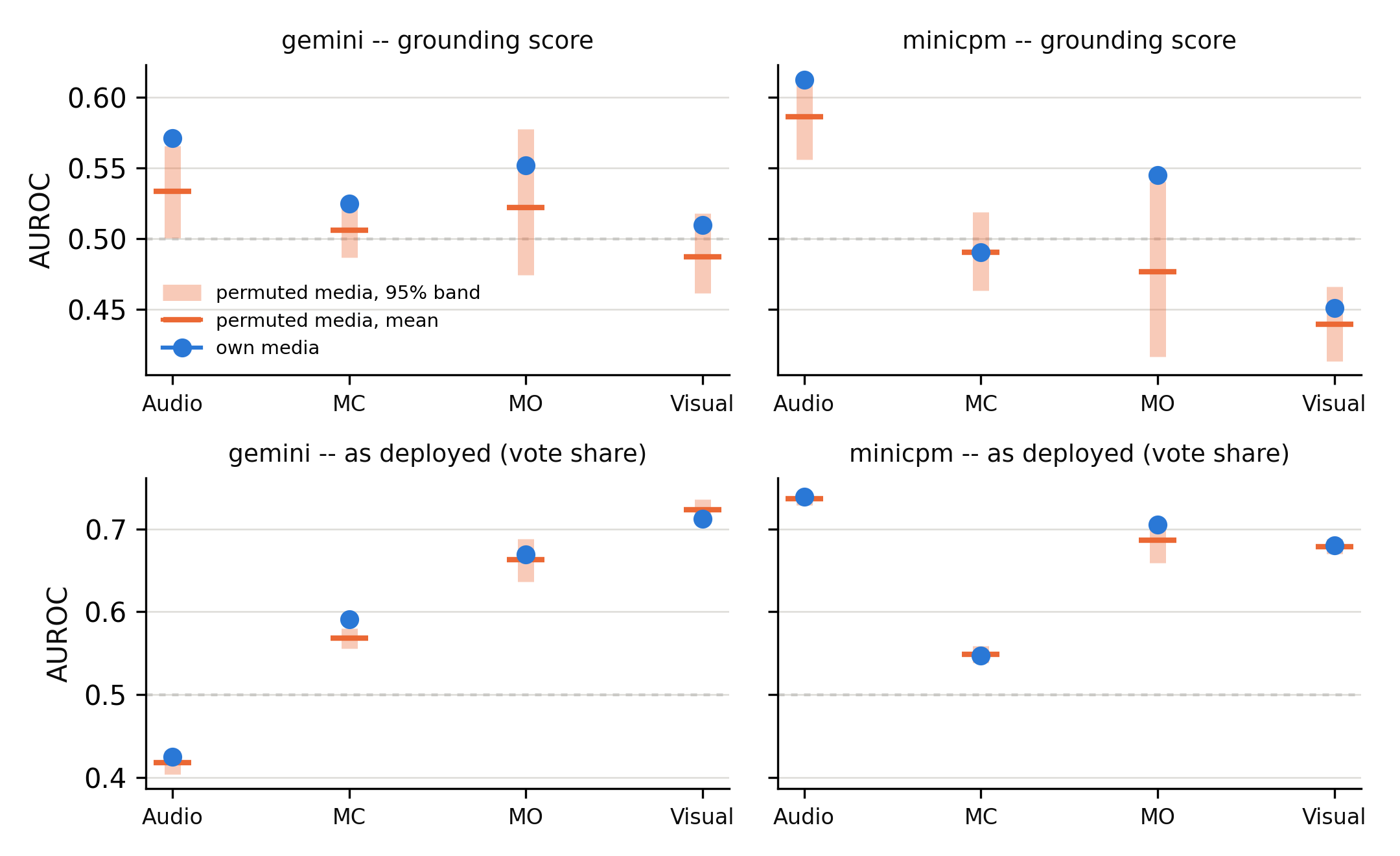}
\caption{Media-permutation control for $G^{\text{avg}}_{\text{ans}}$. Circles are the AUROC obtained with each question's own media; bars are the central $95\%$ of the null obtained by re-joining questions to other questions' media within blocks of equal split and modality signature, with the null mean marked. The upper row reads the grounding score at question level, the lower row passes it through the deployed vote share of Equation~\ref{eq:gwsc}, which normalises within a question and therefore cancels what the media contribute equally to all of its candidates. The dashed line is chance.}
\label{fig:media-permutation}
\end{figure}

\paragraph{Results.} Table~\ref{tab:media-permutation} and Figure~\ref{fig:media-permutation} report both modes. Read as a grounding score, the true join beats the permuted one in eight of the eight generator--split cells for $G^{\max}_{\text{ans}}$ and $G^{\text{avg}}_{\text{ans}}$ and in seven of eight for $G^{\min}_{\text{ans}}$, with split-averaged gains of $+0.026$ to $+0.029$ AUROC under Gemini and $+0.018$ to $+0.033$ under MiniCPM; the split-averaged permutation $p$ ranges from $0.005$ to $0.06$. Read as the deployed confidence, the same comparison gives $+0.002$ to $+0.007$ on the split average, within the permutation band in every configuration. The grounding score is thus demonstrably a function of the question's own media, and the vote-share protocol retains only the part of that function which varies between the candidates of a single question.

Two caveats bound the reading. The absolute discrimination of question-level grounding is low --- $0.517$ to $0.541$ on the split average --- so the media-specific component, while present, is a small signal on its own; and a permutation test of this construction can only establish that the media matter, not how much of the deployed method's performance they explain.

\begin{table*}[t]
\centering
\begin{adjustbox}{width=0.92\textwidth}
\begin{tabular}{ll cc c c c}
\toprule
Score & Split & observed & null mean & null $95\%$ band & $\Delta$ & $p$ \\
\midrule
\multicolumn{7}{l}{\emph{Gemini}} \\
\multicolumn{7}{l}{\quad\emph{grounding score}} \\
$G^{\max}_{\text{ans}}$ & Audio & 0.5711 & 0.5334 & [0.5004, 0.5656] & $+0.0377$ & 0.010 \\
 & MC & 0.5277 & 0.5062 & [0.4865, 0.5293] & $+0.0215$ & 0.040 \\
 & MO & 0.5569 & 0.5224 & [0.4691, 0.5769] & $+0.0345$ & 0.119 \\
 & Visual & 0.5068 & 0.4843 & [0.4590, 0.5157] & $+0.0226$ & 0.095 \\
 & Avg & 0.5406 & 0.5116 & [0.4935, 0.5292] & $+0.0291$ & 0.005 \\
$G^{\text{avg}}_{\text{ans}}$ & Avg & 0.5395 & 0.5122 & [0.4943, 0.5297] & $+0.0273$ & 0.005 \\
$G^{\min}_{\text{ans}}$ & Avg & 0.5382 & 0.5127 & [0.4947, 0.5324] & $+0.0255$ & 0.005 \\
\multicolumn{7}{l}{\quad\emph{deployed confidence}} \\
$G^{\max}_{\text{ans}}$ & Avg & 0.5962 & 0.5930 & [0.5844, 0.6016] & $+0.0032$ & 0.219 \\
$G^{\text{avg}}_{\text{ans}}$ & Avg & 0.5993 & 0.5930 & [0.5830, 0.6018] & $+0.0063$ & 0.075 \\
$G^{\min}_{\text{ans}}$ & Avg & 0.6001 & 0.5929 & [0.5830, 0.6024] & $+0.0072$ & 0.070 \\
\midrule
\multicolumn{7}{l}{\emph{MiniCPM}} \\
\multicolumn{7}{l}{\quad\emph{grounding score}} \\
$G^{\max}_{\text{ans}}$ & Audio & 0.6127 & 0.5861 & [0.5560, 0.6131] & $+0.0266$ & 0.035 \\
 & MC & 0.5136 & 0.4927 & [0.4652, 0.5233] & $+0.0209$ & 0.105 \\
 & MO & 0.4943 & 0.4804 & [0.4189, 0.5394] & $+0.0139$ & 0.348 \\
 & Visual & 0.4472 & 0.4386 & [0.4126, 0.4645] & $+0.0086$ & 0.259 \\
 & Avg & 0.5169 & 0.4995 & [0.4772, 0.5200] & $+0.0175$ & 0.060 \\
$G^{\text{avg}}_{\text{ans}}$ & Avg & 0.5248 & 0.4982 & [0.4773, 0.5194] & $+0.0267$ & 0.015 \\
$G^{\min}_{\text{ans}}$ & Avg & 0.5301 & 0.4975 & [0.4772, 0.5195] & $+0.0327$ & 0.010 \\
\multicolumn{7}{l}{\quad\emph{deployed confidence}} \\
$G^{\max}_{\text{ans}}$ & Avg & 0.6678 & 0.6625 & [0.6547, 0.6687] & $+0.0052$ & 0.090 \\
$G^{\text{avg}}_{\text{ans}}$ & Avg & 0.6679 & 0.6628 & [0.6551, 0.6711] & $+0.0051$ & 0.080 \\
$G^{\min}_{\text{ans}}$ & Avg & 0.6655 & 0.6633 & [0.6554, 0.6712] & $+0.0022$ & 0.264 \\
\bottomrule
\end{tabular}
\end{adjustbox}
\caption{Media-permutation control: AUROC on the true media join against the null distribution obtained from $200$ signature-preserving derangements, canonical partition, test items. Per-split rows are shown for $G^{\max}_{\text{ans}}$ and summarised by the split average for the other two aggregations; the complete cell-level table is released with the artifact. $\Delta$ is the observed value minus the null mean and $p$ is the one-sided permutation $p$-value, $(1+\#\{\text{null} \ge \text{observed}\})/(R+1)$. A join by record position rather than by question identity, which does not preserve the modality signature and is therefore not a draw from this null, gives a split-averaged grounding-score AUROC of $0.459$ under MiniCPM, below the null mean of $0.498$.}
\label{tab:media-permutation}
\end{table*}

\section{Transferability of the Learned Aggregation Weights}
\label{app:transferability}

The aggregated score $\mathcal{C}_{\text{chain}}$ is fitted per split on labelled validation data, so a natural question is whether the learned weights transfer to splits or answer-generating models other than the one they were fitted on. We test both directions on the canonical partition: for every ordered pair of splits we apply the aggregator fitted on the source split's validation items to the target split's test items (cross-split), and we apply each split's Gemini-fitted aggregator to the corresponding MiniCPM test items and vice versa, aligning features by semantic role (cross-model). The analysis is run in the main-text protocol; the replay reproduces the stored in-split values before any transfer is evaluated.

\paragraph{The fitted aggregator is nearly always a single-feature selector.} A structural observation frames the results: in 8 of the 10 fits (4 splits plus pooled, for each generator) the validation-selected aggregator collapses to a single feature with unit weight (e.g., $G_{\text{avg}}$ on Gemini-Audio and on MiniCPM-Audio, $S_{\text{smooth}}$ on Gemini-MO, $G_{\max}$ on MiniCPM-MO), and on Gemini-MC the fitted logistic regression is effectively constant, reducing the confidence to the majority-vote share. Transferring $\mathcal{C}_{\text{chain}}$ therefore amounts, in practice, to transferring a validation-selected choice of the best single chain-level feature.

\paragraph{Results.} Transfer is close to free on most pairs: the mean off-diagonal change in AUROC under cross-split transfer is $-0.015$ (Gemini) and $-0.001$ (MiniCPM), with all AURAC changes within $\pm 0.026$, and cross-model transfer changes pooled AUROC by at most $\pm 0.006$ (Gemini$\to$MiniCPM $+0.003$, MiniCPM$\to$Gemini $-0.005$). The failures concentrate exactly where the split-specific feature choice was the learned content: aggregators fitted on MC lose $0.04$--$0.06$ AUROC when applied to MO, and MiniCPM-fitted weights lose $0.046$ AUROC on Gemini-Audio, where MiniCPM's selected feature is near-worst for the other generator. Transferring fits obtained on the pooled union of splits across models is uniformly worse ($-0.02$ AUROC in both directions) than transferring the per-split fits. Two caveats apply: the Gemini-MC in-split value rests on tie-breaking of an effectively constant scorer and comparisons on that cell are noise; and even in-split, $\mathcal{C}_{\text{chain}}$ trails the best label-free single score on some splits, so transfer deltas of $\pm 0.02$ are small relative to the gap to oracle per-split score selection. Full matrices are included in the released result files.

The practical reading is twofold: the learned weights are not model-specific in any deep sense, so re-using them across generators is safe \emph{on average}; but because what they encode is a per-split feature choice, they inherit the feature-instability limitation discussed in Appendix~\ref{app:aggregator-evidence}, and a small labelled validation sample on the target configuration remains the reliable way to select or fit the aggregate --- one that Appendix~\ref{app:aggregator-evidence} shows can be as small as a few dozen examples.

\section{Robustness to the Choice of Omnimodal Encoder}
\label{app:lco-encoder}

All embedding-based scores in the main experiments are computed from E5-Omni~\citep{chen-etal-2026-e5} representations. To verify that our findings do not hinge on this particular encoder, we repeat the main-text evaluation (same Gemini-generated candidates, same protocol and 100 randomised runs) replacing E5-Omni with LCO-Embedding-Omni-7B~\citep{NEURIPS2025_e839f274}, a 7-billion-parameter omnimodal embedding model trained with a language-centric representation-learning recipe that differs from E5-Omni in training procedure, training data, and scale (7B vs.\ 3B parameters). Methods that do not depend on the encoder (Self-Consistency, UMPIRE, and the self-verbalisation baselines) are unaffected by this substitution and their rows repeat the main results; we include them for ease of comparison. Tables~\ref{tab:lco-auroc-ece} and~\ref{tab:lco-aurac} mirror Tables~\ref{tab:sf-gemini} and~\ref{tab:sf-aurac}.

\begin{table*}[t]
\centering
\begin{adjustbox}{width=\textwidth}
\begin{tabular}{l cc cc cc cc cc}
\toprule
& \multicolumn{2}{c}{Avg} & \multicolumn{2}{c}{Audio} & \multicolumn{2}{c}{MC} & \multicolumn{2}{c}{MO} & \multicolumn{2}{c}{Visual} \\
\cmidrule(lr){2-3} \cmidrule(lr){4-5} \cmidrule(lr){6-7} \cmidrule(lr){8-9} \cmidrule(lr){10-11}
Method & AUROC & ECE & AUROC & ECE & AUROC & ECE & AUROC & ECE & AUROC & ECE \\
\midrule
Self-Consistency & 0.5927 & 0.3068 & 0.4512 & 0.6203 & \textbf{0.5630} & 0.2723 & 0.6334 & 0.2459 & 0.7234 & \textbf{0.0888} \\
UMPIRE & 0.5825 & 0.4507 & \textbf{0.6318} & 0.7075 & 0.5551 & 0.3999 & 0.6159 & \textbf{0.1307} & 0.5272 & 0.5649 \\
\midrule
Level-Based Conf. & 0.5230 & 0.3192 & 0.5074 & 0.1661 & 0.5154 & 0.3466 & 0.5124 & 0.5777 & 0.5566 & 0.1866 \\
Self-Probing Conf. & 0.5803 & 0.3170 & 0.5705 & \textbf{0.1594} & 0.5495 & 0.3378 & 0.5871 & 0.5688 & 0.6143 & 0.2019 \\
Two-Step CoT Conf. & 0.5435 & 0.3147 & 0.5441 & 0.1611 & 0.5109 & 0.3473 & 0.5411 & 0.5522 & 0.5779 & 0.1981 \\
\midrule
$S_{\text{smooth}}$ & 0.5977 & 0.3013 & 0.4335 & 0.6116 & 0.5571 & 0.2754 & 0.6629 & 0.2264 & 0.7373 & 0.0917 \\
$S_{\text{dens}}$ & 0.5832 & 0.2973 & 0.4257 & 0.6016 & 0.5432 & 0.2792 & 0.6266 & 0.2149 & 0.7374 & 0.0935 \\
$S_{\text{goal}}$ & \textbf{0.6036} & \textbf{0.2968} & 0.4479 & 0.6064 & 0.5486 & 0.2872 & \textbf{0.6917} & 0.2035 & 0.7260 & 0.0899 \\
$G_{\text{avg}}$ & 0.6025 & 0.3013 & 0.4554 & 0.6219 & 0.5530 & 0.2756 & 0.6719 & 0.2153 & 0.7297 & 0.0924 \\
$G_{\max}$ & 0.5993 & 0.3022 & 0.4372 & 0.6139 & 0.5593 & \textbf{0.2674} & 0.6574 & 0.2378 & \textbf{0.7432} & 0.0896 \\
\midrule
$\mathcal{C}_{\text{chain}}$ & 0.6008 & 0.3015 & 0.4411 & 0.6179 & 0.5617 & 0.2746 & 0.6651 & 0.2209 & 0.7352 & 0.0925 \\
\bottomrule
\end{tabular}
\end{adjustbox}
\caption{AUROC ($\uparrow$) and ECE ($\downarrow$) across the UNO-Bench splits with all embedding-based scores computed from LCO-Embedding-Omni-7B instead of E5-Omni (Gemini-generated answers, same protocol as Table~\ref{tab:sf-gemini}). Avg as in Table~\ref{tab:sf-gemini}; bold indicates the best result per column.}
\label{tab:lco-auroc-ece}
\end{table*}

\begin{table}[t]
\centering
\begin{adjustbox}{width=\columnwidth}
\begin{tabular}{l c c c c c}
\toprule
Method & Avg & Audio & MC & MO & Visual \\
\midrule
Self-Consistency & 0.7281 & 0.8188 & 0.6842 & 0.5236 & 0.8860 \\
UMPIRE & 0.7240 & \textbf{0.8940} & 0.6920 & 0.5019 & 0.8080 \\
\midrule
Level-Based Conf. & 0.6760 & 0.8353 & 0.6388 & 0.4156 & 0.8144 \\
Self-Probing Conf. & 0.7042 & 0.8574 & 0.6704 & 0.4537 & 0.8352 \\
Two-Step CoT Conf. & 0.6885 & 0.8486 & 0.6503 & 0.4354 & 0.8198 \\
\midrule
$S_{\text{smooth}}$ & \textbf{0.7296} & 0.8134 & \textbf{0.6966} & 0.5099 & \textbf{0.8983} \\
$S_{\text{dens}}$ & 0.7249 & 0.8102 & 0.6890 & 0.5025 & 0.8981 \\
$S_{\text{goal}}$ & 0.7269 & 0.8117 & 0.6849 & 0.5207 & 0.8902 \\
$G_{\text{avg}}$ & 0.7268 & 0.8203 & 0.6741 & 0.5175 & 0.8952 \\
$G_{\max}$ & 0.7277 & 0.8162 & 0.6669 & \textbf{0.5338} & 0.8938 \\
\midrule
$\mathcal{C}_{\text{chain}}$ & 0.7284 & 0.8166 & 0.6840 & 0.5200 & 0.8931 \\
\bottomrule
\end{tabular}
\end{adjustbox}
\caption{AURAC ($\uparrow$) across the UNO-Bench splits with LCO-Embedding-Omni-7B (Gemini-generated answers). Avg is the unweighted mean of the four per-split values. Bold indicates the best result per column.}
\label{tab:lco-aurac}
\end{table}

Two observations emerge. First, the headline conclusions are stable under the change of encoder. On the split average, $\mathcal{C}_{\text{chain}}$ moves from an AUROC of 0.6046 with E5-Omni to 0.6008 with LCO and from an AURAC of 0.7275 to 0.7284; it remains ahead of Self-Consistency, UMPIRE, and all verbalised baselines on both average metrics, and the embedding scores remain the strongest family on the open-answer omni-modal split, where $S_{\text{goal}}$ with LCO attains the best MO AUROC observed with Gemini-generated answers (0.6917, against 0.6880 for the best E5-Omni score). For the aggregate, the absolute difference between encoders never exceeds 0.03 on any split--metric combination, and on Visual and MO the larger encoder is slightly beneficial.

Second, while the aggregate is stable, \emph{which} individual score leads shifts with the encoder, consistent with the heterogeneity observed in the main text. With LCO, $S_{\text{goal}}$ replaces $G_{\text{avg}}$ as the best average discriminator (0.6036, with $G_{\text{avg}}$ close behind), and $S_{\text{smooth}}$ attains the best average, MC, and Visual AURAC. The grounding scores are the most encoder-sensitive: $G_{\text{avg}}$ loses its average lead (0.6102 to 0.6025), whereas $G_{\max}$ gains substantially on the multiple-choice and Visual splits (0.5582 to 0.5593 and 0.7232 to 0.7432, the best Visual AUROC in the table) and attains the best MO AURAC (0.5338), while losing ground on Audio (0.4351 to 0.4372 remains below chance). That individual scores fluctuate with the encoder while $\mathcal{C}_{\text{chain}}$ does not is precisely the behaviour the aggregation was designed for, and we regard this as evidence that the proposed method is robust to the choice of the external omnimodal embedding model.

\section{Attention-Style Grounding: Exercising the Per-Modality Alignment Profile}
\label{app:grounding-attention}

The remark in the Methodology notes that, under a single shared encoder, the max-pooled omni-modal grounding coincides with its uni-modal counterpart, and that the value of the partitioned formulation lies in enabling aggregations that actually use the per-modality alignment profile $\{A^{(k)}(r_i)\}_{k=1}^{K}$. This appendix evaluates one such aggregation, an attention-style grounding score $G_{\text{att}}$, against the simple grounding scores $G_{\text{avg}}$ and $G_{\max}$ of the main text.

\paragraph{Definition.} The score has two instantiations, matching the structure of the input. In the \emph{omni-modal} setting ($K>1$), each reasoning step is softly routed to its best modality and rewarded for being decisive about it (entropy-gated routing): with $p_i = \mathrm{softmax}_k\!\left(A^{(k)}(r_i)/\tau\right)$ the routing distribution of step $r_i$ over modalities and $H(p_i)$ its Shannon entropy,
\begin{equation}
G_{\text{att}} \;=\; \frac{1}{m}\sum_{i=1}^{m}
\Big(1 - \tfrac{H(p_i)}{\log K}\Big)\,\max_{k} A^{(k)}(r_i),
\end{equation}
so that steps aligning strongly with a \emph{single} modality contribute fully, while generic steps that weakly align with everything are suppressed. In the \emph{uni-modal} setting, where routing over modalities is vacuous, the same sharpening is applied over steps instead (salience-weighted alignment): with $w = \mathrm{softmax}_i\!\left(A(r_i)/\tau\right)$,
\begin{equation}
G_{\text{att}} \;=\; \sum_{i=1}^{m} w_i\, A(r_i),
\end{equation}
which concentrates the score on the few most grounded anchor steps rather than averaging over the whole chain. Both variants use temperature $\tau=0.05$.

\paragraph{Protocol.} $G_{\text{att}}$ is evaluated exactly like the other individual scores in the main tables (sampling-free regime, 100 randomised runs); we verified our replay of the protocol by reproducing the stored $G_{\text{avg}}$ results of every run to numerical precision. Table~\ref{tab:grounding-attention} compares the three grounding scores across the three configurations of this work. Bold indicates the best of the three per column; asterisks follow the 100-run paired $t$-test convention of Table~\ref{tab:auroc-ece}, computed within each block.

\begin{table*}[t]
\centering
\begin{adjustbox}{width=\textwidth}
\begin{tabular}{ll cc cc cc cc cc}
\toprule
& & \multicolumn{2}{c}{Avg} & \multicolumn{2}{c}{Audio} & \multicolumn{2}{c}{MC} & \multicolumn{2}{c}{MO} & \multicolumn{2}{c}{Visual} \\
\cmidrule(lr){3-4} \cmidrule(lr){5-6} \cmidrule(lr){7-8} \cmidrule(lr){9-10} \cmidrule(lr){11-12}
Configuration & Score & AUROC & AURAC & AUROC & AURAC & AUROC & AURAC & AUROC & AURAC & AUROC & AURAC \\
\midrule
\multirow{3}{*}{Gemini, E5-Omni}
& $G_{\text{avg}}$ & \textbf{0.6102} & \textbf{0.7292} & \textbf{0.4817}\textsuperscript{***} & 0.8120 & 0.5576 & 0.6874 & \textbf{0.6705}\textsuperscript{***} & \textbf{0.5206}\textsuperscript{***} & \textbf{0.7311}\textsuperscript{***} & \textbf{0.8969}\textsuperscript{***} \\
& $G_{\max}$ & 0.5918 & 0.7268 & 0.4351 & 0.8107 & \textbf{0.5582} & \textbf{0.6900}\textsuperscript{***} & 0.6507 & 0.5179 & 0.7232 & 0.8888 \\
& $G_{\text{att}}$ & 0.5675 & 0.7107 & 0.4211 & \textbf{0.8145}\textsuperscript{***} & 0.5498 & 0.6674 & 0.5733 & 0.4693 & 0.7256 & 0.8916 \\
\midrule
\multirow{3}{*}{Gemini, LCO}
& $G_{\text{avg}}$ & \textbf{0.6025} & 0.7268 & 0.4554 & \textbf{0.8203} & 0.5530 & \textbf{0.6741} & \textbf{0.6719}\textsuperscript{***} & 0.5175 & 0.7297 & \textbf{0.8952}\textsuperscript{***} \\
& $G_{\max}$ & 0.5993 & \textbf{0.7277} & 0.4372 & 0.8162 & \textbf{0.5593}\textsuperscript{***} & 0.6669 & 0.6574 & \textbf{0.5338}\textsuperscript{***} & \textbf{0.7432}\textsuperscript{***} & 0.8938 \\
& $G_{\text{att}}$ & 0.5938 & 0.7232 & \textbf{0.4593}\textsuperscript{***} & 0.8197 & 0.5538 & 0.6736 & 0.6316 & 0.5048 & 0.7305 & 0.8949 \\
\midrule
\multirow{3}{*}{MiniCPM, E5-Omni}
& $G_{\text{avg}}$ & \textbf{0.6588} & \textbf{0.7015} & \textbf{0.7419}\textsuperscript{***} & \textbf{0.8608}\textsuperscript{***} & \textbf{0.5507} & \textbf{0.4314} & 0.6675 & \textbf{0.7339}\textsuperscript{***} & 0.6750 & \textbf{0.7798}\textsuperscript{***} \\
& $G_{\max}$ & 0.6584 & 0.6913 & 0.7223 & 0.8445 & 0.5439 & 0.4265 & \textbf{0.6978}\textsuperscript{***} & 0.7223 & 0.6695 & 0.7719 \\
& $G_{\text{att}}$ & 0.6427 & 0.6830 & 0.7304 & 0.8387 & 0.5428 & 0.4175 & 0.6216 & 0.7121 & \textbf{0.6760}\textsuperscript{***} & 0.7638 \\
\bottomrule
\end{tabular}
\end{adjustbox}
\caption{The attention-style grounding score $G_{\text{att}}$ against the simple grounding scores, sampling-free regime, AUROC and AURAC ($\uparrow$; means over 100 randomised runs). Avg is the unweighted mean of the four per-split values. Bold marks the best of the three scores per column within each configuration; asterisks follow the 100-run paired $t$-test convention of Table~\ref{tab:auroc-ece} and are not attached to the derived Avg column.}
\label{tab:grounding-attention}
\end{table*}

\paragraph{Findings.} The picture is configuration-dependent, and instructive on both ends. With MiniCPM-generated answers, $G_{\text{att}}$ retains the best Visual AUROC but trails both simple scores on the split average (0.6427 against 0.6588 for $G_{\text{avg}}$ and 0.6584 for $G_{\max}$), its weakness being concentrated on MO, which is precisely the split where $G_{\max}$ is strongest (0.6978). With Gemini-generated answers and the E5-Omni encoder, the plain mean alignment $G_{\text{avg}}$ is clearly superior (0.6102 vs.\ 0.5675 on average), and with the LCO encoder $G_{\max}$ leads on the split average (0.5993 against 0.6025 for $G_{\text{avg}}$ and 0.5938 for $G_{\text{att}}$). The sharpened aggregation is therefore not a uniform improvement: its benefit appears to depend on how sharply the encoder separates modality alignments at the fixed temperature, and we have not tuned $\tau$.

One pattern, however, is consistent across all three configurations: $G_{\text{att}}$ is the \emph{worst} of the three scores on the open-answer omni-modal split (e.g.\ 0.5733 vs.\ 0.6705 for $G_{\text{avg}}$ with E5-Omni), precisely the split where our scores otherwise lead all baselines. This is perhaps unsurprising in retrospect: the entropy gate rewards steps that commit to a single modality, but open-answer omni-modal questions are exactly those whose reasoning must integrate audio and visual evidence within the same step, so the decisiveness prior penalises the behaviour that characterises correct answers there. In our view this is useful evidence about the design space: the per-modality alignment profile is genuinely informative (with MiniCPM it yields the best grounding score we have on the Audio and Visual splits), but a hard one-modality-per-step prior is the wrong inductive bias for genuinely omni-modal reasoning, and modality-aware aggregations that instead reward \emph{balanced} cross-modal alignment are the natural next candidate.

\section{Geometry-Weighted Self-Consistency}
\label{app:gwsc}

In the main protocol the chain-level scores \emph{grade} a fixed answer: the returned answer is the count-majority cluster of the sampled candidates, and the confidence attached to it is the normalised score-mass (sum-share) of that cluster. This appendix evaluates the natural next step, in which the same score-mass is also allowed to \emph{select} the answer: each sampled chain votes with a weight equal to its chain-level score, so a minority cluster of well-grounded, coherent chains can overturn the count majority. The confidence formation is identical to the main protocol (the winning cluster's sum-share); the only difference is which cluster wins. 

\paragraph{Method.} Standard self-consistency treats all sampled reasoning chains equally. Given $N$ sampled generations $\{(R^{(j)},a^{(j)})\}_{j=1}^{N}$, we compute for each the chain-level feature vector
\[
\mathbf{x}^{(j)} =
\left[
S_{\text{smooth}}^{(j)},
S_{\text{dens}}^{(j)},
S_{\text{goal}}^{(j)},
G_{\text{avg}}^{(j)},
G_{\max}^{(j)}
\right]
\]
and a confidence score $c^{(j)}$, obtained either from the logistic-regression aggregator of the main text, $c^{(j)}=\mathcal{C}_{\text{chain}}(\mathbf{x}^{(j)})$, or directly from a single feature in the individual-score ablations. The standard self-consistency estimate, $\hat{P}(y|q,V)=\frac{1}{N}\sum_{j}\mathbf{1}(a^{(j)}=y)$, is then replaced by its geometry-weighted counterpart:
\begin{equation}
\hat{P}(y|q,V)
=
\frac{
\sum_{j=1}^{N}
c^{(j)}\mathbf{1}(a^{(j)}=y)
}{
\sum_{j=1}^{N} c^{(j)}
},
\end{equation}
which assigns larger weights to answers supported by coherent and well-grounded reasoning chains; the answer returned is $\arg\max_y \hat{P}(y|q,V)$ and its confidence is $\hat{P}$ itself. The relation to the main protocol is exact: there, the same $\hat{P}(y|q,V)$ is computed but the count-majority answer is returned (with $\hat{P}$ of that answer as its confidence), so the two settings differ only in whether the geometric evidence may change the selected answer, not in how confidence is formed. The evaluation protocol is identical to the main experiments ($N=6$, 100 randomised runs).

\begin{table*}[t]
\centering
\begin{adjustbox}{width=\textwidth}
\begin{tabular}{l cc cc cc cc cc}
\toprule
& \multicolumn{2}{c}{Avg} & \multicolumn{2}{c}{Audio} & \multicolumn{2}{c}{MC} & \multicolumn{2}{c}{MO} & \multicolumn{2}{c}{Visual} \\
\cmidrule(lr){2-3} \cmidrule(lr){4-5} \cmidrule(lr){6-7} \cmidrule(lr){8-9} \cmidrule(lr){10-11}
Method & AUROC & ECE & AUROC & ECE & AUROC & ECE & AUROC & ECE & AUROC & ECE \\
\midrule
Self-Consistency & 0.5927 & 0.3068 & 0.4512 & 0.6203 & \textbf{0.5630} & 0.2723 & 0.6334 & 0.2459 & 0.7234 & \textbf{0.0888}\textsuperscript{***} \\
Level-Based & 0.5230 & 0.3192 & 0.5074 & 0.1661 & 0.5154 & 0.3466 & 0.5124 & 0.5777 & 0.5566 & 0.1866 \\
Self-Probing & 0.5803 & 0.3170 & 0.5705 & \textbf{0.1594}\textsuperscript{***} & 0.5495 & 0.3378 & 0.5871 & 0.5688 & 0.6143 & 0.2019 \\
Two-Step CoT & 0.5435 & 0.3147 & 0.5441 & 0.1611 & 0.5109 & 0.3473 & 0.5411 & 0.5522 & 0.5779 & 0.1981 \\
UMPIRE & 0.5825 & 0.4507 & \textbf{0.6318}\textsuperscript{***} & 0.7075 & 0.5551 & 0.3999 & 0.6159 & \textbf{0.1307}\textsuperscript{***} & 0.5272 & 0.5649 \\
\midrule
$S_{\text{smooth}}$ & 0.6068 & \textbf{0.2980} & 0.4551 & 0.6124 & 0.5614 & 0.2758 & \textbf{0.6795}\textsuperscript{***} & 0.2109 & 0.7311 & 0.0928 \\
$S_{\text{dens}}$ & 0.5923 & 0.3057 & 0.4355 & 0.6174 & 0.5405 & 0.2860 & 0.6563 & 0.2271 & \textbf{0.7368}\textsuperscript{*} & 0.0923 \\
$S_{\text{goal}}$ & 0.5990 & 0.3052 & 0.4563 & 0.6122 & 0.5465 & 0.2895 & 0.6577 & 0.2224 & 0.7356 & 0.0969 \\
$G_{\text{avg}}$ & \textbf{0.6084} & 0.3005 & 0.4769 & 0.6061 & 0.5541 & 0.2826 & 0.6699 & 0.2179 & 0.7328 & 0.0954 \\
$G_{\max}$ & 0.5918 & 0.3043 & 0.4351 & 0.6148 & 0.5582 & \textbf{0.2718} & 0.6507 & 0.2355 & 0.7232 & 0.0952 \\
\midrule
$\mathcal{C}_{\text{chain}}$ & 0.6011 & 0.3014 & 0.4621 & 0.6107 & 0.5629 & 0.2745 & 0.6521 & 0.2267 & 0.7275 & 0.0936 \\
\bottomrule
\end{tabular}
\end{adjustbox}
\caption{Geometry-weighted majority voting: AUROC ($\uparrow$) and ECE ($\downarrow$) across the UNO-Bench splits with Gemini-generated answers. Every confidence method re-weights the votes of the $n=6$ candidates. Avg is the unweighted mean of the four per-split values (pooled-union results in Appendix~\ref{app:overall}). Bold indicates the best result per column; asterisks here denote 100-run paired $t$-test comparisons with the runner-up ($p<0.01$, $p<0.001$, and $p<0.0001$ for one, two, and three asterisks), unlike the item-level convention of the main tables, and are not attached to the derived Avg column.}
\label{tab:auroc-ece}
\end{table*}

\begin{table}[t]
\centering
\begin{adjustbox}{width=\columnwidth}
\begin{tabular}{l c c c c c}
\toprule
Method & Avg & Audio & MC & MO & Visual \\
\midrule
Self-Consistency & 0.7281 & 0.8188 & 0.6842 & 0.5236 & 0.8860 \\
Level-Based & 0.6760 & 0.8353 & 0.6388 & 0.4156 & 0.8144 \\
Self-Probing & 0.7042 & 0.8574 & 0.6704 & 0.4537 & 0.8352 \\
Two-Step CoT & 0.6885 & 0.8486 & 0.6503 & 0.4354 & 0.8198 \\
UMPIRE & 0.7240 & \textbf{0.8940}\textsuperscript{***} & \textbf{0.6920} & 0.5019 & 0.8080 \\
\midrule
$S_{\text{smooth}}$ & 0.7276 & 0.8112 & 0.6841 & 0.5200 & 0.8951 \\
$S_{\text{dens}}$ & 0.7268 & 0.8120 & 0.6752 & \textbf{0.5254} & 0.8945 \\
$S_{\text{goal}}$ & \textbf{0.7309} & 0.8161 & 0.6902 & 0.5188 & \textbf{0.8987}\textsuperscript{***} \\
$G_{\text{avg}}$ & 0.7293 & 0.8131 & 0.6870 & 0.5202 & 0.8969 \\
$G_{\max}$ & 0.7268 & 0.8107 & 0.6900 & 0.5179 & 0.8888 \\
\midrule
$\mathcal{C}_{\text{chain}}$ & 0.7279 & 0.8144 & 0.6867 & 0.5192 & 0.8912 \\
\bottomrule
\end{tabular}
\end{adjustbox}
\caption{Geometry-weighted majority voting: AURAC ($\uparrow$) across the UNO-Bench splits with Gemini-generated answers. Avg, bold, and asterisks as in Table~\ref{tab:auroc-ece}.}
\label{tab:aurac}
\end{table}

\begin{table}[t]
\centering
\begin{adjustbox}{width=\columnwidth}
\begin{tabular}{l r r r}
\toprule
Split & $\Delta$AUROC (\%) & $\Delta$ECE (\%) & $\Delta$AURAC (\%) \\
\midrule
Audio   & $+2.44$\textsuperscript{***} & $-1.55$\textsuperscript{***} & $-0.54$\textsuperscript{***} \\
MC      & $-0.03$ {\scriptsize(n.s.)}  & $+0.82$\textsuperscript{*}   & $+0.36$\textsuperscript{*} \\
MO      & $+2.97$\textsuperscript{***} & $-7.71$\textsuperscript{***} & $-0.82$\textsuperscript{***} \\
Visual  & $+0.56$\textsuperscript{***} & $+5.44$\textsuperscript{***} & $+0.60$\textsuperscript{***} \\
\midrule
Average & $+1.49$ & $-0.75$ & $-0.10$ \\
\bottomrule
\end{tabular}
\end{adjustbox}
\caption{Mean paired relative difference (\%) of the geometry-weighted $\mathcal{C}_{\text{chain}}$ with respect to Self-Consistency across 100 runs (Gemini-generated answers). Positive $\Delta$AUROC/$\Delta$AURAC and negative $\Delta$ECE favour the weighted variant. Asterisks: paired $t$-test, \textsuperscript{*}$p<0.01$, \textsuperscript{**}$p<0.001$, \textsuperscript{***}$p<0.0001$. The Average row is the unweighted mean of the four per-split deltas and carries no significance test; pooled-union deltas are reported in Appendix~\ref{app:overall}.}
\label{tab:paired}
\end{table}

\paragraph{Results.} Tables~\ref{tab:auroc-ece} and~\ref{tab:aurac} report the geometry-weighted results with Gemini-generated answers, in the format of the main tables; Table~\ref{tab:paired} gives the paired relative differences of $\mathcal{C}_{\text{chain}}$ against plain Self-Consistency across the 100 matched runs. The weighted $\mathcal{C}_{\text{chain}}$ raises the split-averaged AUROC from 0.5927 to 0.6011, with the largest gains on the open-answer omni-modal split (AUROC 0.6334 to 0.6521), while the split-averaged AURAC is essentially unchanged (0.7279 vs.\ 0.7281), the AUROC gains on Audio and MO coming with small AURAC losses on those same splits (Table~\ref{tab:paired}); the run-level analysis marks the AUROC improvements as highly systematic. The gains are nonetheless small in absolute terms, and most of the weighted scheme's performance is inherited from the strength of self-consistency itself.

\paragraph{Item-level significance.} Since the 100 runs share the same underlying question pool, we complement the run-level tests with a paired bootstrap over test questions on a canonical partition ($B=10^4$), analogous to the main-text convention; these comparisons are computed on the pooled union of splits and carry the caveat of Appendix~\ref{app:overall}. At the item level, the pooled gains of the weighted $\mathcal{C}_{\text{chain}}$ over Self-Consistency are significant with MiniCPM-generated answers (AUROC $+0.014$, $p=0.006$; AURAC $+0.023$, $p=0.0007$) and marginal with Gemini ($+0.014$, $p=0.017$; $+0.011$, $p=0.026$), while no single-split difference reaches significance and the result does not replicate on the canonical partition of the alternative-encoder configuration of Appendix~\ref{app:lco-encoder}. One honestly-selected per-split result stands out: with the LCO encoder, validation selects $G_{\max}$ as the weighting signal on Audio, and the choice holds up at the item level ($+0.120$ AUROC over Self-Consistency, $p=0.003$).


\section{Geometry-Weighted Results with MiniCPM}
\label{app:minicpm}

To complement the geometry-weighted results of Appendix~\ref{app:gwsc}, we repeat that evaluation protocol using $n=6$ candidate answers generated by MiniCPM instead of Gemini. As in the main text, MiniCPM being open-source means that UMPIRE here runs in its \emph{default} configuration, computing internal states from the very model that generated the answers, rather than through the external proxy model required for closed models such as Gemini. All other settings are unchanged (means across 100 randomized evaluation runs; asterisks follow the 100-run $t$-test convention of Table~\ref{tab:auroc-ece}). Tables~\ref{tab:minicpm-auroc-ece}--\ref{tab:minicpm-paired} mirror Tables~\ref{tab:auroc-ece}--\ref{tab:paired}.

\begin{table*}[t]
\centering
\begin{adjustbox}{width=\textwidth}
\begin{tabular}{l cc cc cc cc cc}
\toprule
& \multicolumn{2}{c}{Avg} & \multicolumn{2}{c}{Audio} & \multicolumn{2}{c}{MC} & \multicolumn{2}{c}{MO} & \multicolumn{2}{c}{Visual} \\
\cmidrule(lr){2-3} \cmidrule(lr){4-5} \cmidrule(lr){6-7} \cmidrule(lr){8-9} \cmidrule(lr){10-11}
Method & AUROC & ECE & AUROC & ECE & AUROC & ECE & AUROC & ECE & AUROC & ECE \\
\midrule
Majority Vote (Self-Consistency) & 0.6559 & 0.2553 & 0.7236 & 0.1310 & 0.5461 & 0.3198 & 0.6782 & 0.4051 & 0.6755 & 0.1653 \\
Level-Based Conf. & 0.6200 & 0.2699 & 0.6306 & 0.1796 & 0.6024 & 0.3905 & 0.5956 & 0.2745 & 0.6512 & 0.2350 \\
Self-Probing Conf. & 0.6563 & 0.2843 & 0.6941 & 0.1753 & \textbf{0.6331}\textsuperscript{***} & 0.4658 & 0.5933 & 0.2521 & 0.7047 & 0.2440 \\
Two-Step CoT Conf. & 0.6294 & 0.3133 & 0.6791 & 0.1830 & 0.6252 & 0.4804 & 0.5609 & 0.3028 & 0.6525 & 0.2871 \\
UMPIRE & 0.6570 & \textbf{0.1577} & \textbf{0.7507}\textsuperscript{***} & 0.1550 & 0.5840 & 0.2898 & 0.5875 & \textbf{0.1323}\textsuperscript{***} & \textbf{0.7058} & \textbf{0.0538}\textsuperscript{***} \\
\midrule
$S_{\text{smooth}}$ & 0.6564 & 0.2628 & 0.7270 & 0.1406 & 0.5628 & 0.3116 & 0.6531 & 0.4357 & 0.6829 & 0.1632 \\
$S_{\text{dens}}$ & 0.6538 & 0.2590 & 0.7344 & \textbf{0.1293} & 0.5572 & 0.3141 & 0.6347 & 0.4317 & 0.6887 & 0.1610 \\
$S_{\text{goal}}$ & 0.6522 & 0.2627 & 0.7295 & 0.1354 & 0.5451 & 0.3155 & 0.6462 & 0.4408 & 0.6878 & 0.1593 \\
$G_{\text{avg}}$ & 0.6588 & 0.2649 & 0.7419 & 0.1346 & 0.5507 & 0.3179 & 0.6675 & 0.4382 & 0.6750 & 0.1690 \\
$G_{\max}$ & 0.6584 & 0.2564 & 0.7223 & 0.1332 & 0.5439 & 0.3225 & \textbf{0.6978}\textsuperscript{***} & 0.4013 & 0.6695 & 0.1686 \\
\midrule
$\mathcal{C}_{\text{chain}}$ & \textbf{0.6603} & 0.2573 & 0.7361 & 0.1331 & 0.5512 & 0.3175 & 0.6702 & 0.4165 & 0.6836 & 0.1621 \\
\bottomrule
\end{tabular}
\end{adjustbox}
\caption{AUROC ($\uparrow$) and ECE ($\downarrow$) across the UNO-Bench splits with MiniCPM-generated answers. Avg is the unweighted mean of the four per-split values (pooled-union results in Appendix~\ref{app:overall}). Best result per column in bold; asterisks as in Table~\ref{tab:auroc-ece}, not attached to the derived Avg column.}
\label{tab:minicpm-auroc-ece}
\end{table*}

\begin{table*}[t]
\centering
\begin{adjustbox}{width=0.85\textwidth}
\begin{tabular}{l c c c c c}
\toprule
Method & Avg & Audio & MC & MO & Visual \\
\midrule
Majority Vote (Self-Consistency) & 0.6865 & 0.8500 & 0.4135 & 0.7149 & 0.7677 \\
Level-Based Conf. & 0.6573 & 0.7918 & 0.4343 & 0.6652 & 0.7380 \\
Self-Probing Conf. & 0.6931 & 0.8311 & \textbf{0.4815}\textsuperscript{***} & 0.6733 & 0.7866 \\
Two-Step CoT Conf. & 0.6695 & 0.8234 & 0.4692 & 0.6402 & 0.7453 \\
UMPIRE & 0.6968 & \textbf{0.8816}\textsuperscript{***} & 0.4451 & 0.6621 & \textbf{0.7985}\textsuperscript{***} \\
\midrule
$S_{\text{smooth}}$ & 0.6976 & 0.8630 & 0.4273 & 0.7310 & 0.7691 \\
$S_{\text{dens}}$ & 0.6924 & 0.8597 & 0.4219 & 0.7191 & 0.7690 \\
$S_{\text{goal}}$ & 0.6866 & 0.8463 & 0.4027 & 0.7265 & 0.7709 \\
$G_{\text{avg}}$ & \textbf{0.7015} & 0.8608 & 0.4314 & \textbf{0.7339}\textsuperscript{***} & 0.7798 \\
$G_{\max}$ & 0.6913 & 0.8445 & 0.4265 & 0.7223 & 0.7719 \\
\midrule
$\mathcal{C}_{\text{chain}}$ & 0.6928 & 0.8548 & 0.4180 & 0.7232 & 0.7753 \\
\bottomrule
\end{tabular}
\end{adjustbox}
\caption{AURAC ($\uparrow$) across the UNO-Bench splits with MiniCPM-generated answers. Avg as in Table~\ref{tab:minicpm-auroc-ece}. Best result per column in bold; asterisks as in Table~\ref{tab:auroc-ece}, not attached to the derived Avg column.}
\label{tab:minicpm-aurac}
\end{table*}

\begin{table}[t]
\centering
\begin{adjustbox}{width=\columnwidth}
\begin{tabular}{l r r r}
\toprule
Split & $\Delta$AUROC (\%) & $\Delta$ECE (\%) & $\Delta$AURAC (\%) \\
\midrule
Audio   & $+1.72$\textsuperscript{***} & $+1.61$\textsuperscript{**}  & $+1.25$\textsuperscript{***} \\
MC      & $+0.93$\textsuperscript{***} & $-0.70$\textsuperscript{***} & $+2.09$\textsuperscript{***} \\
MO      & $-1.16$\textsuperscript{**}  & $+2.87$\textsuperscript{***} & $+1.12$\textsuperscript{***} \\
Visual  & $+1.21$\textsuperscript{***} & $-1.93$\textsuperscript{***} & $+0.51$\textsuperscript{**} \\
\midrule
Average & $+0.68$ & $+0.46$ & $+1.24$ \\
\bottomrule
\end{tabular}
\end{adjustbox}
\caption{Mean paired relative difference (\%) of $\mathcal{C}_{\text{chain}}$ with respect to Self-Consistency across 100 runs with MiniCPM-generated answers. Positive $\Delta$AUROC/$\Delta$AURAC and negative $\Delta$ECE favor our method. Asterisks as in Table~\ref{tab:paired}. The Average row is the unweighted mean of the four per-split deltas and carries no significance test; pooled-union deltas are reported in Appendix~\ref{app:overall}.}
\label{tab:minicpm-paired}
\end{table}

The MiniCPM results confirm the findings of Appendix~\ref{app:gwsc} while sharpening the comparison with UMPIRE. First, the benefit of our geometric re-weighting over plain self-consistency is more uniform than with Gemini: the paired analysis (Table~\ref{tab:minicpm-paired}) shows significant relative gains on all four splits for AURAC (average $+1.24\%$, against $-0.10\%$ with Gemini) and on the Audio, multiple-choice, and Visual splits for AUROC (average $+0.68\%$); the single exception is AUROC on the open-answer omni-modal split ($-1.16\%$\textsuperscript{**}), where AURAC nevertheless improves ($+1.12\%$\textsuperscript{***}). Second, as expected, UMPIRE benefits substantially from its native configuration: with internal states from the generating model itself it becomes the strongest method on the unimodal Audio and Visual splits and achieves the best average calibration (ECE 0.1577), while $G_{\text{avg}}$ attains the best average AURAC (0.7015, against 0.6968 for UMPIRE) and our aggregated score retains the best average AUROC in the table (0.6603). Crucially, however, its weakness in genuinely omni-modal settings persists: the open-answer omni-modal split remains UMPIRE's worst setting (AUROC 0.5875 vs.\ 0.6702 for our method), and paired tests confirm that $\mathcal{C}_{\text{chain}}$ outperforms it there by $+13.8\%$ AUROC\textsuperscript{***} and $+9.3\%$ AURAC\textsuperscript{***}. This demonstrates that UMPIRE's difficulty in handling more than one extra-textual modality at a time is structural rather than an artifact of the proxy-model setting used for closed models---and that our method retains its omni-modal advantage while requiring no access to internal states whatsoever, and therefore applying unchanged to closed-source models where UMPIRE's default setting is simply unavailable. Finally, $G_{\max}$ behaves very differently across the two generators: it is the weakest signal on Gemini-Audio and the strongest single signal on MiniCPM-MO (0.6978, above Self-Consistency's 0.6782), which indicates that the discriminative power of individual embedding-based components depends on the answer-generating model and motivates selecting among them rather than fixing one a priori.

\section{Effect of the Number of Sampled Candidates}
\label{app:n-sweep}

\begin{figure*}[t]
\centering
\includegraphics[width=0.32\textwidth]{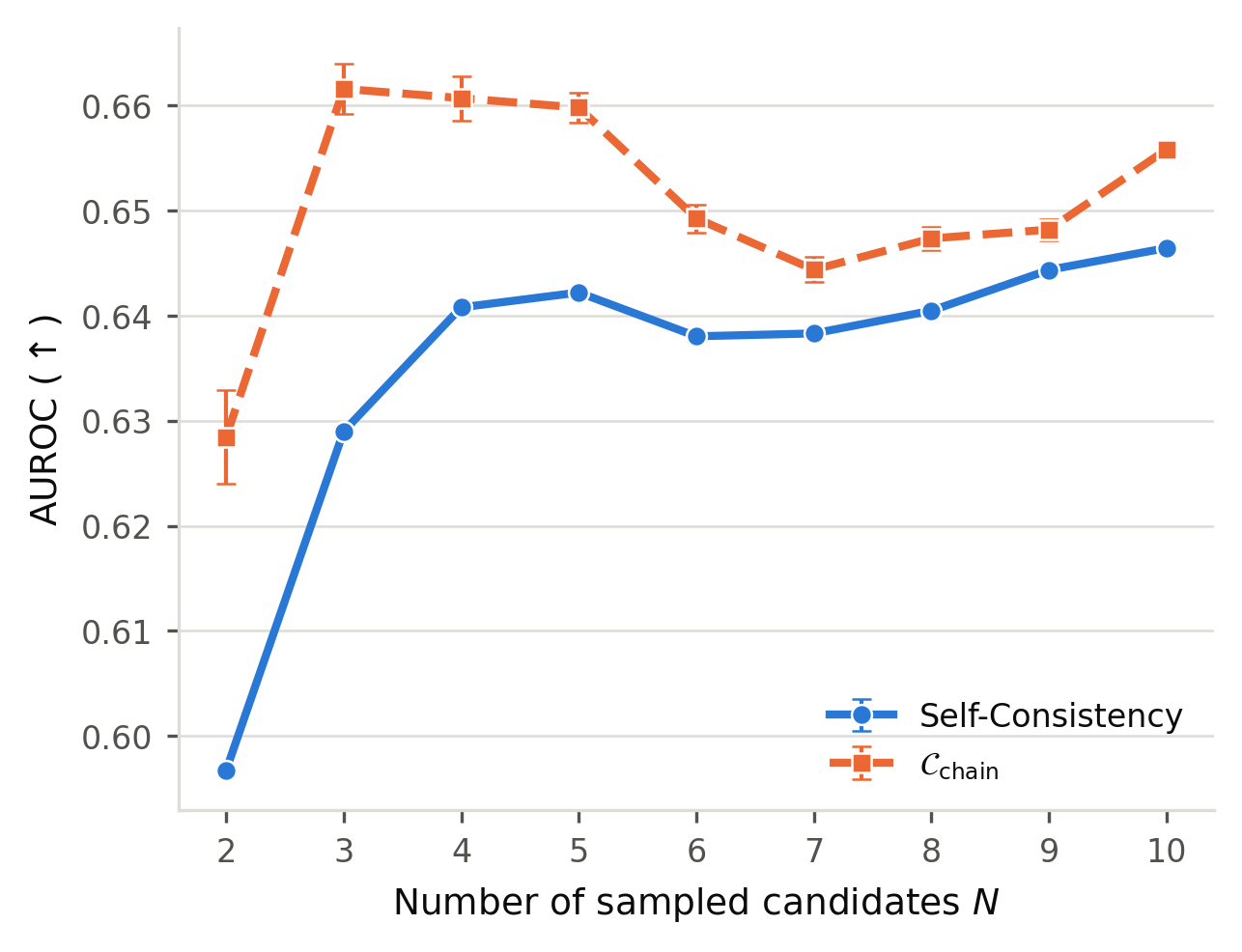}
\hfill
\includegraphics[width=0.32\textwidth]{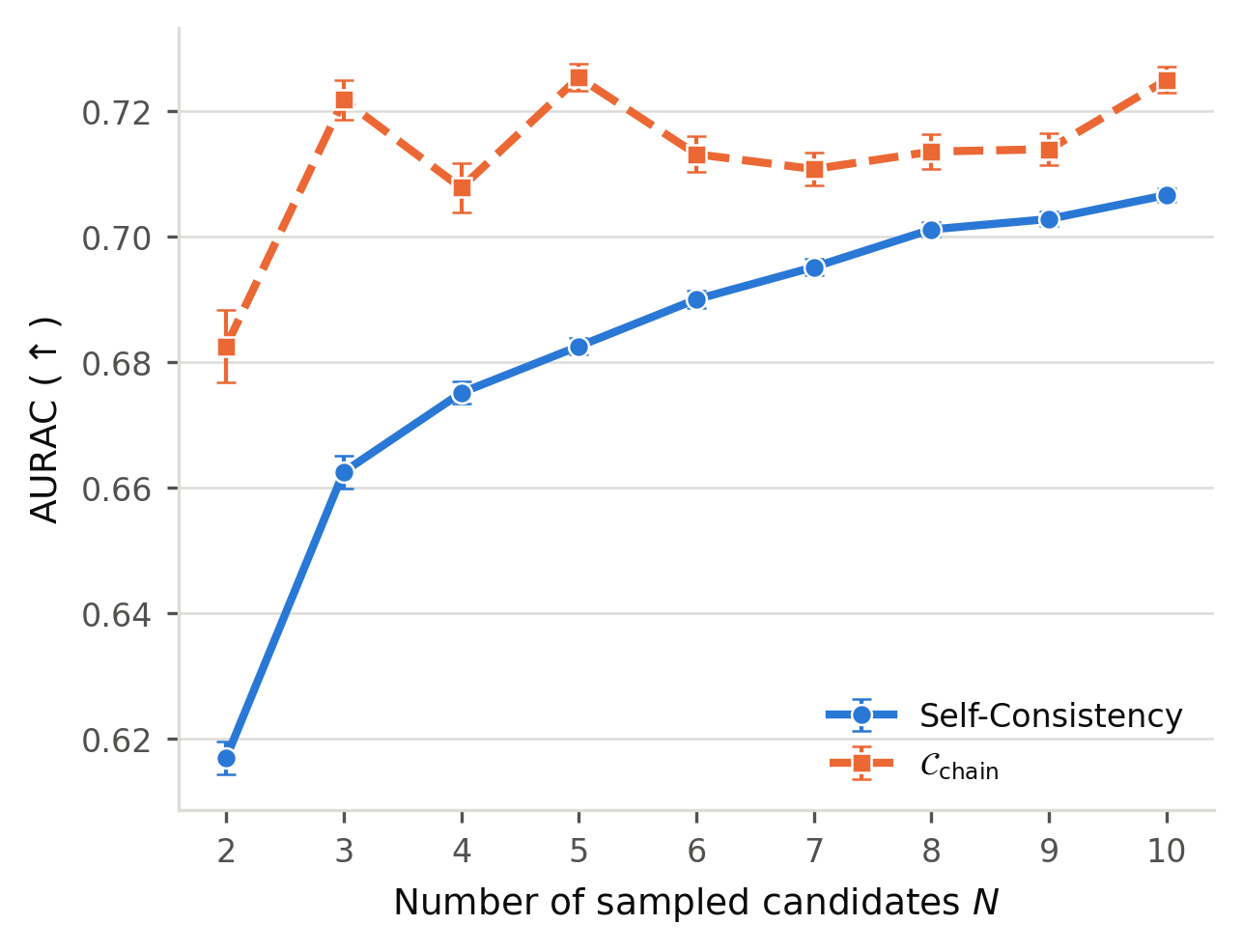}
\hfill
\includegraphics[width=0.32\textwidth]{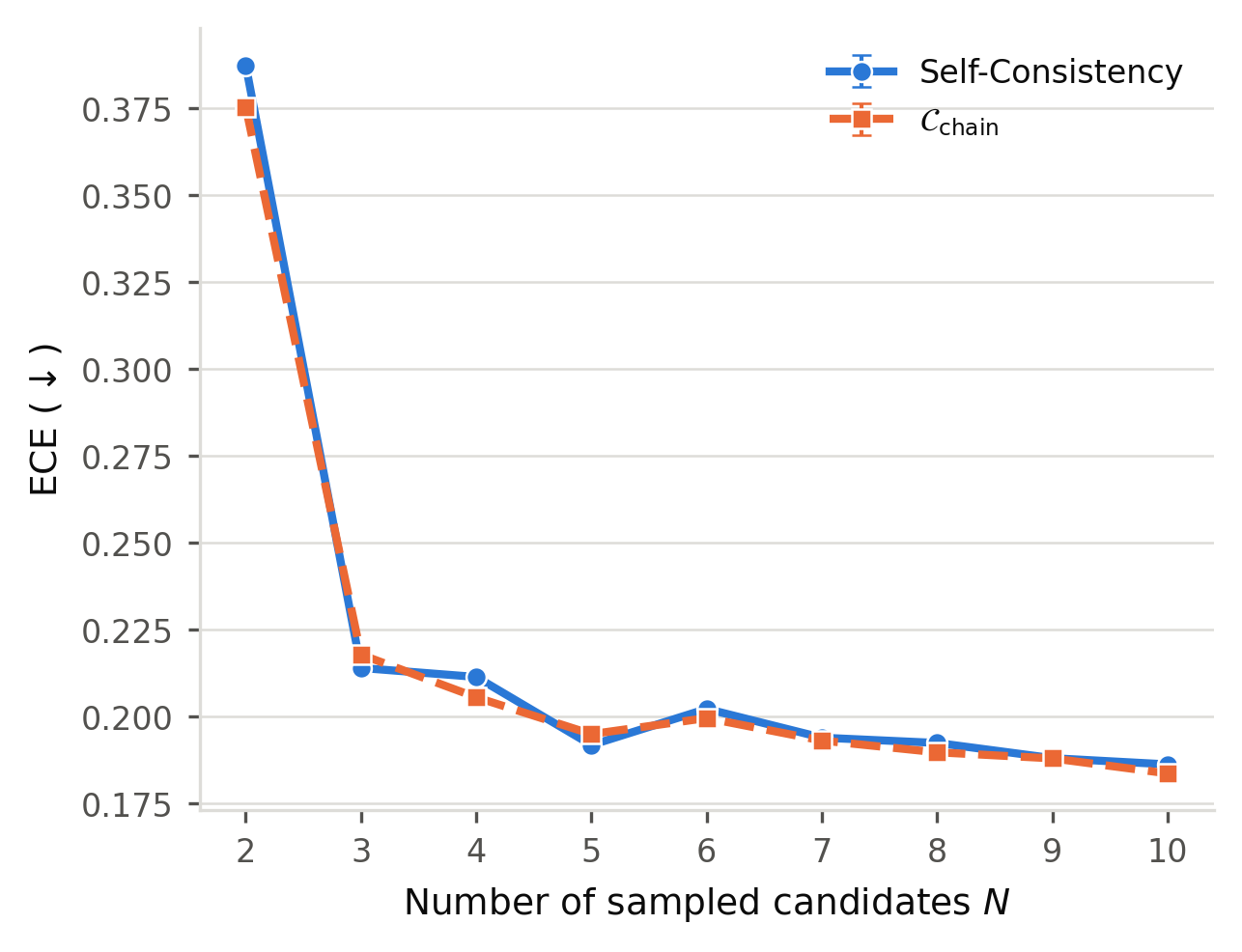}
\caption{AUROC ($\uparrow$), AURAC ($\uparrow$), and ECE ($\downarrow$) pooled over the union of splits (Appendix~\ref{app:overall}) as a function of the number of sampled candidates $N$ (MiniCPM-generated answers; means over 100 randomized evaluation runs per budget, error bars show $\pm 1$ standard error of the mean).}
\label{fig:n-sweep}
\end{figure*}

The sampling-based methods of this work fix the sampling budget at $N=6$ candidates per question. To assess how the comparison between plain self-consistency and the geometry-weighted variant of Appendix~\ref{app:gwsc} evolves with the sampling budget, we repeat that evaluation with MiniCPM-generated answers for $N \in \{2,\dots,10\}$, running 100 randomized evaluation repetitions per budget with the same protocol as the main experiments. Figure~\ref{fig:n-sweep} reports AUROC, AURAC, and ECE pooled over the union of splits as a function of $N$ for self-consistency and $\mathcal{C}_{\text{chain}}$. Pooled point estimates carry the heterogeneity caveat of Appendix~\ref{app:overall}; we rely on them here only for the \emph{paired} comparison between two methods evaluated on identical pools at each budget, for which the between-split component is shared.

Three observations emerge. First, $\mathcal{C}_{\text{chain}}$ exceeds self-consistency on AUROC and AURAC at every tested budget, while the two methods are essentially indistinguishable on ECE (differences within $0.012$, of either sign). Second, and most practically relevant, the advantage of geometry-weighted aggregation is largest in the low-budget regime: at $N=2$ the AUROC gap is $+3.2$ points (0.628 vs.\ 0.597) and the AURAC gap $+6.6$ points (0.683 vs.\ 0.617), and at $N=3$ they are $+3.3$ and $+5.9$ points respectively. This is intuitive: with few samples, raw vote counts are extremely coarse (with $N=2$, majority voting cannot distinguish most configurations at all), and the continuous embedding-based confidence provides exactly the tie-breaking signal that vote frequencies lack. Third, as $N$ grows the two curves converge, with $\mathcal{C}_{\text{chain}}$ retaining a small but consistent edge (e.g., AUROC 0.656 vs.\ 0.646 at $N=10$): vote frequencies become increasingly informative on their own, progressively subsuming the benefit of re-weighting. The variability of $\mathcal{C}_{\text{chain}}$ across runs is also larger at small $N$, reflecting the higher variance of fitting the aggregation weights on fewer effective votes. Taken together, these results indicate that our re-weighting is most valuable precisely where sampling is most expensive to scale---small candidate pools---which is the operative regime for costly omni-modal API models.

\section{Additional Baselines}
\label{app:additional-baselines}

To better isolate the contribution of our embedding-based signals from that of generic semantic-dispersion information between samples, we additionally compare against four established black-box baselines that, like self-consistency, operate on the $N$ sampled answers; the two dispersion-based ones (RDS and Semantic Volume) are also reported in the main Tables~\ref{tab:sf-gemini} and~\ref{tab:sf-minicpm}, and this appendix provides their in-depth analysis in the Gemini setting. The first two use the semantic clusters of the sampled answers. \textbf{Semantic Entropy} (SE) follows the discrete formulation of \cite{farquhar2024semanticentropy}: given the cluster assignments produced by the same clustering step used throughout our experiments, the confidence is $\exp(-H)$, where $H$ is the entropy of the empirical cluster distribution. \textbf{Number of Semantic Sets} (NumSets) uses the count of distinct clusters $k$ as an uncertainty signal, mapped to a confidence as $1-(k-1)/(N-1)$. The other two operate on continuous per-candidate \emph{answer embeddings}, computed with the same E5-Omni encoder used for our scores: \textbf{RDS} (radial dispersion score) is the mean Euclidean distance of the $N$ candidate answer embeddings from their centroid, and \textbf{Semantic Volume} is the log-determinant of the Gram matrix of the mean-centred embeddings, $\log\det(XX^{\top}+\varepsilon I)$ with $\varepsilon=10^{-6}$; both are uncertainty scores, negated to obtain confidences. All four scores are question-level and are evaluated with the identical protocol, preprocessing, and 100-run repetition scheme of the geometry-weighted experiments of Appendix~\ref{app:gwsc} (Gemini-generated answers); the Self-Consistency and $\mathcal{C}_{\text{chain}}$ rows are re-computed within the same runs and match the values of Appendix~\ref{app:gwsc} up to run-to-run noise. We further evaluate a \emph{hybrid} aggregate ($\mathcal{C}_{\text{chain}}$ + dispersion): the logistic-regression aggregator of $\mathcal{C}_{\text{chain}}$ refitted, under the same protocol, on seven features, i.e.\ the five chain-level scores plus the two dispersion scores added as question-level features. Table~\ref{tab:additional-baselines} reports AUROC and ECE.

\begin{table*}[t]
\centering
\begin{adjustbox}{width=\textwidth}
\begin{tabular}{l cc cc cc cc cc}
\toprule
& \multicolumn{2}{c}{Avg} & \multicolumn{2}{c}{Audio} & \multicolumn{2}{c}{MC} & \multicolumn{2}{c}{MO} & \multicolumn{2}{c}{Visual} \\
\cmidrule(lr){2-3} \cmidrule(lr){4-5} \cmidrule(lr){6-7} \cmidrule(lr){8-9} \cmidrule(lr){10-11}
Method & AUROC & ECE & AUROC & ECE & AUROC & ECE & AUROC & ECE & AUROC & ECE \\
\midrule
Self-Consistency & 0.5927 & 0.3068 & 0.4512 & 0.6203 & 0.5630 & 0.2723 & 0.6334 & 0.2459 & 0.7234 & \textbf{0.0888}\textsuperscript{***} \\
Semantic Entropy & 0.5882 & 0.3882 & 0.4471 & 0.6891 & 0.5531 & 0.3593 & 0.6319 & 0.3316 & 0.7207 & 0.1727 \\
Num.\ Semantic Sets & 0.5810 & 0.3106 & 0.4441 & 0.6237 & 0.5438 & 0.2914 & 0.6292 & 0.2274 & 0.7070 & 0.0998 \\
RDS & 0.6038 & 0.2405 & 0.6654 & \textbf{0.2760}\textsuperscript{***} & 0.5434 & \textbf{0.1382}\textsuperscript{***} & 0.5894 & 0.1951 & 0.6169 & 0.3526 \\
Semantic Volume & 0.6017 & \textbf{0.2211} & \textbf{0.6671}\textsuperscript{***} & 0.4689 & 0.5417 & 0.1802 & 0.5918 & \textbf{0.1223}\textsuperscript{***} & 0.6062 & 0.1130 \\
\midrule
$\mathcal{C}_{\text{chain}}$ & 0.6012 & 0.3014 & 0.4626 & 0.6107 & \textbf{0.5631} & 0.2744 & \textbf{0.6517}\textsuperscript{**} & 0.2269 & 0.7275 & 0.0937 \\
$\mathcal{C}_{\text{chain}}$ + dispersion & \textbf{0.6431} & 0.3474 & 0.6386 & 0.7205 & 0.5591 & 0.2964 & 0.6441 & 0.2474 & \textbf{0.7304}\textsuperscript{***} & 0.1253 \\
\bottomrule
\end{tabular}
\end{adjustbox}
\caption{AUROC ($\uparrow$) and ECE ($\downarrow$) of the additional black-box baselines across the UNO-Bench splits (Gemini-generated answers; means over 100 randomized evaluation runs). All methods in this table require the $N=6$ sampled candidates. Avg is the unweighted mean of the four per-split values (pooled-union results in Appendix~\ref{app:overall}). Bold indicates the best result per column; asterisks follow the 100-run paired $t$-test convention of Table~\ref{tab:auroc-ece} and are not attached to the derived Avg column.}
\label{tab:additional-baselines}
\end{table*}

Three findings emerge, one of which qualifies our earlier reading of this comparison. First, the discrete cluster-based baselines add nothing over plain self-consistency: SE tracks it closely on AUROC (0.5882 vs.\ 0.5927 on average) while degrading calibration, NumSets is uniformly weaker, and the same holds for AURAC (pooled values 0.8041 and 0.8002 vs.\ 0.8059; per-split AURAC was not stored for these baselines). For this family, self-consistency is effectively the best choice in its simplicity: with a modest sampling budget the cluster distribution carries essentially the same information as the majority-vote share, and re-parameterisations of it add nothing.

Second, and in contrast, the \emph{continuous} embedding-dispersion baselines are considerably stronger than we anticipated. Semantic Volume attains the best average ECE in the table (0.2211), and both RDS and Semantic Volume repair the Audio failure mode that affects every other sampling-derived method (AUROC 0.6654 and 0.6671\textsuperscript{***}, Audio AURAC up to 0.9017, where self-consistency sits below chance): the continuous dispersion of answer embeddings evidently carries signal on Audio that the discrete vote frequency inverts. Under the pooled-union protocol this Audio repair, combined with the between-split ranking effects analysed in Appendix~\ref{app:overall}, even hands Semantic Volume the best pooled AUROC of the table; on the split average its discrimination is instead on par with $\mathcal{C}_{\text{chain}}$ (0.6017--0.6038 for the two dispersion baselines vs.\ 0.6012). Inter-sample dispersion measured in the embedding space is therefore a genuinely informative signal beyond cluster agreement, and the improvements of our method over self-consistency cannot be dismissed as generic dispersion information only where these baselines are also strong. Crucially, however, the dispersion baselines collapse exactly where our within-chain scores are strongest: $\mathcal{C}_{\text{chain}}$ retains a clear advantage on the Visual (AUROC 0.7275\textsuperscript{***} vs.\ 0.6169), multiple-choice (0.5631 vs.\ 0.5434), and open-answer omni-modal (0.6517\textsuperscript{***} vs.\ 0.5918) splits, with the corresponding AURAC gaps equally wide (e.g.\ MO 0.5193 vs.\ 0.4734). The two signal families are thus complementary rather than redundant: between-sample dispersion excels where answer diversity is informative (Audio), while the geometry of the reasoning chain excels where the difficulty lies in assessing an individual answer against multimodal evidence. We also note that RDS and Semantic Volume, like self-consistency and our weighted scheme, require the full set of $N$ sampled candidates plus one embedding pass per candidate; unlike our method, however, they use only the answers' embeddings and discard the reasoning trace entirely. The MiniCPM-side results (Table~\ref{tab:sf-minicpm}) qualify the strength of this family further: with a generator whose agreement signal is not inverted on Audio, the dispersion baselines drop to the bottom of the sampling-based family on the split average (0.5764--0.5882) and fall to chance on both omni-modal splits, indicating that much of their Gemini advantage stems from repairing the Audio inversion.

Third, the hybrid aggregate gives an initial picture of how far simple feature-level fusion can combine the two families. Its dominant effect is to remove the worst-case failure of the chain-based aggregate: on Audio, where $\mathcal{C}_{\text{chain}}$ sits below chance, the hybrid reaches 0.6386 AUROC and 0.8529 AURAC (paired gains of $+0.176$ and $+0.038$ over $\mathcal{C}_{\text{chain}}$, both $p<10^{-70}$), and it additionally attains the best Visual AUROC in the table (0.7304\textsuperscript{***}) and, as a direct consequence of the Audio repair, the best average AUROC (0.6431). No split is left at or below chance, which is true of no other method in this work. These gains come at a cost: small but significant losses against $\mathcal{C}_{\text{chain}}$ on pooled AURAC ($-0.013$), open-answer AUROC ($-0.008$), and calibration on every split, together with visibly larger run-to-run variance, which we attribute to the aggregator having to arbitrate between heterogeneous signal families from roughly 150 validation questions per split. In turn, the hybrid does not reach the per-split maxima of the specialised methods (Semantic Volume retains the best Audio AUROC and average ECE, $\mathcal{C}_{\text{chain}}$ the best open-answer AUROC). We read this as evidence that the complementarity is real and exploitable, and that extracting it fully requires a more careful fusion than feature concatenation (e.g.\ per-split gating between the two families), which we leave to future work.

\section{Reproducibility Details}
\label{app:reproduce}
\subsection{Code and Data release}
The code required to replicate the results in this work can be accessed in an anonymised form at \url{https://anonymous.4open.science/r/Multimodal-CoT-Confidence-Scoring-C12E}.
The dataset used is, as mentioned, the publicly available UNO-Bench \citep{chen2025unobenchunifiedbenchmarkexploring}, with further description of pre-processing and details for reproducing the data on which we run our experiments available in Appendix \ref{app:experiments}.

\subsection{Software Environment}
All experiments were implemented in Python and executed using widely adopted open-source machine learning and scientific computing libraries. Table~\ref{tab:packages} summarizes the primary software dependencies used in our implementation.

\begin{table}[ht]
\centering
\small
\begin{tabular}{ll}
\toprule
\textbf{Package} & \textbf{Version} \\
\midrule
Python & 3.12 \\
numpy & 2.3.2 \\
pandas & 2.3.1 \\
scipy & 1.16.1 \\
scikit-learn & 1.7.1 \\
torch & 2.8.0 \\
torchvision & 0.23.0 \\
transformers & 4.56.0 \\
huggingface-hub & 0.34.4 \\
sentence-transformers & 5.1.0 \\
accelerate & 1.10.0 \\
tokenizers & 0.22.0 \\
safetensors & 0.6.2 \\
google-genai & 1.30.0 \\
tqdm & 4.67.1 \\
Pillow & 11.3.0 \\
opencv-python & 4.12.0 \\
matplotlib & 3.10.5 \\
seaborn & 0.13.2 \\
PyYAML & 6.0.2 \\
\bottomrule
\end{tabular}
\caption{Primary software dependencies used in our implementation.}
\label{tab:packages}
\end{table}

All pretrained models and datasets were obtained through their official releases via the Hugging Face ecosystem or the corresponding model providers. Experiments were conducted using deterministic random seeds whenever applicable. The complete source code, configuration files, and dependency specifications will be released to facilitate reproducibility.

\end{document}